%% file: acl_latex.tex
\pdfoutput=1

\documentclass[11pt]{article}

\usepackage[]{acl}

\usepackage{times}
\usepackage{latexsym}

\usepackage[T1]{fontenc}

\usepackage[utf8]{inputenc}

\usepackage{microtype}

\usepackage{subcaption}
\usepackage{multirow}
\usepackage{pifont}
\usepackage{amssymb}
\usepackage{bm}
\usepackage{amsmath}
\usepackage{booktabs}
\usepackage{makecell}
\usepackage{colortbl}
\usepackage{array}
\usepackage{graphicx}
\usepackage[figuresright]{rotating}
\usepackage{xspace} 
\usepackage{bbold}
\usepackage{tabularx}
\usepackage{fontawesome}

\usepackage{CJKutf8}

\definecolor{pinegreen}{RGB}{15,153,15}
\definecolor{mygray}{gray}{.9}
\definecolor{target}{RGB}{0,0,146}
\definecolor{myblue}{RGB}{25,101,255}
\definecolor{myorange}{RGB}{239,134,63}

\newcommand*{\circled}[1]{\lower.7ex\hbox{\tikz\draw (0pt, 0pt)%
		circle (.5em) node {\makebox[1em][c]{\small #1}};}}

\newcommand{\tbbg}{\cellcolor{gray!10}}
\newcommand{\tbbgg}{\cellcolor{gray!30}}

\newcommand{\bestd}[1]{\textcolor{bestd}{\textbf{#1}}}
\newcommand{\bestc}[1]{\textcolor{bestc}{\textbf{#1}}}
\definecolor{bestd}{RGB}{237,100,152}
\definecolor{bestc}{RGB}{0,126,219}

\newcommand{\myroman}[1]{\uppercase\expandafter{\romannumeral#1}}

\makeatletter
\newcommand*\bigcdot{\mathpalette\bigcdot@{.5}}
\newcommand*\bigcdot@[2]{\mathbin{\vcenter{\hbox{\scalebox{#2}{$\m@th#1\bullet$}}}}}
\makeatother

\newcolumntype{P}[1]{>{\centering\arraybackslash}p{#1}}

\newcommand{\logofig}{\includegraphics[height=0.7em,trim=0 2em 0 0]{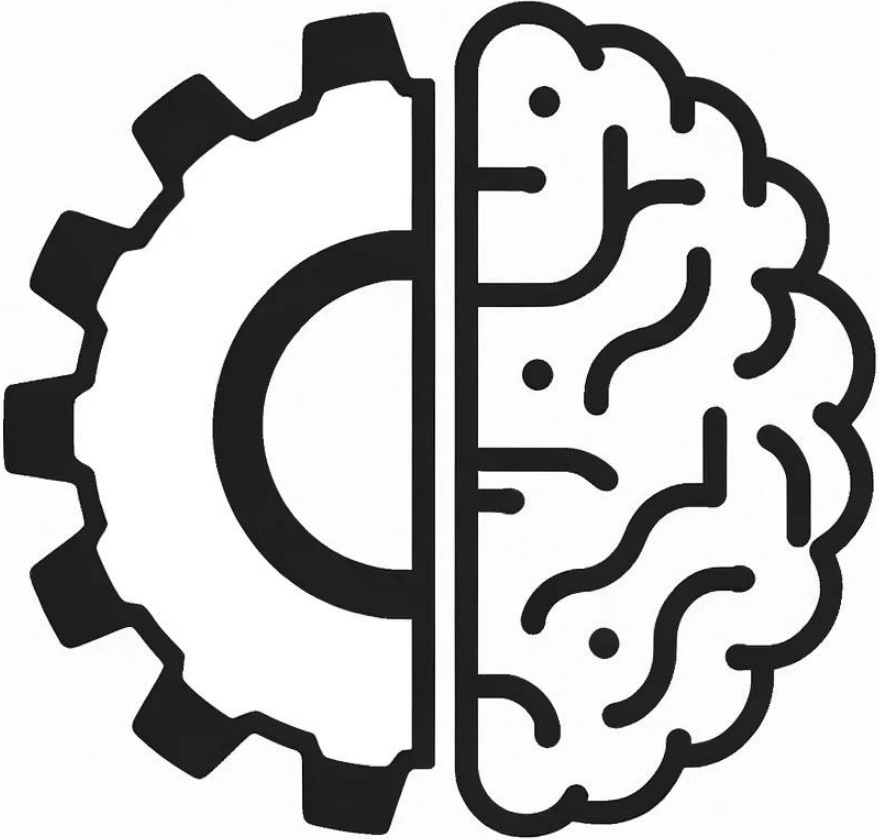}}
\newcommand{\logott}{\textsc{T}\logofig\textsc{MBench}\xspace}
\newcommand{\logo}{\textsc{T}\logofig\textsc{MBench}\xspace}

\newcolumntype{\myline}{!{\vrule width 0.08em}}

\title{\logott: Benchmarking Theory of Mind in Large Language Models}

 \author{Zhuang Chen$^1$\thanks{\quad Equal contribution.} \quad Jincenzi Wu$^2$\footnotemark[1]  \quad Jinfeng Zhou$^1$\footnotemark[1] \quad Bosi Wen$^1$\footnotemark[1] \\
\textbf{Guanqun Bi$^{1,3}$  \quad Gongyao Jiang$^4$\quad Yaru Cao$^{1,5}$}\\ \textbf{Mengting Hu$^6$  \quad Yunghwei Lai$^7$ \quad Zexuan Xiong$^1$} \quad \textbf{Minlie Huang$^1$} \thanks{\quad Corresponding author.}\\
$^1$CoAI Group, DCST, IAI, BNRIST, Tsinghua University\\ 
$^2$The Chinese University of Hong Kong \quad $^3$IIE, CAS \quad $^4$Tianjin University \\ 
$^5$Northwest Minzu University \quad $^6$Nankai University \quad $^7$Beijing Institute of Technology \\
{zhchen-nlp@mail.tsinghua.edu.cn \quad aihuang@tsinghua.edu.cn}}

\begin{document}
\maketitle
\begin{abstract}
Theory of Mind (ToM) is the cognitive capability to perceive and ascribe mental states to oneself and others. Recent research has sparked a debate over whether large language models (LLMs) exhibit a form of ToM. However, existing ToM evaluations are hindered by challenges such as constrained scope, subjective judgment, and unintended contamination, yielding inadequate assessments. To address this gap, we introduce \logo with three key characteristics: a systematic evaluation framework encompassing 8 tasks and 31 abilities in social cognition, a multiple-choice question format to support automated and unbiased evaluation, and a build-from-scratch bilingual inventory to strictly avoid data leakage. 
Based on \logo, we conduct extensive experiments to evaluate the ToM performance of 10 popular LLMs across tasks and abilities. We find that even the most advanced LLMs like GPT-4 lag behind human performance by over 10\% points, indicating that LLMs have not achieved a human-level theory of mind yet. 
Our aim with \logo is to enable an efficient and effective evaluation of LLMs' ToM capabilities, thereby facilitating the development of LLMs with inherent social intelligence.

\end{abstract}

\section{Introduction}
Theory of Mind (ToM) is a fundamental cognitive process, involving the capability to attribute mental states – beliefs, intents, desires, emotions, knowledge, etc. – to oneself and others \cite{premack1978tom}. 
ToM is essential for human social cognition \cite{baron1985does} and plays an important role in social activities like empathetic communication \cite{decety2004functional}, relationship maintenance \cite{slaughter2002theory}, decision making \cite{carlson2001individual}, and childhood education \cite{caputi2012longitudinal}. 


With the advent of the era of large language models (LLMs), powerful LLMs like GPT-4 \cite{achiam2023gpt} and LLaMA \cite{touvron2023llama} have demonstrated comparable performance to humans in solving tasks. Consequently, researchers have grown increasingly curious that ``\textit{Do large language models have a theory of mind?}'', while the subsequent assessments have yielded inconsistent and even contradictory results.  \citet{kosinski2023theory} and \citet{bubeck2023sparks} find that GPT-3.5/4 achieve remarkably high performance on the false belief task \cite{wimmer1983beliefs}, suggesting that ToM may have spontaneously emerged in LLMs. However, later studies reveal that LLMs' ToM performance dramatically decreases when faced with trivial alterations \cite{ullman2023large,shapira2023clever}. This suggests a phenomenon called ``Clever Hans'' \cite{lapuschkin2019unmasking} denoting that LLMs rely on unexpected spurious correlations \cite{simon1977spurious} rather than truly possessing a ``Mind Reading'' ability.
This ongoing debate underscores the pressing need for a holistic ToM benchmark.

\begin{figure*}[t]
	\centering
	\vspace{0mm}
	\hspace{-4mm}
	\setlength{\abovecaptionskip}{0mm}
	\setlength{\belowcaptionskip}{0mm}
	\includegraphics[width=0.95\linewidth]{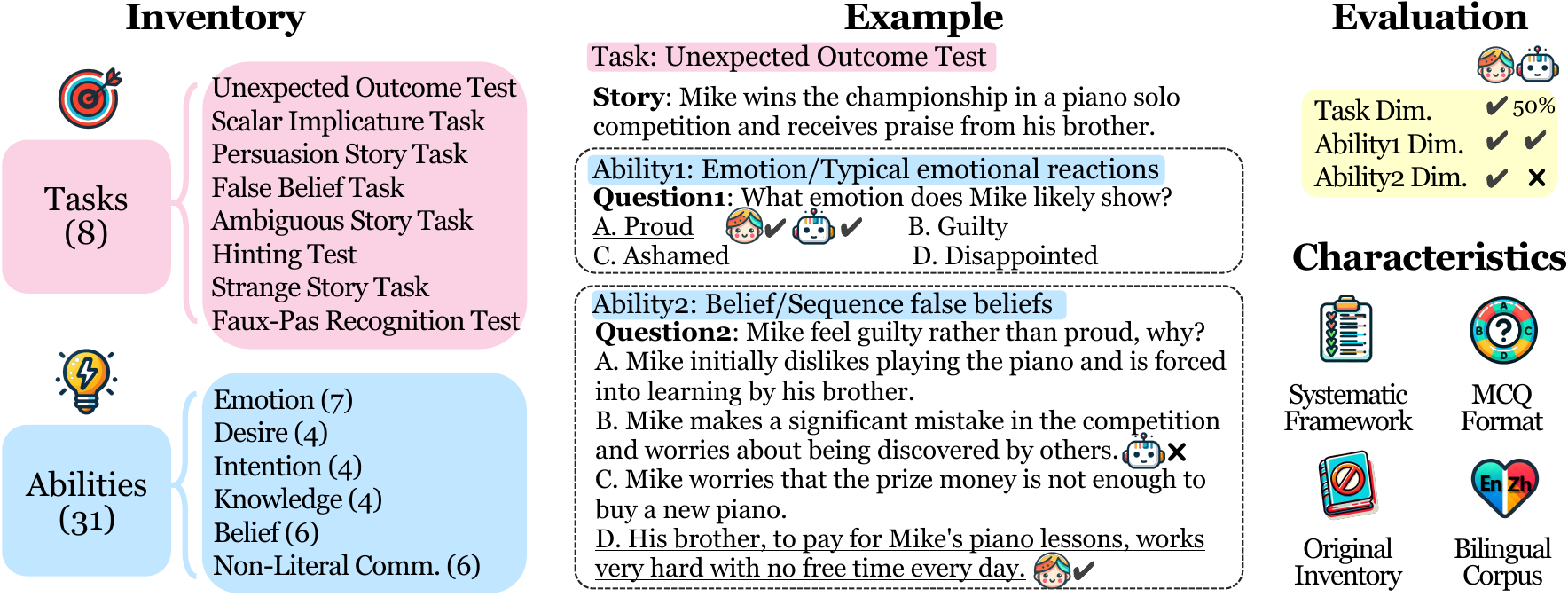}
	\centering
	\caption{\logo is a systematic, automated, and original bilingual ToM benchmark for LLMs, covering 8 tasks and 31 abilities. \logo contains 2,860 testing samples involving diverse real-world social scenarios.}
	\label{fig-intro}
	\vspace{-3mm}
\end{figure*}

Despite its critical importance, the development of such a benchmark still faces three major challenges.
\textbf{1) Constrained scope}. Existing studies mainly evaluate LLMs on specific ToM dimensions like emotions \cite{sap2019socialiqa} and beliefs \cite{wu2023hi}. While inspiring, they fall short of providing a complete ToM assessment. 
\textbf{2) Subjective judgment}. Most existing inventories for ToM evaluation involve open-ended questions where responses require manual scoring by experts \cite{beaudoin2020systematic}. 
While this approach works on small-scale and focused ToM tests for children, it is impractical for benchmarking due to the high annotation cost required for comprehensively evaluating LLMs. Additionally, human evaluations towards diverse answers may introduce potential inconsistency and bias \cite{klie2023annotation}. 
\textbf{3) Unintended contamination}. Although merely combining different inventories can create a corpus involving multiple ToM aspects, it is still not qualified for being a benchmark. Since most ToM inventories in psychology research are released far before the launch of LLMs, 
they are likely to have appeared in the training process of LLMs, whether during pre-training, SFT, or RLHF \cite{golchin2023time,li2023task}. We present a simple verification in Appendix \ref{appendix-contamination}. The potential contamination indicates a risk that LLMs might complete ToM evaluations in an ``open-book'' status, leading to inflated performance \cite{shapira2023clever}.


To address the above challenges,  we propose \logott, a ToM benchmark for LLMs featuring three key characteristics.
\textbf{1) Systematic framework}. We first review broad psychological literature and identify 8 well-defined theory-of-mind tasks in \logo. Then, since tasks are the means and abilities are the core, we further ground \logo in a well-structured psychological framework, ATOMS \cite{beaudoin2020systematic}, expanding our evaluation to encompass 31 core theory-of-mind abilities. We hereby design both task- and ability-oriented evaluations to comprehensively assess the ToM capabilities of LLMs.
\textbf{2) Multi-choice question format}.
We define every test sample in \logo as a story, followed by a question, and several plausible options where only one answer is correct and the others are high-quality but misleading wrong answers. This multiple-choice question (MCQ) format enables automated and efficient evaluation, avoiding the high labor costs and potential inconsistencies in manual scoring.
\textbf{3) Build-from-scratch inventory}. We do not use any existing inventories. Instead, after being trained by professional psychologists who are proficient in ToM, we authors draw inspiration from daily social scenarios and manually create 2,860 original test samples, each available in both Chinese and English. 
We also implement strict annotation and validation procedures to ensure the data quality.
The resulting inventory includes the above-mentioned 8 tasks and 31 abilities for ToM evaluation, covering a wide range of real-world human social scenarios across diverse topics.


Using \logott, we conduct both task- and ability-oriented evaluations of 10 popular LLMs with vanilla and CoT prompting. We also recruit 20 graduate students to provide an expected human baseline ToM performance. The experimental results show that, the GPT-4 series outperforms other LLMs across ToM tasks and abilities, yet their average performance is significantly lower than human levels by over 10\% points. 
Further analysis shows that when facing a more rigorous and comprehensive understanding of social scenarios, the gap in ToM performance between LLMs and humans widens. Moreover, inspecting the internal attention of LLMs reveals their decision-making processes notably differ from those of humans.

\begin{figure*}[t]
	\centering
	\vspace{0mm}
	\hspace{0mm}
	\setlength{\abovecaptionskip}{1mm}
	\setlength{\belowcaptionskip}{1mm}
	\includegraphics[width=1.0\linewidth]{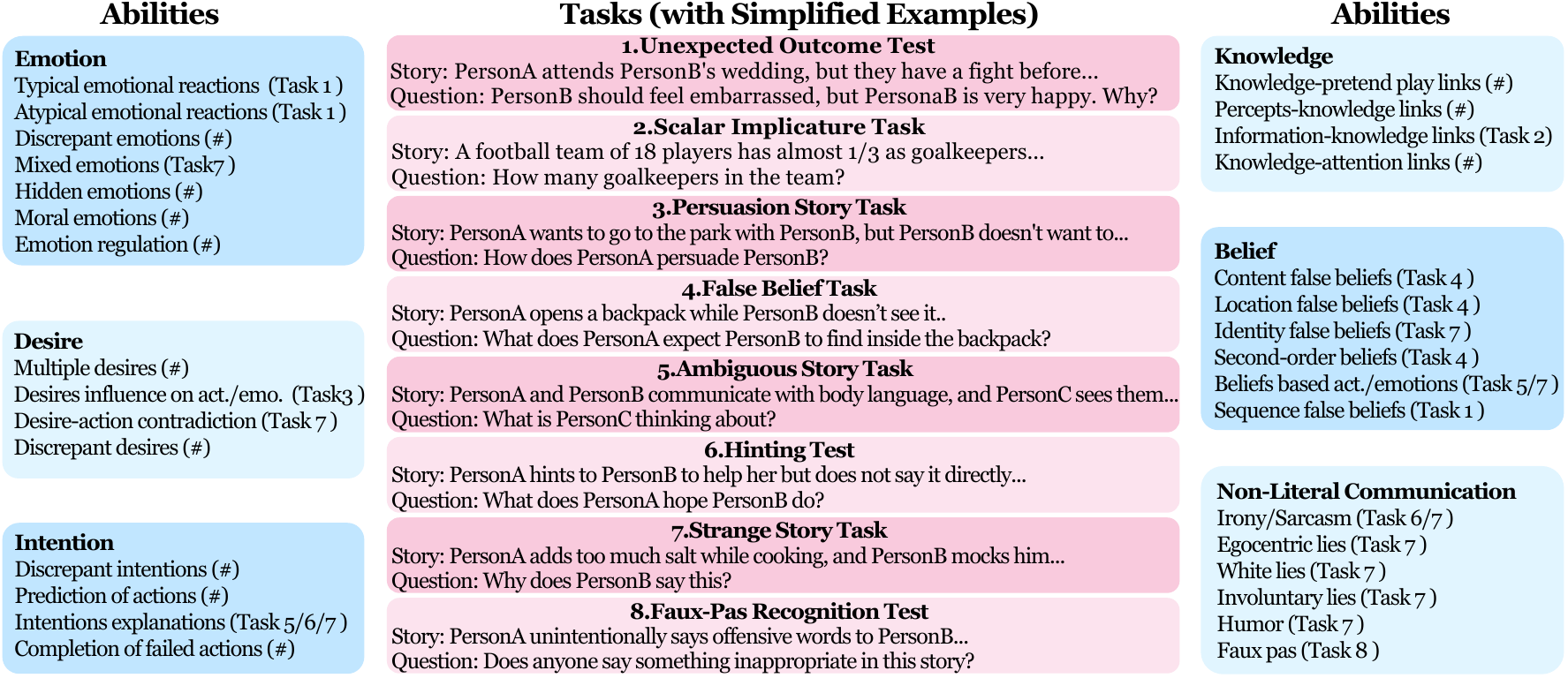}
	\centering
	\caption{The mapping between 8 tasks and 31 ATOMS abilities. The suffix after each ability indicates its occurrence within specific tasks, whereas those with ``\#'' are not covered by tasks and are evaluated with extra test samples.}
	\label{fig-atmos}
	\vspace{-4mm}
\end{figure*}

To the best of our knowledge, \logo is the first systematic ToM benchmark for automated ToM evaluation of LLMs. 
We hope that \logo will drive construction through evaluation, facilitating the development of LLMs in the realm of social intelligence, thereby complementing their task-solving capabilities. A broader objective is to enable LLMs to transcend the superficial layer of natural language and delve deeper into understanding the mental states and cognitive processes behind human words,
hereby foster more efficient and seamless human-AI interaction. We will release the code and data of \logo at \url{{https://github.com/zhchen18/ToMBench}}.

\section{\logo Framework}

\subsection{8 Theory-of-Mind Tasks}

The essence of the theory of mind is \textit{social understanding}, which refers to the ability of humans to perceive and ascribe mental states to themselves and others in social scenarios. In \logo, we first identify 8 social cognitive tasks that have been widely proven in psychology to effectively assess ToM capabilities: \textbf{Unexpected Outcome Test} \cite{dyck2001autism}, \textbf{Scalar Implicature Task} \cite{goodman2013knowledge}, \textbf{Persuasion Story Task} \cite{kolodziejczyk2016young}, \textbf{False Belief Task} \cite{wimmer1983beliefs}, \textbf{Ambiguous Story Task} \cite{bosacki1999theory}, \textbf{Hinting Test} \cite{corcoran1995schizophrenia}, \textbf{Strange Story Task} \cite{happe1994advanced}, \textbf{Faux-pas Recognition Test} \cite{baron1999recognition}. We then construct test samples for these tasks based on definitions, descriptions, and examples in the original psychology literature. 
Due to space constraints, we present the details of all tasks in Appendix \ref{appendix-detail-task}.

\subsection{31 Theory-of-Mind Abilities} 

Drawing from expertise in psychology, we gain insight into the principle that tasks are means, while abilities are the core \cite{quesque2020theory}. 
Thus, we further resort to a well-defined psychological framework, ``Abilities in the Theory-of-Mind Space (ATOMS)'' \cite{beaudoin2020systematic}. 
Originally designed for an extensive analysis of psychological ToM measures towards young children, ATOMS outlines 7 distinct ability dimensions: \textit{Emotion}, \textit{Desire}, \textit{Intention}, \textit{Percept}, \textit{Knowledge}, \textit{Belief}, and \textit{Non-literal Communication}, embracing 39 specific theory-of-mind abilities. After removing the \textit{Percept} dimension requiring visual cues and some mixed abilities, we retain a total of 6 dimensions and 31 abilities in \logo. 

However, as shown in Figure \ref{fig-atmos}, 
the above-mentioned 8 tasks only encompass 19/31 abilities in ATOMS. 
Therefore, to address the remaining 12 abilities, we again refer back to the original literature to supplement extra ability-specific test samples, thereby extending the evaluation scope of \logo to a complete set of 31 abilities. Due to space limitation, we here briefly introduce 6 ability dimensions, while the complete details of 31 specific abilities can be found in Appendix \ref{appendix-detail-ability}.



\textbf{1) Emotion} involves \texttt{7} abilities to understand that situational factors influence people's emotional states, that people can experience complex emotions, and that people can regulate emotional expressions. 
\textbf{2) Desire} involves \texttt{4} abilities to understand that people have subjective desires, preferences, and wants that influence their emotions and actions. 
\textbf{3) Intention} involves \texttt{4} abilities to understand that people undertake actions in pursuit of goals and intentions. 
\textbf{4) Knowledge} involves \texttt{4} abilities to understand that others have access to different knowledge based on their perceptions, information they have received, or familiarity with things. 
\textbf{5) Belief} involves \texttt{6} abilities to understand that people can hold beliefs about the world that are different from reality or different from one's own beliefs. 
\textbf{6) Non-literal Communication} involves \texttt{6} abilities to understand that communication can convey meaning beyond the literal words spoken. 

\section{\logo Construction}
\subsection{Overview}

\paragraph{Principles} We authors act as workers and build \logo from scratch. We do not use any existing inventories from psychological literature due to the potential risk of data contamination and the limited size of test samples. To ensure the effectiveness of inventory, all workers have undergone training by psychology experts, gaining a solid understanding of ToM. Furthermore, all workers have thoroughly reviewed the related psychology literature, adhering strictly to the specific definitions, descriptions, and examples of tasks and abilities. 

\begin{figure}[t]
	\centering
	\vspace{0mm}
	\hspace{0mm}
	\setlength{\abovecaptionskip}{2mm}
	\setlength{\belowcaptionskip}{2mm}
	\includegraphics[width=0.57\textwidth]{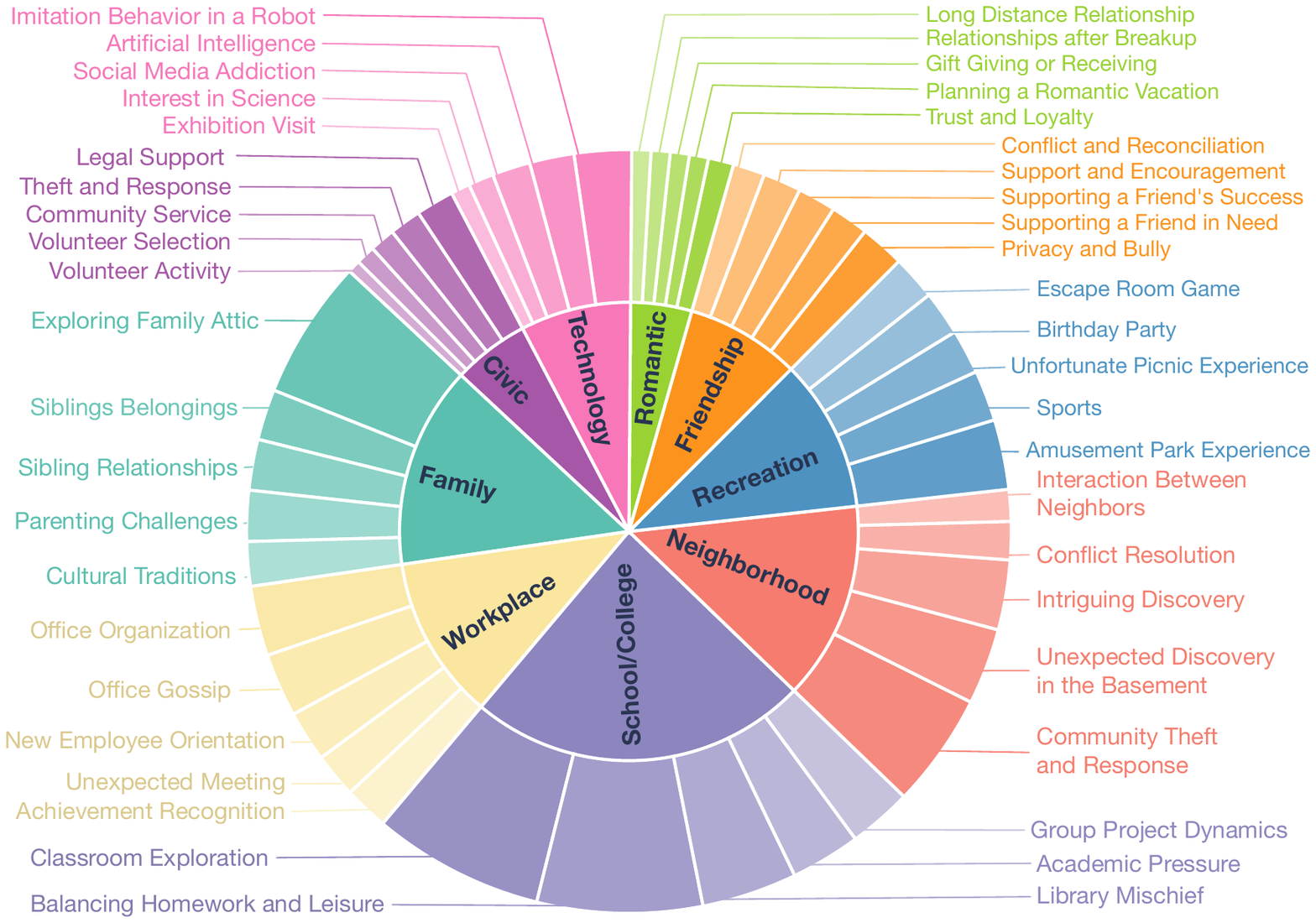}
	\centering
	\caption{Topics of social scenarios in \logo. Under 9 primary topics, we highlight the top-5 sub-topics with the highest frequency.}
	\label{fig-topic}
	\vspace{-6mm}
\end{figure}

\paragraph{Procedure} We first craft samples for 8 tasks, with each worker responsible for at least one, covering 19 abilities in total. Owing to the natural difficulty of tasks and abilities, we ensure that each task has $\geq$100 samples, and each ability within a task has $\geq$20 samples. This step results in the creation of 2,470 samples. Subsequently, we add extra samples for the remaining 12 abilities not previously covered, with $\geq$20 samples for each ability, bringing the total number of samples to 2,860.

\paragraph{Statistics} In Table \ref{tab-statistics}, we present the data statistics of \logo. Due to space constraints, the detailed statistics of the 31 abilities are present in Appendix \ref{appendix-statistics}. 
As shown in Figure \ref{fig-topic}, \logo includes diverse daily topics, such as \textit{school}, \textit{workplace}, \textit{family}, \textit{neighborhood}, etc. These real-life social scenarios are crucial for an effective ToM evaluation, just as we mentioned earlier, the essence of the theory of mind is {social understanding}.





\begin{table}[t]
 \vspace{0.8mm}
	\hspace{6mm}	
	\setlength{\abovecaptionskip}{4mm}
	\setlength{\tabcolsep}{0.5mm}
	\resizebox{0.43\textwidth}{!}{
		\begin{tabular}{llllll}
			\toprule
			& \textbf{\#S} & \textbf{\#Q} & \makecell{\textbf{ASL}\\\textbf{(En)}} & \makecell{\textbf{ASL}\\\textbf{(Zh)}}  & \textbf{Agr.}\\ \hline
			\textbf{Task View} & 934 &  2,470 & 61.22 & 97.69  & 99.4\% \\ \hline
			Unexpected Outcome Test & 100 & 300 & 38.46 & 62.01 & 100.0\% \\
			Scalar Implicature Task& 100 & 200 & 47.17 & 76.89 & 100.0\% \\
			Persuasion Story Task & 100 & 100 & 36.58 & 51.35 & 95.0\% \\
			False Belief Task & 100 & 600 & 49.15 & 77.54 & 100.0\% \\
			Ambiguous Story Task & 100 & 200 & 102.57 & 164.07 & 100.0\% \\
			Hinting Test & 93 & 103 & 49.63 & 79.92 & 100.0\% \\
			Strange Story Task & 201 & 407 & 70.42 & 112.97 & 100.0\% \\
			Faux-pas Recognition Test & 140 & 560 & 95.77 & 156.79 & 98.2\% \\ \midrule		
			\textbf{Ability View} & 1,584 & 2,860 & 66.57 & 107.21 & 99.4\% \\ \midrule
			Emotion & 300 & 420 & 52.34 & 83.50 & 99.8\% \\
			Desire & 160 & 180 & 50.19 & 74.91 & 97.2\% \\
			Intention & 273 & 340 & 82.56 & 131.20 & 100.0\% \\
			Knowledge & 170 & 290 & 56.38 & 94.26 & 100.0\% \\
			Belief & 440 & 882 & 55.70 & 88.99 & 100.0\% \\
			Non-Literal Communication & 241 & 748 & 88.02 & 143.91 & 99.4\% \\ \bottomrule
			
		\end{tabular}
	}
	\caption{Data statistics. {\#S}: Number of stories, {\#Q}: Number of questions, {ASL(En/Zh)}: Average story length (English/Chinese). Agr.: Final agreement.}
	\label{tab-statistics}
	\vspace{-3.6mm}
\end{table}



\subsection{Data Collection}


We here detail the process of data collection. We construct \logo in the form of multiple-choice questions to avoid the high costs of manual scoring,  and ensure an unbiased and consistent evaluation. Each sample is defined as a combination of a {story}, a {question}, and several {options}.

\paragraph{Story} \hspace{-3mm}describes a scenario from everyday life, including characters' actions and interactions, to set the context for evaluation. The inspiration for stories primarily comes from posts on social platforms like Reddit, Twitter, Zhihu, and Weibo, such as ``\textit{What kind of hints have you understood from others}?'' The design of stories refers to the setting in psychology literature while ensuring variety in social scenarios.
\paragraph{Question} \hspace{-3mm}asks subjects to understand specific aspects of the story, which strictly adheres to psychological definitions of ToM tasks and abilities. Every question is designed to be answerable by humans. A story can correspond to multiple questions, each exploring different aspects of a social scenario to assess a comprehensive understanding, where each question represents a unique test sample. 
\paragraph{Options} \hspace{-2mm}include one correct answer and several misleading incorrect answers. The incorrect options are designed to be plausible, avoiding outliers that are easily dismissed. Options are generally of two types: for true/false questions, such as ``\textit{Is what PersonA said true?}'', the options are simply yes/no. For explanatory questions, such as ``\textit{Why did PersonB say this?}'', there are four options provided.

\subsection{Data Validation \& Translation}
After the data collection, we conduct two rounds of validation to ensure the data quality. In 1st round, worker A would first complete all samples created by worker B. For stories, questions, and options where there are disagreements, workers A and B would discuss and modify them to reach a consensus as much as possible. In 2nd round, for samples where consensus is still not reached, another worker C would discuss with workers A and B to determine the final answer. After two rounds of discussion, the final average agreement reaches 99.4\%. The inventory is initially crafted in Chinese. We then carefully use GPT-4-0613 to translate it into English, and manually check all translated samples to support bilingual ToM evaluation. Note that we do not provide correct answers in translation, thus there is no data leakage. The translation prompts can be found in Appendix \ref{appendix-translation}. Since we authors act as workers, we do not pay any other individuals for data collection.






\input{tables/task-results-soft.tex} 

\subsection{Evaluation Method}
In \logo, we organize the test samples in two perspectives: the task view groups them into 8 theory-of-mind tasks, and the ability view categorizes them into 31 specific theory-of-mind abilities. The former approach is more general and commonly used in psychology research; the latter is more comprehensive, allowing us to inspect the performance of each specific ability. Accordingly, we report both \textbf{task-oriented} and \textbf{ability-oriented} results by averaging the samples related to a specific task or ability. For evaluation, we present LLMs with a story, a question, and several options, then ask them to pick the correct answer.

\section{Experiments}

\subsection{Experimental Setup}
We evaluate a total of 10 popular LLMs, including GPT-4-1106 \cite{GPT4-1106}, GPT-4-0613 \cite{achiam2023gpt}, GPT-3.5-Turbo-1106 \cite{GPT4-1106}, GPT-3.5-Turbo-0613 \cite{GPT35-0613}, ChatGLM3-6B \cite{ChatGLM3}, LLaMA2-13B-Chat \cite{touvron2023llama}, Baichuan2-13B-Chat \cite{Baichuan2}, Qwen-14B-Chat \cite{bai2023qwen}, Mistral-7B-v0.2 \cite{jiang2023mistral}, and Mixtral-8x7B-Instruct-v0.1 \cite{Mixtral}. For GPT* and other open LLMs, we strictly abide by their terms and get access through official APIs and model weights, respectively.

We employ two prompting methods: the \textbf{vanilla} prompting directly asks LLMs to give a choice, while the \textbf{CoT} prompting elicits step-by-step reasoning. The prompts are detailed in Appendix \ref{appendix-prompt}.
For all models (except GPT-4-*), to avoid the bias from option IDs \cite{zheng2023large}, we shuffle the option orders five times and choose the most frequently selected option as the final answer. 
For the GPT-4-* models, our pilot experiments show very consistent answers across different option orders, so we only use the result from one round of answering.  Accuracy is used as the metric. 

To establish a human baseline, we recruit 20 native graduate students (each paid with \$15) to complete the Chinese \logo together. No extra tutorials or examples are provided to ensure a fair comparison. We directly use this result as human performance since studies have found that cultural and language differences do not significantly affect ToM task performance across native-English and native-Chinese speakers \cite{bradford2018cross}.

\input{tables/ability-results-category.tex}

\subsection{Main Results}
We show the ToM performance of LLMs across 8 tasks (Table \ref{tab-task-soft}) and 6 ability categories (Table \ref{tab-ability-category}), with detailed results on the 31 specific abilities presented in Appendix \ref{appendix-all-ability-results}. We now dissect the results and highlight several critical observations.
\paragraph{Human vs. LLMs}
The average ToM performances of all LLMs are significantly lower than that of humans, with the smallest gap being 10.1\% in task view (Human 85.4\% vs. GPT-4-1106 75.3\%) and 10.8\% (Human 86.1\% vs. GPT-4-0613 + CoT 75.3\%) in ability view, respectively. Interestingly, in the false belief task (\textbb{FBT}), LLMs like GPT-4-1106 even outperform humans, which we believe is explainable. In existing ToM inventories, there are relatively abundant samples for \textbb{FBT}, such as the Sally-Anne \cite{baron1985does} and Smarties \cite{wimmer1983beliefs} tests. Moreover, the stories used in \textbb{FBT} are strictly templated, which further benefits LLMs' generalization after being trained on existing samples. For example, they often involve PersonA moving something to a new location without PersonB knowing, and then asking about PersonB's belief.  It's normal for LLMs to surpass humans in specific tasks and abilities, but this shouldn't lead us to conclude that LLMs possess a human-level ToM. Instead, we should consider a general and robust ToM across tasks and abilities as the criterion for assessment.


\paragraph{LLMs' ToM Performance}
GPT-4 series stands out with superior results in LLMs, and surpasses the second-tier GPT-3.5-Turbo series by over 10\% points. 
Among the open LLMs, Qwen-14B-Chat impressively outperforms other LLMs and even closely matches GPT-3.5-Turbo. Mistral-7B is also notable, not only surpassing ChatGLM3-6B but even outperforming LLaMA2-13B-Chat and Baichuan2-13B-Chat. 
The MoE-type Mixtral-8x7B 
does not outperform Mistral-7B, which contrasts to its task-solving performance and also underscores the unique nature of ToM evaluation. 

\paragraph{Vanilla vs. CoT Prompting}
Both task- and ability-oriented results indicate that CoT prompting almost never improves ToM performance and can even lead to declines. We believe CoT reasoning primarily serves to provide intrinsic context, breaking down complex tasks into simpler ones. However, if an LLM inherently lacks a robust ToM, its CoT decomposition process may struggle to align with genuine human cognitive reasoning, thereby failing to boost performance. We present a failure case of CoT prompting in Appendix \ref{appendix-cot-case}.

\paragraph{Differences Across ToM Tasks}
In Table \ref{tab-task-soft}, we calculate the ``Grand Mean'' performance of all LLMs on specific tasks. Notably, the {faux-pas recognition test} (\textbb{FRT}) shows relatively high accuracy, partly because \textbb{FRT} includes a notable number of True/False questions, reducing its difficulty. The {unexpected outcome test} (\textbb{UOT}) follows, mainly because a substantial portion of \textbb{UOT} involves direct emotion understanding tasks, a domain where NLP has abundant datasets, thus LLMs perform well. The poorest performance is observed in the {scalar implicature task} (\textbb{SIT}), which requires understanding a character's grasp of information based on quantifiers, involving mathematical reasoning where LLMs are typically weak. The {persuasion story task} (\textbb{PST}) also proves challenging, focusing on the complex psychological activities of characters persuading others and requiring a comprehensive understanding of social contexts. 

\paragraph{Differences Across ToM Abilities}
Further exploring Table \ref{tab-ability-category}, the best-performing ability dimensions are \textit{Non-literal Communication} (overlaps with \textbb{FRT}) and \textit{Emotion}, consistent with the task-oriented results. The weakest category is \textit{Knowledge} (overlaps with \textbb{SIT})—not referring to the LLMs' own knowledge but testing whether LLMs understand the information propagation between characters in stories, which LLMs often struggle with. In Appendix \ref{appendix-all-ability-results}, we delve deeper into LLMs' performance across the 31 abilities.

\paragraph{The Necessity of Bilingual Inventory}
The averaged results show LLMs performing similarly in Chinese and English, 
but this does not imply that LLMs perform identically in both languages. To assess the significance of any differences, we first conduct a Kolmogorov-Smirnov Test \cite{an1933sulla} on Chinese and English results, finding that neither task-oriented nor ability-oriented results follow a normal distribution (\textit{p}<0.01). Thus, we employ the Wilcoxon Signed-Rank Test \cite{wilcoxon1947individual} to examine the differences caused by languages, yielding \textit{p}=0.019 (task) and \textit{p}=0.016 (ability), both below the significance threshold of 0.05. Obviously, although the bilingual data are generated through translation with minimal semantic differences, LLMs exhibit significant performance discrepancies between Chinese and English contexts.

\subsection{In-Depth Analysis}

\paragraph{A Harder Coherent Test for LLMs}
When evaluating task-oriented ToM performance, we use the simple average accuracy across all questions at that time. In fact, for all ToM tasks except \textbb{PST}, most stories are associated with multiple coherent questions probing different aspects of the social scenario within the story. A common and intuitive perspective is that for a story, subjects need to answer all coherent questions correctly to demonstrate a complete understanding \cite{kim2023fantom}, rather than just making educated guesses. Therefore, we further design a more challenging story-level coherent test for evaluation, where an LLM is considered to have failed a story if it incorrectly answers any question associated with that story. 

\begin{figure}[t]
	\centering
	\vspace{1mm}
	\hspace{-4mm}
	\setlength{\abovecaptionskip}{0mm}
	\setlength{\belowcaptionskip}{0mm}
	\includegraphics[width=0.48\textwidth]{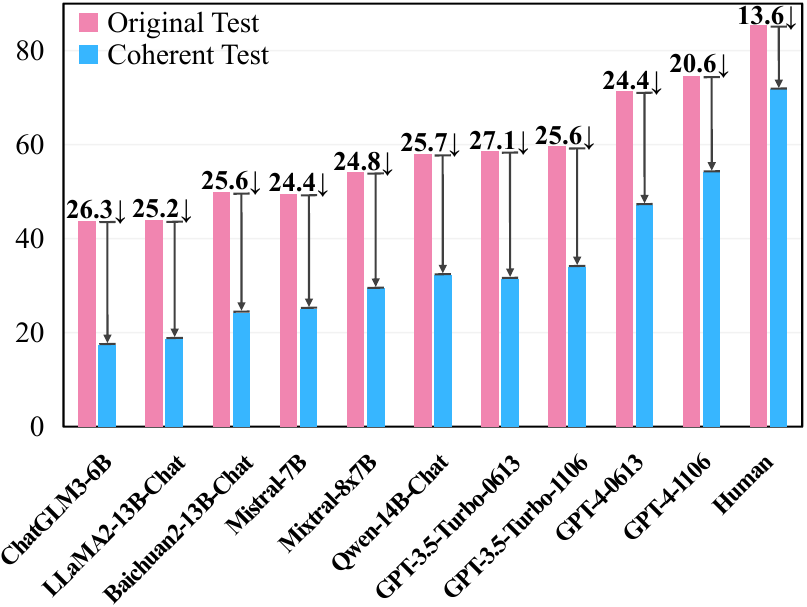}
	\centering
	\caption{
		The performance variance under the {coherent test}. Full results are present in  Appendix \ref{appendix-coherent-results}, Table \ref{tab-task-coherent-results}.}
\label{fig-coherent-results}
\vspace{-4mm}
\end{figure}

We present the average performance variation of all LLMs in Figure \ref{fig-coherent-results}. Under the coherent test, the ToM performance of all LLMs drops dramatically by at least 20\% points, while humans only experience a decrease of 13.6\% points. This widens the performance gap between the top-performing LLMs and humans from the original 10.1\% to 16.2\%, highlighting LLMs' limitations in fully comprehending social scenarios like humans.



\paragraph{LLMs Fail on Trivial ToM Questions}
We now delve into an ability where LLMs perform particularly poorly: \textit{Knowledge/Knowledge-Pretend Links}. It's important to note that this deliberate selection does not introduce bias against LLMs because, methodologically, failing even one ToM task implies the absence of a general ToM ability \cite{shapira2023clever}. Specifically, this ability tests whether subjects can understand that a person or object cannot imitate something they are unaware of. 
This task is intuitive and straightforward for humans (achieving 93.3\% accuracy) but proves difficult for LLMs. GPT-4-1106 and GPT-4-0613 achieve only 26.7\% and 3.3\% accuracy on English samples, respectively, which is nearly equivalent to or even worse than random guessing (full results can be found in Table \ref{tab-all-ability-results}). 

\begin{figure}[t]
	\centering
	\vspace{0mm}
	\hspace{0mm}
	\setlength{\abovecaptionskip}{2mm}
	\setlength{\belowcaptionskip}{0mm}
	\includegraphics[width=0.48\textwidth]{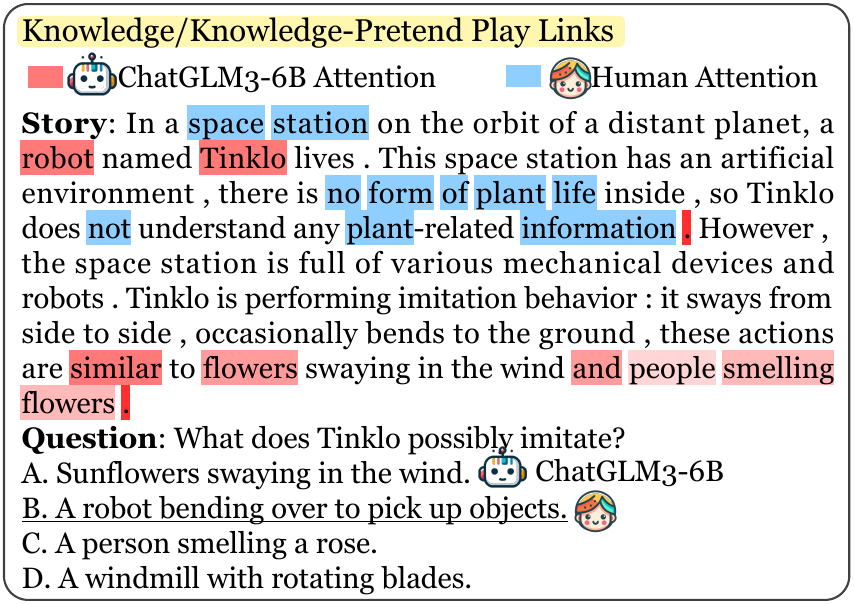}
	\centering
	\caption{
		The difference between the human and LLM's attentions. Color intensity denotes attention weights.}
	\label{fig-attention}
	\vspace{-2mm}
\end{figure}

To understand the reasons behind this poor performance, in Figure \ref{fig-attention}, we visualize the decision-making process of an open LLM, ChatGLM3-6B, of which the weights are accessible. Specifically, when ChatGLM3-6B generates an incorrect option A, we average the attention scores across the attention heads in 20 layers, identifying the top-10 keywords that ChatGLM3-6B focuses on when answering the question. Clearly, LLMs still rely on semantic associations to answer ToM questions, aligning with previous research on LLMs' illusory theory of mind \cite{kim2023fantom}.
For comparison, we also ask a student to provide the top-10 keywords. The results show that the student could correctly answer the question after reading only half of the story, further emphasizing the distinction between human and LLMs in ToM capabilities.

\section{Related Work}

As LLMs have reached and sometimes even surpassed human performance in task solving, researchers have started to explore whether LLMs possess a ToM and exhibit human-like social intelligence. \citet{Sap2022NeuralTO} tested GPT-3 with SocialIQA \cite{sap2019socialiqa} and ToMi \cite{Le2019RevisitingTE}, finding its performance significantly lower than humans. However, when it comes to instruction-tuned davince-002/003, \citet{kosinski2023theory} reported their performance as comparable to children aged 7 and 10, respectively, according to the experiments on 40 samples testing false beliefs. \citet{bubeck2023sparks} discovered that GPT-4 achieved impressive results on the Sally-Anne test with 10 samples. \citet{van2023theory} found that GPT-3.5/4 outperformed students aged 7 to 10 on various early inventories. This sparked some discussions about the ``\textit{theory of mind may have spontaneously emerged in large language models}.'' 

However, opposing views emerged quickly. \citet{ullman2023large} and \citet{shapira2023clever} showed that trivial alterations to test samples could drastically decrease LLM's ToM performance, highlighting their reliance on spurious correlations. \citet{jones2023epitome} found poor performance on tasks like scalar implicature, while \citet{ma2023tomchallenges}, \citet{wu2023hi}, and \citet{kim2023fantom} observed limited abilities in reconstructed/high-order/conversational false belief tests, respectively. The debate on whether LLMs possess ToM underscores the need for a holistic benchmark. \citet{ma2023towards} made an inspiring attempt by testing about 9 ToM abilities of simulated geometric bodies in virtual 2D grids, but this setting cannot reflect human activities in real-world social scenarios. 

Different from existing ToM assessments for LLMs, our \logo includes 8 tasks and 31 abilities, comprising 2,860 multi-choice questions built from scratch and covering diverse real-world social scenarios. \logo establishes a systematic, automated, and original ToM benchmark that can continuously contribute to future research.

\vspace{3mm}

\section{Conclusions}
We introduce \logo, the first systematic ToM benchmark for LLMs with an MCQ-type inventory entirely built from scratch. 
We reveal that even the most advanced LLMs significantly lag behind human performance in ToM. Further in-depth analysis indicates that LLMs still struggle to comprehensively understand social scenarios, and tend to rely on semantic associations rather than human-like cognitive processes when addressing ToM questions. 
By proposing \logo, we aim to provide an effective and efficient platform for evaluating the ToM capabilities of LLMs and spur further research into LLMs that innately understand ToM, thereby improving how artificial intelligence can serve us in a human-friendly way.

\section*{Limitations}
We discuss the limitations of our work as follows.

\paragraph{Evaluation Protocol} In \logo, we aim to cover a broad range of well-defined ToM tasks frequently used in psychology. It is important to recognize that additional ToM tasks exist, such as the imposing memory task \cite{kinderman1998theory}, which introduces unique evaluation protocols. Additionally, \logo includes 31 abilities from the ATOMS framework that can be evaluated through textual modality. However, visual ToM tests are also classic and common in psychology, such as visual perspective taking - acknowledging that others have different visual percepts and adopting the visual perspective of another person. As multimodal LLMs like GPT-4V rapidly develop, ToM evaluations based on images and videos will also be applicable for LLMs. We advocate for future research to broaden the range of ToM assessments for LLMs, covering more tasks and abilities for a thorough evaluation.

\paragraph{Inventory Size} Due to the difficulty and cost of manual construction, although \logo contains a total of 2,860 samples, the minimum number for some abilities is only 20. This may lead to inadequate testing of specific abilities. There is an opportunity for future work to expand the sample size, allowing for a more detailed and accurate analysis of each ability.

\paragraph{Inventory Languages} \logo currently supports Chinese and English evaluation, where the Chinese data is originally created and the English data is obtained via translation. Although we have carefully designed the prompt for GPT-4 to complete the translation and manually reviewed all data, some cultural differences may still exist, e.g. fictitious person names, fictitious location names, etc. Future efforts could focus on localizing evaluation content and expanding language coverage to better reflect diverse cultural contexts.

\paragraph{LLMs and Prompting Methods} In the experiments, we evaluate 10 popular representative LLMs, though additional LLMs could be included in future analyses. We use vanilla and CoT prompting methods for evaluation, while the recently emerging prompting methods such as ``perspective taking'' \cite{wilf2023think} and ``foresee and reflect'' \cite{zhou2023far} could also be explored to enhance the LLMs' ToM performance.

\paragraph{ToM Understanding vs. Application} Our examination of LLMs' ToM capabilities in \logo is conducted by posing ToM-related questions and evaluating the responses. It's crucial to note that an LLM's intrinsic awareness of ToM principles does not necessarily translate to effective extrinsic ToM performance in human-AI interactions. These aspects represent progressive steps—from understanding to application. While \logo aims to address the foundational aspect, future research could explore more complex interactions, e.g., considering LLMs as autonomous social agents and assessing their dynamic ToM abilities in customized social scenarios.

\section*{Ethical Considerations}
We here elaborate on the potential ethical issues.

\paragraph{ToM and Anthropomorphism} 
The theory of mind is a unique social cognitive ability inherent to humans. Evaluating LLMs' ToM capabilities through \logo might lead to anthropomorphic interpretations, where LLMs are ascribed to human-like mental states. However, it's crucial to clarify that our intention is not to humanize LLMs. Instead, we aim to enhance LLMs' ability to understand and interpret human mental states and cognitive processes, thereby improving human-machine interaction. 
This endeavor is about bridging the communication gap, not erasing the fundamental differences between humans and machines.

\paragraph{Content Safety} We authors have undergone professional training in psychology to understand ToM and reviewed broad psychological literature related to the tasks and abilities. We ensure that all test samples in \logo are free from unsafe, toxic, biased, and offensive content. All social scenarios in test stories are fictitious and do not involve any specific countries, locations, events, characters, or other identifiable information. 

\paragraph{Human Baseline} To establish the human baseline for \logo, we recruit graduate students who are native speakers of Chinese. This decision is made based on objective conditions available to us and does not reflect any intention of bias or unfairness towards individuals of any race, nationality, or other personal characteristics. Participants are selected based on the order of registration and get paid timely after the experiment. Participants complete experiments in an online document without providing any sensitive or personal information.

\section*{Acknowledgments}
This work was supported by the National Science Foundation for Distinguished Young Scholars (No. 62125604) and the NSFC Key Project (No. 61936010). This work was also supported by Tsinghua Precision Medicine Foundation, Tsinghua University - Beijing Tsingshang Architectural Decoration Engineering Co., Ltd. Joint Institute for Smart Scene Innovation Design, and China Postdoctoral Science Foundation (No. 2023M741944).



\bibliography{anthology,custom}
\bibliographystyle{acl_natbib}

\appendix

\section{Inventory Contamination in LLMs}
\label{appendix-contamination}
We use the Sally-Anne Test, a commonly applied false belief task, as a simple yet general example for observation. As shown in Table \ref{tab-contam-case}, when prompted with the Sally-Anne Test scenarios, GPT-3.5-Turbo and GPT-4 can generate test samples that closely match the original examples, highlighting the potential risk of contamination in LLMs. For more inspection methods to identify contamination, please refer to related literature on this subject \cite{golchin2023time,li2023task}.


\begin{table*}[t]
\vspace{1mm}
\small
\renewcommand\arraystretch{1.0}
\setlength{\abovecaptionskip}{1mm}
\setlength{\belowcaptionskip}{0mm}
\centering
\setlength{\tabcolsep}{1.0mm}
\begin{tabular}{|m{0.986\textwidth}|}
    \hline		
    \multicolumn{1}{|c|}{\textbf{Ground-Truth Sample in Sally-Anne Test}} \\ \hline
  Sally first placed a marble into her basket.  Then she left the scene, and the marble was transferred by Anne and hidden in her box. Then, when Sally returned, the experimenter asked the critical Belief Question: ``Where will Sally look for her marble?''. The answer is ``basket''. \\ \hline
    \multicolumn{1}{|c|}{\textbf{GPT-3.5-Turbo-0613 output with the instruction ``Give a Sally-Anne Test example and its answer.''}} \\ \hline
\textbf{Example}:\\
Sally puts a marble in her basket and leaves the room. While she is gone, Anne moves the marble from Sally’s basket to her own box. Where will Sally look for the marble when she returns? \\
\textbf{Answer}:\\
Sally will look in her own basket.\\ \hline
    \multicolumn{1}{|c|}{\textbf{GPT-4-0613 output with the instruction ``Give a Sally-Anne Test example and its answer.''}} \\ \hline
\textbf{Sally-Anne Test Example}:\\
Sally has a basket. Anne has a box. Sally puts a marble into her basket and then leaves to go for a walk. While Sally is away, Anne takes the marble from the basket and puts it into her box. Sally comes back.\\
\textbf{Question}: Where will Sally look for her marble? \\
\textbf{Answer}: Sally will look for her marble in the basket.\\ \hline
    
\end{tabular}
\caption{A case of contamination in LLMs for the frequently used Sally-Anne Test.}
\label{tab-contam-case}
\vspace{-4mm}
\end{table*}

\section{Details of Theory-of-Mind Tasks}
\label{appendix-detail-task}
In \logo, we refer to the ``Abilities in Theory of Mind Space (ATOMS)'' framework \cite{beaudoin2020systematic} and define ToM as ``\textit{a person's cognitive representation of self and others' mental states}''. Under this definition, both representing mental states (such as inferring a person's emotional reactions based on situations) and distinguishing one's own and others' mental states (such as understanding someone's false beliefs) are considered ToM tasks. Thereby, we identify 8 social cognitive tasks from a broad view of psychology literature for ToM Evaluation.
\paragraph{Unexpected Outcome Test} \cite{dyck2001autism} evaluates participants' ability to infer characters' mental states in scenarios with apparent discrepancies between emotion-eliciting contexts and the emotions actually elicited. Example data is present in Table \ref{tab-example-uot}.

\paragraph{Scalar Implicature Task} \cite{goodman2013knowledge} involves scenarios where a speaker uses terms like ``some'' to imply ``not all'' in situations of incomplete information. It tests participants' ability to infer meanings beyond literal expressions in conversational contexts. Example data is present in Table \ref{tab-example-sit}.

\paragraph{Persuasion Story Task} \cite{kolodziejczyk2016young} assesses participants' ability to understand and choose effective persuasion strategies, reflecting their understanding of how to influence others' mental states and attitudes. Example data is present in Table \ref{tab-example-pst}.

\paragraph{False Belief Task} \cite{wimmer1983beliefs} examines whether participants can distinguish between their own beliefs (true beliefs) and others' beliefs (false beliefs) when they differ. Example data is present in Table \ref{tab-example-fbt}.

\paragraph{Ambiguous Story Task} \cite{bosacki1999theory} presents ambiguous social vignettes, followed by questions that gauge participants' understanding of others' mental states, like emotions and beliefs, in uncertain situations. Example data is present in Table \ref{tab-example-ast}.

\paragraph{Hinting Test} \cite{corcoran1995schizophrenia} assesses participants' ability to infer mental states from indirect hints in social interactions, reflecting their understanding of implied meanings beyond literal statements. Example data is present in Table \ref{tab-example-ht}.

\paragraph{Strange Story Task} \cite{happe1994advanced} requires participants to infer characters' mental states in stories that include complex social communications such as lies, white lies, misunderstandings, persuasion, irony, contrary emotions, jokes, double bluffs, pretense, figures of speech, forgetting, and appearance versus reality. Example data is present in Table \ref{tab-example-sst1} (part1) and Table \ref{tab-example-sst2} (part2).

\paragraph{Faux-pas Recognition Test} \cite{baron1999recognition} tests participants' ability to recognize when characters in social stories commit a faux pas, reflecting their understanding of social norms and others' perspectives. Example data is present in Table \ref{tab-example-frt}.

\section{Details of Theory-of-Mind Abilities}
In \logo, we further identify 31 social cognitive abilities in 6 dimensions according to the ATOMS \cite{beaudoin2020systematic} framework in psychology for ToM Evaluation.  
Among them, 19 abilities are included in previous tasks, and example data can be found from the corresponding tasks. Additionally, we have constructed extra test data for the rest 12 abilities, and example data can be found in Table \ref{tab-example-emotion} (Emotion), Table \ref{tab-example-desire} (Desire), Table \ref{tab-example-intention} (Intention), and Table \ref{tab-example-knowledge} (Knowledge).

\textbf{Emotion} involves the abilities to understand that situational factors influence people's emotional states, that people can experience complex emotions, and that people can regulate emotional expressions. 
This dimension encompasses 7 abilities.

(I) Typical emotional reactions \cite{knafo2009empathy}: Inferring a person’s emotional reactions based on situations that typically elicit certain emotions/inferring a preceding event based on a person’s emotional reaction.

 (II) Atypical emotional reactions \cite{denham1986social}: Inferring or explaining a person’s emotional reactions based on situations eliciting emotions that are atypical compared to what is usually expected.

 (III) Discrepant emotions \cite{borke1971interpersonal}: Understanding that people may have discrepant feelings about an event.

 (IV) Mixed emotions \cite{gordis1989young}: Understanding that people may feel mixed emotions or different emotions successively.

 (V) Hidden emotions \cite{harris1986children}: Understanding that other people may hide their emotions.

 (VI) Moral emotions \cite{pons2000test}: Understanding that negative feelings might arise following a reprehensible action.

 (VII) Emotion regulation \cite{pons2000test}: Understanding that others might use strategies to regulate their emotions.

\textbf{Desire} involves the abilities to understand that people have subjective desires, preferences, and wants that influence their emotions and actions. 
This dimension encompasses 4 abilities.

(I) Discrepant desires \cite{repacholi1997early}: Understanding that different people may have discrepant desires.

(II) Multiple desires \cite{bennett1993children}: Understanding the co-existence of multiple desires simultaneously or successively in one person.

(III) Desires influence on emotions and actions \cite{wellman1988young}: Understanding that people’s emotions and actions are influenced by their desires/preferences. 

(IV) Desire-action contradiction \cite{colonnesi2008precursors}: Producing plausible explanations when actions contradict stated desires/preferences.

\textbf{Intention} involves the abilities to understand that people undertake actions in pursuit of goals and intentions. 
This dimension encompasses 4 abilities. 

(I) Completion of failed actions \cite{meltzoff1995understanding}: Understanding another person’s intent, as demonstrated by completing their failed action.

(II) Discrepant intentions \cite{killen2011accidental}: Understanding that identical actions/results can be achieved with different intentions.

(III) Prediction of actions \cite{phillips2002infants}: Predicting people’s actions based on their intentions.

(IV) Intentions explanations \cite{smiley2001intention}: Producing plausible intention explanations for different types of observed social events.

\textbf{Knowledge} involves the abilities to understand that others have access to different knowledge based on their perceptions, information they have received, or familiarity with things. 
This dimension encompasses 4 abilities. 

(I) Knowledge-pretend play links \cite{aronson1999preschoolers}: Understanding that someone who does not know something exists cannot engage in ``pretend play'' that incorporates that knowledge.

(II) Percepts-knowledge links \cite{pillow1989early}: Understanding that someone who does not have access to perceptual information (i.e., by looking, hearing, etc.) may not have access to knowledge.

(III) Information-knowledge links \cite{peskin2014keeping}: Understanding that someone who was not informed or is not familiar with something may not know.

(IV) Knowledge-attention links \cite{moll2006infants}: Understanding that something new is more interesting to someone than something already known.

\textbf{Belief} involves the abilities to understand that people can hold beliefs about the world that are different from reality or different from one's own beliefs. 
This dimension encompasses 6 abilities

(I) Content false beliefs \cite{hogrefe1986ignorance}: Familiar container with an unexpected content: Understanding the false belief held by someone who never opened the container.

(II) Location false beliefs \cite{wimmer1983beliefs}: Understanding the false belief held by someone who did not witness or was not informed of a displacement or change of action.

(III) Identity false beliefs \cite{flavell1986development}: Understanding that when something looks/sounds/smells like something else, a person may hold a false belief about its identity.

(IV) Second-order beliefs \cite{perner1985john}: Understanding the second-order belief or false belief held by someone who doe not know somebody else was informed (e.g., of a misleading identity, a misleading location, etc.).

(V) Beliefs based action/emotions \cite{swettenham1996can}: Predicting another emotions or actions based on their stated beliefs/Inferring another person’s belief based on their stated action or emotion.

(VI) Sequence false beliefs \cite{brambring2010validity}: Understanding the false belief created when a predictable sequence of stimuli is broken with the intrusion of an unexpected stimulus.

\textbf{Non-literal Communication} involves the abilities to understand that communication can convey meaning beyond the literal words spoken. 
This dimension encompasses 6 abilities.

(I) Irony/sarcasm \cite{sullivan1995children}: Understanding that other people may lie in order to be ironic/sarcastic.

(II) Egocentric lies \cite{happe1994advanced}: Understanding that someone may consciously lie in order to avoid a problem or to get its way.

(III) White lies \cite{happe1994advanced}: Understanding that someone may lie in order to spare another’s feelings.

(IV) Involuntary lies \cite{happe1994advanced}: Understanding that someone may tell a ``lie'' without knowing.

(V) Humor \cite{happe1994advanced}: Understanding that someone may tell a ``lie'' in order to make a joke.

(VI) Faux pas \cite{baron1999recognition}: Ability to recognize faux-pas (social gaffe) situations.

\label{appendix-detail-ability}
\begin{table*}[t]
\vspace{0mm}
\small
\renewcommand\arraystretch{1.4}
\setlength{\abovecaptionskip}{1mm}
\setlength{\belowcaptionskip}{0mm}
\centering
\setlength{\tabcolsep}{1.0mm}
\begin{tabular}{|m{0.986\textwidth}|}
\hline

\textbf{Story}: Xiao Ming receives a bicycle on his birthday.  \\ 
\colorbox{bestc!20}{Ability: Emotion/Typical emotional reactions} \\
\textbf{Question-1}: What is Xiao Ming's emotion?     \\
(A) Embarrassed \textbf{(B) Happy} (C) Disappointed (D) Regretful \\ 
\colorbox{bestc!20}{Ability: Belief/Sequence false beliefs} \\
\textbf{Question-2}: He should be very happy, but he is very disappointed, why?  \\
(A) Xiao Ming worries that riding a bicycle affects his studies.	(B) Xiao Ming fears that riding a bicycle to school makes his classmates laugh at him.	(C) Xiao Ming thinks the color of the bicycle does not match his clothes.	\textbf{(D) Xiao Ming hopes for a computer as a gift, not a bicycle.} \\
\colorbox{bestc!20}{Ability: Emotion/Atypical emotional reactions} \\
\textbf{Question-3}: Xiao Ming is having a birthday, he hopes for a computer or a new game as a birthday gift, on his birthday he receives a bicycle. What is Xiao Ming's emotion at this time?  \\ 
(A) Embarrassed (B) Happy \textbf{(C) Disappointed} (D) Regretful \\ \hline

\end{tabular}
\caption{Example data of the {{\colorbox{bestd!20}{Unexpected Outcome Test (\textbb{UOT})}}}.}
\label{tab-example-uot}
\vspace{0mm}
\end{table*}

\begin{table*}[t]
	\vspace{0mm}
	\small
	\renewcommand\arraystretch{1.4}
	\setlength{\abovecaptionskip}{1mm}
	\setlength{\belowcaptionskip}{0mm}
	\centering
	\setlength{\tabcolsep}{1.0mm}
	\begin{tabular}{|m{0.986\textwidth}|}
		\hline		
		\textbf{Story}: Almost every letter to Laura Company contains a check. Today, Laura receives 5 letters. Laura tells you on the phone ``I look at 3 out of 5 letters. There are checks in 2 of the letters.'' \\ 
		\colorbox{bestc!20}{Ability: Knowledge/Information-knowledge links (Question-1, 2)} \\
		\textbf{Question-1}: Before Laura calls you, how many of these 5 letters do you think contain checks?     \\
		(A) 0 (B) 1 (C) 2 \textbf{(D) 4} \\ 
		\textbf{Question-2}: After Laura calls you, how many of these 5 letters do you think contain checks?  \\
		(A) 0	(B) 1	(C) 2	\textbf{(D) 4} \\\hline
		
	\end{tabular}
\caption{Example data of the \colorbox{bestd!20}{Scalar Implicature Task (\textbb{SIT})} .}
	\label{tab-example-sit}
	\vspace{0mm}
\end{table*}

\begin{table*}[t]
	\vspace{0mm}
	\small
	\renewcommand\arraystretch{1.4}
	\setlength{\abovecaptionskip}{1mm}
	\setlength{\belowcaptionskip}{0mm}
	\centering
	\setlength{\tabcolsep}{1.0mm}
	\begin{tabular}{|m{0.986\textwidth}|}
		\hline		
		\textbf{Story}: Xiao Hong is a 6-year-old child. Today is Saturday. Mom and Dad have free time today, they do not know what they can do together. Maybe go for ice cream? Xiao Hong really, really wants to go to the amusement park today. However, Dad thinks the amusement park has a lot of noise. He says: ``Xiao Hong, this is not a good idea. I think there is a lot of noise in the amusement park.'' \\
		\colorbox{bestc!20}{Ability: Desire/Desires influence on actions/emotions} \\
		\textbf{Question}: How does Xiao Hong persuade her dad?     \\
		(A) Xiao Hong can look up some information, proving that the current amusement parks take many measures to reduce noise, such as setting up soundproof walls, using quieter equipment, etc. \textbf{(B) Xiao Hong can tell her dad that she has not been to the amusement park for a long time, this is a very special wish for her, she really wants to go.} (C) If Dad really does not want to go to the amusement park, Xiao Hong can suggest going to other places, such as parks or zoos, so everyone is happy. (D) Xiao Hong can suggest: ``How about we go in the early morning or on weekdays? I hear that there are fewer people at that time, and it is quieter. We can avoid peak hours, and get a quieter experience.'' \\ \hline
		
	\end{tabular}
	\caption{Example data of the \colorbox{bestd!20}{Persuasion Story Task (\textbb{PST})}.}
	\label{tab-example-pst}
	\vspace{0mm}
\end{table*}

\begin{table*}[t]
	\vspace{0mm}
	\small
	\renewcommand\arraystretch{1.4}
	\setlength{\abovecaptionskip}{1mm}
	\setlength{\belowcaptionskip}{0mm}
	\centering
	\setlength{\tabcolsep}{1.0mm}
	\begin{tabular}{|m{0.986\textwidth}|}
		\hline		
		\textbf{Story-a}: Li Lei and Han Meimei are wandering in the living room, they see the cabinet, box and handbag, they find a sweatshirt in the cabinet, Han Meimei leaves the living room, Li Lei moves the sweatshirt to the handbag. \\ 
		\colorbox{bestc!20}{Ability: Belief/Location false beliefs} \\
		\textbf{Question-a1}: Where is the sweatshirt now?     \\
		(A) Briefcase (B) Box (C) Cabinet \textbf{(D) Handbag} \\ 
		\colorbox{bestc!20}{Ability: Belief/Second-order beliefs} \\
		\textbf{Question-a2}: After Han Meimei returns to the living room, where does Li Lei think Han Meimei looks for the sweatshirt?     \\
		(A) Box (B) Wardrobe (C) Handbag \textbf{(D) Cabinet} \\ \hline 
		\textbf{Story-b}: Xiao Li finds a storage cabinet in the study, the label on the storage cabinet is carrot, Xiao Li cannot see what is inside the storage cabinet, Xiao Li opens the storage cabinet and finds a coat, there are no carrots in the storage cabinet, Xiao Li closes the storage cabinet and puts it back in its place, Xiao Zhang enters the study and sees the storage cabinet. \\ 
		\colorbox{bestc!20}{Ability: Belief/Content false beliefs} \\
		\textbf{Question-b1}: What is in the storage cabinet?     \\
		\textbf{(A) Coat} (B) Pencil (C) Carrot (D) Tape \\ 
		\colorbox{bestc!20}{Ability: Belief/Second-order beliefs} \\
		\textbf{Question-b2}: After Xiao Zhang opens the storage cabinet, what does Xiao Zhang think Xiao Li expects to find in the storage cabinet?     \\
		(A) Mask (B) Marker (C) Coat \textbf{(D) Carrot} \\ \hline 
		
	\end{tabular}
	\caption{Example data of the \colorbox{bestd!20}{False Belief Task (\textbb{FBT})} .}
	\label{tab-example-fbt}
	\vspace{0mm}
\end{table*}

\begin{table*}[t]
	\vspace{0mm}
	\small
	\renewcommand\arraystretch{1.4}
	\setlength{\abovecaptionskip}{1mm}
	\setlength{\belowcaptionskip}{0mm}
	\centering
	\setlength{\tabcolsep}{1.0mm}
	\begin{tabular}{|m{0.986\textwidth}|}
		\hline		
		\textbf{Story}: Dehua and Sanming are company employees, they are competing for a promotion opportunity. Today is Sanming's birthday, he enjoys a small birthday cake alone. Lingling is the company supervisor, she is having a private conversation with Dehua in the pantry. Sanming sees from a distance that Lingling quietly gives Dehua a document, and with a smile, gently pats Dehua's back, then returns to her office. Dehua quickly checks the document, shows a smile, and carefully puts the document into his file bag. \\ 
		\colorbox{bestc!20}{Ability: Intention/Intentions explanations} \\
		\textbf{Question-1}: Why does Lingling give Dehua that document?     \\
		(A) Lingling returns Dehua's personal document left in the conference room. (B) Lingling gives Dehua a daily notice unrelated to promotion. (C) Lingling and Dehua are planning a surprise for Sanming. \textbf{(D) Lingling gives Dehua extra information.} \\ 
		\colorbox{bestc!20}{Ability: Belief/Beliefs based action/emotions} \\
		\textbf{Question-2}: What do you think Sanming thinks?     \\
		(A) Sanming thinks everything is normal because that is Dehua's document. (B) Sanming feels angry because he is isolated. (C) Sanming feels very happy because his cake tastes good. \textbf{(D) Sanming reevaluates his promotion opportunity.} \\ \hline 	
		
	\end{tabular}
	\caption{Example data of the \colorbox{bestd!20}{Ambiguous Story Task (\textbb{AST})}.}
	\label{tab-example-ast}
	\vspace{0mm}
\end{table*}

\begin{table*}[t]
\vspace{0mm}
\small
\renewcommand\arraystretch{1.4}
\setlength{\abovecaptionskip}{1mm}
\setlength{\belowcaptionskip}{0mm}
\centering
\setlength{\tabcolsep}{1.0mm}
\begin{tabular}{|m{0.986\textwidth}|}
	\hline		
	\textbf{Story}: Cao Sheng and Wang Hong are a couple. One day, Wang Hong makes dinner, Cao Sheng takes a bite and says, ``Does our home not buy salt for a long time?'' \\ 
	\colorbox{bestc!20}{Ability: Intention/Intentions explanations} \\
	\textbf{Question}: What does Cao Sheng really want to say when he says this?     \\
	(A) Cao Sheng is asking if the salt in the house is used up. (B) Cao Sheng is expressing his increased need for salt. \textbf{(C) Cao Sheng implies that the dinner Wang Hong makes does not have enough salt.} (D) Cao Sheng is reminding them that they need to buy more ingredients. \\ \hline
	
\end{tabular}
\caption{Example data of the \colorbox{bestd!20}{Hinting Test (\textbb{HT})} .}
\label{tab-example-ht}
\vspace{0mm}
\end{table*}

\begin{table*}[t]
	\vspace{0mm}
	\small
	\renewcommand\arraystretch{1.4}
	\setlength{\abovecaptionskip}{1mm}
	\setlength{\belowcaptionskip}{0mm}
	\centering
	\setlength{\tabcolsep}{1.0mm}
	\begin{tabular}{|m{0.986\textwidth}|}
		\hline		
		\textbf{Story-a}: Li Tong and Wang Hong are best friends. They both participate in the same painting competition. Now, Li Tong really wants to win this competition, but when the competition results come out, the winner of the competition is her best friend Wang Hong, not her. Li Tong is very upset that she does not win, but she feels happy for her friend who wins. Li Tong says to Wang Hong: ``Good job, I am really happy you win!'' Li Tong says to her mom: ``I am very sad, I do not win the competition!'' Is what Li Tong says to Wang Hong true? \\
		\colorbox{bestc!20}{Ability: Emotion/Mixed emotions (Question-a1, a2)} \\
		\textbf{Question-a1}: Is what Li Tong says to Wang Hong true?     \\
		\textbf{(A) Yes} (B) No \\
		\textbf{Question-a2}: Why does Li Tong say she is both happy and sad?     \\
		(A) She is happy that Wang Hong wins the competition, she is sad and very jealous that Wang Hong wins the competition. (B) She is happy that Wang Hong wins the competition and is not sad about it. (C) She is happy that Wang Hong wins the competition but is sad that Wang Hong does not comfort her for not winning the competition. \textbf{(D) She is happy that Wang Hong wins the competition but is sad that she does not win the competition.} \\ \hline 
		
		\textbf{Story-b}: Xiao Ming's teacher often gives students a lot of homework. This weekend, Xiao Ming wants to go to his friend Xiao Chen's house to play games together. But Xiao Ming still has a lot of homework that he does not finish. Xiao Ming's father sees him working hard all day on his homework, and asks Xiao Ming if he finishes his homework? Does he want to go out and play for a rest? Xiao Ming says, ``No, I do not want to.'' \\
		\colorbox{bestc!20}{Ability: Desire/Desire-action contradiction (Question-b1, b2)} \\
		\textbf{Question-b1}: Is what Xiao Ming says to his father true?     \\
		(A) Yes \textbf{(B) No} \\
		\textbf{Question-b2}: Why does Xiao Ming say he does not want to go out and play?     \\
		(A) Because he thinks the homework is too hard, so he lies and says he does not want to go. \textbf{(B) Because he finds it hard to finish the homework.} (C) Because he forgets the homework. (D) Because he does not want his father to think he is a child who loves to play. \\ \hline 
		
		\textbf{Story-c}: At a birthday party, Wen Qing pretends to be a superhero and says, ``I am a wise and powerful superhero!''. Her friends all watch happily. \\
		\colorbox{bestc!20}{Ability: Desire/Desire-action contradiction (Question-c1, c2)} \\
		\textbf{Question-c1}: Is what Wen Qing says true?     \\
		(A) Yes \textbf{(B) No} \\
		\textbf{Question-c2}: Why does Wen Qing say this?     \\
		(A) Wen Qing misunderstands that she is a wise and powerful superhero. (B) Wen Qing lies because she wants to make her friends happy. (C) Wen Qing says this because she is a real superhero. \textbf{(D) Wen Qing plays the role of a superhero because she wants to make her friends happy.} \\ \hline 
		
		\textbf{Story-d}: Anming's mother spends a long time making Anming's favorite fried fish and chips. But when she brings the food to Anming, Anming watches TV, she doesn't even look up, and doesn't say thank you. Anming's mother says angrily: ``You are really polite!'' \\
		\colorbox{bestc!20}{Ability: Non-Literal Communication/Irony or Sarcasm (Question-d1, d2)} \\
		\textbf{Question-d1}: Does mother tell the truth?     \\
		(A) Yes \textbf{(B) No} \\
		\textbf{Question-d2}: Why does the mother say this?     \\
		(A) She thinks Anming is a very polite child. (B) She lies to encourage Anming to be a very polite child. (C) She wants to mock her own food is not delicious. \textbf{(D) She wants to mock Anming is a child without manners.} \\ \hline 
		
	\end{tabular}
	\caption{Example data of the \colorbox{bestd!20}{Strange Story Task (\textbb{SST}) Part1 }.}
	\label{tab-example-sst1}
	\vspace{0mm}
\end{table*}

\begin{table*}[t]
	\vspace{0mm}
	\small
	\renewcommand\arraystretch{1.1}
	\setlength{\abovecaptionskip}{1mm}
	\setlength{\belowcaptionskip}{0mm}
	\centering
	\setlength{\tabcolsep}{1.0mm}
	\begin{tabular}{|m{0.986\textwidth}|}
		\hline		
		\textbf{Story-a}: One day, Lily plays at home and accidentally knocks over and breaks her mother's favorite crystal vase. Oh no, her mother definitely gets angry when she finds out! Therefore, when Lily's mother comes home, sees the broken vase and asks Lily what happens, Lily says, ``The dog knocks it over, it is not my fault!'' \\
		\colorbox{bestc!20}{Ability: Non-Literal Communication/Egocentric lies (Question-a1, a2)} \\
		\textbf{Question-a1}: Is what Lily says true?     \\
		(A) Yes \textbf{(B) No} \\
		\textbf{Question-a2}: Why does Lily say this?     \\
		(A) Lily sees the dog knock over the vase and wants to protect the dog from her mother's blame. (B) Lily jokes to protect herself from her mother's blame. (C) Lily lies to protect the dog from her mother's blame. \textbf{(D) Lily lies to protect herself from her mother's blame.} \\ \hline 
		
		\textbf{Story-b}: Mr. Zhao invites his business partner, Manager Wang, to a high-end restaurant for dinner. At the dinner table, Mr. Zhao notices that Manager Wang shows great interest in a dish, but in fact, this dish is a bit salty. Then, Mr. Zhao immediately praises that the taste of this dish is just right. \\
		\colorbox{bestc!20}{Ability: Non-Literal Communication/White lies (Question-b1, b2)} \\
		\textbf{Question-b1}: Does Mr. Zhao tell the truth?     \\
		(A) Yes \textbf{(B) No} \\
		\textbf{Question-b2}: Why does Mr. Zhao say this?     \\
		(A) Mr. Zhao thinks this dish is really delicious, so he tells the truth. (B) Mr. Zhao wants to please Manager Wang in order to gain an advantageous position in future business cooperation. (C) Mr. Zhao is a regular customer of this restaurant, out of respect for the restaurant, he always praises every dish. \textbf{(D) Mr. Zhao does this out of politeness and social customs, in order to avoid embarrassment and maintain a harmonious dining atmosphere, so he chooses to praise this dish.} \\ \hline 
		
		\textbf{Story-c}: Little Chen promises his grandma to water her flowers. He goes to his friend's house to play, rushes out of the house, and forgets about this. He thinks he waters the flowers. Grandma asks, ``Do you water my flowers?'' Little Chen says, ``I water them.'' \\
		\colorbox{bestc!20}{Ability: Non-Literal Communication/Involuntary lies (Question-c1, c2)} \\
		\textbf{Question-c1}: Is what Little Chen says true?     \\
		(A) Yes \textbf{(B) No} \\
		\textbf{Question-c2}: Why does Little Chen say this?     \\
		(A) Little Chen intentionally lies because he does not want his grandma to think he does not do well. \textbf{(B) Little Chen forgets that he does not water the flowers because he is too happy playing at his friend's house.} (C) Little Chen intentionally lies because he wants to go play. (D) Little Chen thinks he waters the flowers because he usually is responsible. \\ \hline 
		
		\textbf{Story-d}: In a new company, Xiao Zhang just joins and is not very familiar with his colleagues. At lunchtime, he notices a colleague sitting alone, so he takes the initiative to go and sit down, laughing and saying, "Do you know? When I come today, I see a cat wearing a tie at the door, it must be the CEO of the company!" The colleague laughs uncontrollably. \\
		\colorbox{bestc!20}{Ability: Non-Literal Communication/Humor (Question-d1, d2)} \\
		\textbf{Question-d1}: Is what Xiao Zhang says true?     \\
		(A) Yes \textbf{(B) No} \\
		\textbf{Question-d2}: Why does Xiao Zhang say this?     \\
		\textbf{(A) Xiao Zhang says this because he wants to close the distance with his colleagues and enhance their friendship and interaction.} (B) Xiao Zhang says this because he really sees a cat wearing a tie and finds it very interesting and wants to share with his colleagues. (C) Xiao Zhang says this because he wants to show off his observation skills and make his colleagues like him. (D) Xiao Zhang says this because he thinks the company's CEO should be a cat, expressing dissatisfaction with the company's management. \\ \hline 
		
	\end{tabular}
	\caption{Example data of the \colorbox{bestd!20}{Strange Story Task (\textbb{SST}) Part2} .}
	\label{tab-example-sst2}
	\vspace{0mm}
\end{table*}

\begin{table*}[t]
	\vspace{0mm}
	\small
	\renewcommand\arraystretch{1.1}
	\setlength{\abovecaptionskip}{1mm}
	\setlength{\belowcaptionskip}{0mm}
	\centering
	\setlength{\tabcolsep}{1.0mm}
	\begin{tabular}{|m{0.986\textwidth}|}
		\hline		
		\textbf{Story}: Last Friday, all classmates in the fifth grade of elementary school participate in a Chinese composition competition. Xiao Ming really wants to win first place in this competition. The competition results come out, Xiao Hong wins the competition, and Xiao Ming happens to take a leave of absence and does not go to school that day. When going to school on Monday, Xiao Hong is the first to see Xiao Ming and says: ``Xiao Ming, I read your composition and it is not bad, just a bit regrettable.'' Xiao Ming is very surprised when he hears this and says: "What do you mean, can you be more clear?" Xiao Hong looks at Xiao Ming and says: ``Oh, it is nothing, let's go to class.'' \\ 
		\colorbox{bestc!20}{Ability: Non-Literal Communication/Faux pas (Question 1, 2, 3, 4)} \\
		\textbf{Question-1}: Does anyone say something inappropriate in this story?     \\
		\textbf{(A) Someone in the story says something inappropriate.} (B) No one in the story says something inappropriate. \\ 
		\textbf{Question-2}: If there is one, which sentence is inappropriate?     \\
		\textbf{(A) Xiao Ming, I read your composition and it is not bad, it is just a bit regrettable.} (B) What do you mean, can you explain it more clearly? (C) Oh, it is nothing, let's go to class. (D) No one in the story says anything inappropriate. \\  	
		
		\textbf{Question-3}: Who wins the composition competition?     \\
		(A) Xiao Ming (B) Xiao Long \textbf{(C) Xiao Hong} (D) The story does not mention \\  	
		
		\textbf{Question-4}: Does Xiao Hong know that Xiao Ming really wants to win this composition competition?     \\
		(A) Knows \textbf{(B) Does not know} \\ \hline 	
		
	\end{tabular}
	\caption{Example data of the \colorbox{bestd!20}{Faux-Pas Recognition Test (\textbb{FRT})}.}
	\label{tab-example-frt}
	\vspace{0mm}
\end{table*}

\begin{table*}[t]
	\vspace{0mm}
	\small
	\renewcommand\arraystretch{1.1}
	\setlength{\abovecaptionskip}{1mm}
	\setlength{\belowcaptionskip}{0mm}
	\centering
	\setlength{\tabcolsep}{1.0mm}
	\begin{tabular}{|m{0.986\textwidth}|}
		\hline		
		\textbf{Story-a}: Xiao Hong is supposed to help her club prepare for this event, but she goes to visit a friend instead. \\ 
		\colorbox{bestc!20}{Ability: Emotion/Discrepant emotions (Question-a1, a2)} \\
		\textbf{Question-a1}: What kind of emotion does Xiao Hong's friend have?     \\
		(A) Angry (B) Proud \textbf{(C) Grateful} (D) Regretful\\ 
		\textbf{Question-a2}: What kind of emotion does the club member have?     \\
		\textbf{(A) Angry} (B) Proud (C) Grateful (D) Regretful\\ \hline		
			
		\textbf{Story-b}: Xinxin wants to go to her friend's party tonight, but she has a stomachache. She knows, if she tells her mother she has a stomachache, her mother does not let her go. She tries to hide her feelings, so her mother lets her go to the party. \\ 
		\colorbox{bestc!20}{Ability: Emotion/Hidden emotions (Question-b1, b2)} \\
		\textbf{Question-b1}: What are Xinxin's real feelings?     \\
		\textbf{(A) Sad} (B) Scared (C) Happy (D) Disappointed\\ 
		\textbf{Question-b2}: Why is Xinxin's true feeling sad?     \\
		(A) Because she does not like parties. \textbf{(B) Because of the stomachache.} (C) She cannot attend the party. (D) Because she fears her mother worries. \\  \hline	
		
		\textbf{Story-c}:James shows his painting at the school art exhibition. One day, he forgets to take the painting home. Nick, while cleaning the exhibition room, mistakes the painting for abandoned and puts it in the discard box. Meanwhile, Ella knows this is James's work. Ella's own painting also shows at the art exhibition, and she does not want James's painting to be more popular than hers, so she decides not to tell Nick. When James finds the painting is missing, he feels very upset and starts asking classmates. Later, Nick realizes he may have mishandled James's painting. \\ 
		\colorbox{bestc!20}{Ability: Emotion/Moral emotions (Question-c1, c2)} \\
		\textbf{Question-c1}: What kind of emotion does Nick feel after realizing he may have mishandled James's painting?     \\
		(A) Nick possibly feels satisfied because he completes the task of cleaning the exhibition room, helping to clear useless items. (B) Nick possibly feels indifferent, and he indeed does not know at the time, nor does he intentionally take James's. \textbf{(C) Nick possibly feels guilty and anxious because he unintentionally causes loss to others.} (D) Nick possibly feels confused because he is not sure if he really handles James's painting.\\ 
		\textbf{Question-c2}: Ella knows that is James's work, but does not tell Nick, what kind of emotion does she feel?     \\
		(A) Ella possibly feels happy because she harbors hostility towards James. (B) Ella possibly feels worried, fearing that James finds out she knows the truth but does not tell Nick. \textbf{(C) Ella possibly feels satisfied and complacent because she sees James's painting is mishandled and he cannot show his talent.} (D) Ella possibly feels indifferent because she thinks this has nothing to do with her. \\  \hline	
		
		\textbf{Story-d}: The company's project is about to end, and Zhang Hua and Li Jun work overtime until very late. Chen Yu takes a day off for personal reasons before. Zhang Hua seems to casually mention Chen Yu in the office, but there is a hint of questioning in his words. After hearing this, Li Jun looks at Chen Yu's empty seat, then frowns. After a while, Zhang Hua pretends to laugh easily, takes out a leave slip of Chen Yu from the drawer, shows it to Li Jun, and then puts it back. Li Jun seems to think about it, does not speak anymore, and continues to work. \\ 
		\colorbox{bestc!20}{Ability: Emotion/Emotion regulation} \\
		\textbf{Question-d}: What does Li Jun do next?     \\
		\textbf{(A) He reassesses the situation. He thinks about the reasonableness of Chen Yu's leave, realizes that everyone may need to rest for personal reasons, and this does not mean that they are irresponsible or not focused on work.} (B) He checks Chen Yu's social media to see if he really takes leave for personal reasons. (C) He asks other colleagues about their views on Chen Yu's leave. (D) He ignores this issue and focuses on his own work.\\  \hline

	\end{tabular}
	\caption{\texttt{4} abilities in \colorbox{bestc!20}{Emotion} that are not covered by the 8 tasks and evaluated with extra samples.}
	\label{tab-example-emotion}
	\vspace{0mm}
\end{table*}

\begin{table*}[t]
	\vspace{0mm}
	\small
	\renewcommand\arraystretch{1.1}
	\setlength{\abovecaptionskip}{1mm}
	\setlength{\belowcaptionskip}{0mm}
	\centering
	\setlength{\tabcolsep}{1.0mm}
	\begin{tabular}{|m{0.986\textwidth}|}
		\hline		
		\textbf{Story-a}: The little boy Mark always wants to eat a hamburger at lunchtime. When he walks to the school cafeteria to buy a hamburger, his good friend Lenny comes to him and says, "Mark, if you help me answer a few questions on my math homework, I give you a hamburger." So Mark decides to help Lenny with his math homework. \\ 
		\colorbox{bestc!20}{Ability: Desire/Multiple desires} \\
		\textbf{Question-a}: What does Mark do after he helps Lenny finish his math homework?     \\
		(A) Mark continues to go to the cafeteria to buy a hamburger. \textbf{(B) Mark eats the hamburger from Lenny.} (C) Mark goes to the library to read. (D) Mark goes to the playground to play.\\ \hline		
		
		\textbf{Story-b}: Li Hua is a person who is passionate about outdoor adventures, and he goes hiking or camping every weekend. He always seeks fresh and exciting experiences. Wang Fang, a quiet librarian, spends her weekends in a quiet library, reading and organizing books. She likes a calm and orderly environment. This weekend, Li Hua invites Wang Fang to go out for the weekend. \\ 
		\colorbox{bestc!20}{Ability: Desire/Discrepant desires} \\
		\textbf{Question-b}: Where does Li Hua plan to spend the weekend with Wang Fang?     \\
		(A) Hiking \textbf{(B) Library} (C) Sports Park (D) Skiing\\ \hline

	\end{tabular}
	\caption{\texttt{2} abilities in \colorbox{bestc!20}{Desire} that are not covered by previous 8 tasks and evaluated with extra samples.}
	\label{tab-example-desire}
	\vspace{0mm}
\end{table*}

\begin{table*}[t]
	\vspace{0mm}
	\small
	\renewcommand\arraystretch{1.1}
	\setlength{\abovecaptionskip}{1mm}
	\setlength{\belowcaptionskip}{0mm}
	\centering
	\setlength{\tabcolsep}{1.0mm}
	\begin{tabular}{|m{0.986\textwidth}|}
		\hline		
		\textbf{Story-a}: Emily finds a book she really wants to read in the library, but discovers that the book is already borrowed by someone else. Emily sees a book with a similar cover next to it, thinks it is the same book, and takes it away. After returning home, Emily realizes it is a different book, but is too lazy to return it. Meanwhile, Kate is busy working in the library. She knows this book belongs to a blind person who is waiting for this book. Kate, who is recently cheated by a blind person she does not know, notices that Emily takes the wrong book, but she chooses not to tell Emily. \\ 
		\colorbox{bestc!20}{Ability: Intention/Discrepant intentions (Question-a1, a2)} \\
		\textbf{Question-a1}: What is the possible intention behind Emily's behavior?     \\
		(A) Emily possibly chooses not to return it due to negligence or thinking that this book is not important. (B) Emily possibly takes it away without careful inspection because she is eager to read this book. (C) Emily thinks this book belongs to the library, so there is no need to return it. \textbf{(D) Emily possibly mistakenly thinks she finds the book she wants, only realizing she takes the wrong one after returning home.}\\ 		
		\textbf{Question-a2}: What is the possible intention behind Kate's behavior?     \\
		(A) Kate possibly does not have time to tell Emily she takes the wrong book because of her busy work. \textbf{(B) Kate has a dislike for the blind, even if she knows Emily takes the wrong book, she chooses not to tell her.} (C) Kate thinks Emily realizes her mistake and returns the book on her own, so she chooses not to tell her. (D) Kate possibly chooses not to tell Emily she takes the wrong book because of a conflict with Emily.\\ \hline	
		
		\textbf{Story-b}: Qingqing, Dapeng, and Bald Qiang are all hosts on the same channel. They often appear in the same program, but rarely have the opportunity to go out for a meal together. One day, after the program recording ends, Dapeng sees Qingqing is about to leave, he quickly goes over to talk to her. At this time, Bald Qiang also comes over. \\ 
		\colorbox{bestc!20}{Ability: Intention/Prediction of actions} \\
		\textbf{Question-b}: What does Dapeng want to do?     \\
		\textbf{(A) Invites Qingqing and Bald Qiang to have dinner together.} (B) Asks Qingqing for her phone number. (C) Asks Qingqing to share her thoughts on today's program. (D) Dapeng asks Qingqing to introduce him to Bald Qiang.\\ \hline

		\textbf{Story-c}: Xiaohua is in the kitchen preparing dinner when suddenly he hears the doorbell ring. He turns down the fire and goes to open the door, finding it is his friend Xiaoli. Xiaoli holds some movie tickets in his hand and invites Xiaohua to watch a movie. Xiaohua thinks for a moment and tells Xiaoli that he must finish dinner first. At this time, Xiaohua's sister also runs over and says she can help cook, letting Xiaohua go play. Xiaohua looks at the half-cooked dinner in the kitchen and hesitates. \\ 
		\colorbox{bestc!20}{Ability: Intention/Completion of failed actions} \\
		\textbf{Question-c}: What does Xiaohua do next after hearing Xiaoli's invitation?     \\
		(A) He continues to cook dinner. \textbf{(B) He accepts Xiaoli's invitation and goes to watch a movie.} (C) He lets his sister finish dinner and he rests. (D) He goes to discuss with other friends whether to go to the movies.\\ \hline
		
	\end{tabular}
	\caption{\texttt{3} abilities in \colorbox{bestc!20}{Intention} that are not covered by the 8 tasks and evaluated with extra samples.}
	\label{tab-example-intention}
	\vspace{0mm}
\end{table*}

\begin{table*}[t]
	\vspace{0mm}
	\small
	\renewcommand\arraystretch{1.1}
	\setlength{\abovecaptionskip}{1mm}
	\setlength{\belowcaptionskip}{0mm}
	\centering
	\setlength{\tabcolsep}{1.0mm}
	\begin{tabular}{|m{0.986\textwidth}|}
		\hline		
		\textbf{Story-a}: Tara is a curious robot, living in an underwater city called Aquatica in the deep sea. In Aquatica, there are no birds or flying creatures, and Tara never understands them. However, Aquatica is full of all kinds of marine life. Tara is doing imitation behavior: it swings its arms up and down gracefully, like a forward motion, very similar to the flapping of bird wings. \\ 
		\colorbox{bestc!20}{Ability: Knowledge/Knowledge-pretend play links} \\
		\textbf{Question-a}: What is Tara likely imitating?     \\
		\textbf{(A) Fish sliding fins.} (B) Soaring eagle. (C) Butterfly flapping wings. (D) Bat in flight.\\ \hline	
		
		\textbf{Story-b}: Xiaoli and Lilei are wandering in the hall, they see the cabinet and the box, they find vests and sweaters in the cabinet, Lilei leaves the hall, Xiaoli moves the vests and sweaters to the box. \\ 
		\colorbox{bestc!20}{Ability: Knowledge/Percepts-knowledge links (Question b1, b2)} \\
		\textbf{Question-b1}: What does Lilei see in the box?     \\
		(A) Vest (B) Sweater (C) Box (D) Sees nothing\\
		\textbf{Question-b2}: After Li Lei returns to the hall, does Li Lei know where the vest is now?     \\
		(A) Knows, in the briefcase. (B) Knows, in the box. (C) Knows, in the cabinet. \textbf{(D) Does not know.}\\ \hline

		\textbf{Story-c}: On a winter vacation evening, Wang Lei, Liu Ting, and Xiao Ming play toys together at home. They first play with the first novel toy - an electric car, then they play with the second novel toy - a puzzle. At this time, Wang Lei says, "I have something to do and need to leave the room", and he leaves after closing the door. Liu Ting says, ``Oh, he leaves, it doesn't matter, we play together.'' Liu Ting and Xiao Ming play with the third novel toy - colored pencils. After playing, they put all three toys in a plate on the table. At this time, Wang Lei comes back, looks at the three toys in the tray, and excitedly says, ``Oh, look! Look at that toy!'', ``Xiao Ming, can you pass it to me?''.\\ 
		\colorbox{bestc!20}{Ability: Knowledge/Knowledge-attention links} \\
		\textbf{Question-c}: What does Xiao Ming most likely do?     \\
		(A) Xiao Ming picks out the electric car and gives it to Wang Lei. (B) Xiao Ming picks out the puzzle and gives it to Wang Lei. \textbf{(C) Xiao Ming picks out the colored pencils and gives it to Wang Lei.} (D) Xiao Ming randomly picks one of the three toys and gives it to Wang Lei.\\ \hline
		
	\end{tabular}
	\caption{\texttt{3} abilities in \colorbox{bestc!20}{Knowledge} that are not covered by the 8 tasks and evaluated with extra samples.}
	\label{tab-example-knowledge}
	\vspace{0mm}
\end{table*}

\section{Detailed Data Statistics}
\label{appendix-statistics}
We present the detailed data statistics of 31 theory-of-mind abilities in Table \ref{tab-all-statistics}. For each task, we collect $\geq$100 samples, covering 19/31 abilities. Additionally, for the rest 12 abilities not covered in the 8 tasks, we further collect $\geq$20 extra samples per ability.

\renewcommand\arraystretch{0.9}
\begin{table*}[t]
	 \small
  \centering
	\hspace{0mm}	
	\setlength{\abovecaptionskip}{4mm}
	\setlength{\tabcolsep}{5mm}
		\begin{tabular}{llllll}
			\toprule
\textbf{Ability}& \textbf{\#S} & \textbf{\#Q} & \makecell{\textbf{ASL}\\\textbf{(En)}} & \makecell{\textbf{ASL}\\\textbf{(Zh)}}  & \textbf{Agr.}\\ \midrule
\multicolumn{6}{l}{\textbf{Emotion}} \\  \midrule
Typical emotional reactions & 100 & 100 & 32.14 & 52.45 & 100.0\% \\
Atypical emotional reactions & 100 & 100 & 50.78 & 81.12 & 100.0\% \\
Discrepant emotions & 20 & 40 & 27.00 & 41.55 & 100.0\% \\
Mixed emotions & 20 & 40 & 68.92 & 104.85 & 100.0\% \\
Hidden emotions & 20 & 80 & 41.90 & 68.50 & 100.0\% \\
Moral emotions & 20 & 40 & 108.50 & 175.07 & 97.5\% \\
Emotion regulation & 20 & 20 & 108.00 & 168.60 & 100.0\% \\ \midrule

\multicolumn{6}{l}{\textbf{Desire}} \\  \midrule
Multiple desires & 20 & 20 & 64.25 & 100.90 & 100.0\% \\
Desires influence on actions/emotions & 100 & 100 & 36.58 & 51.35 & 95.0\% \\
Desire-action contradiction  & 20 & 40 & 62.38 & 91.38 & 100.0\% \\
Discrepant desires & 20 & 20 & 79.80 & 133.75 & 100.0\% \\ \midrule

\multicolumn{6}{l}{\textbf{Intention}} \\  \midrule
Discrepant intentions & 20 & 40 & 101.28 & 162.07 & 100.0\% \\
Prediction of actions & 20 & 20 & 84.70 & 135.00 & 100.0\% \\
Intentions explanations & 213 & 260 & 78.58 & 125.43 & 100.0\% \\
Completion of failed actions & 20 & 20 & 94.70 & 140.60 & 100.0\% \\ \midrule

\multicolumn{6}{l}{\textbf{Knowledge}} \\  \midrule
Knowledge-pretend play links & 30 & 30 & 79.63 & 166.33 & 100.0\% \\
Percepts-knowledge links & 20 & 40 & 39.08 & 61.50 & 100.0\% \\
Information-knowledge links  & 100 & 200 & 47.17 & 76.89 & 100.0\% \\
Knowledge-attention links & 20 & 20 & 148.20 & 225.45 & 100.0\% \\ \midrule

\multicolumn{6}{l}{\textbf{Belief}} \\  \midrule
Content false beliefs & 50 & 200 & 61.62 & 96.00 & 100.0\% \\
Location false beliefs & 50 & 200 & 36.69 & 59.08 & 100.0\% \\
Identity false beliefs  & 20 & 40 & 70.92 & 120.80 & 100.0\% \\
Second-order beliefs & 100 & 200 & 61.63 & 96.00 & 100.0\% \\
Beliefs based action/emotions & 50 & 142 & 95.45 & 154.11 & 100.0\% \\
Sequence false beliefs & 100 & 100 & 32.46 & 52.45 & 100.0\% \\ \midrule
\multicolumn{6}{l}{\textbf{Non-Literal Communication}} \\  \midrule
Irony/Sarcasm & 20 & 26 & 55.81 & 90.23 & 100.0\% \\
Egocentric lies & 20 & 40 & 77.60 & 131.05 & 100.0\% \\
White lies & 20 & 40 & 66.12 & 102.70 & 100.0\% \\
Involuntary lies & 21 & 42 & 54.76 & 84.12 & 100.0\% \\
Humor & 20 & 40 & 67.78 & 115.45 & 100.0\% \\
Faux pas & 140 & 560 & 95.77 & 156.79 & 98.2\% \\ \bottomrule
			
		\end{tabular}
	
	\caption{Data statistics. {\#S}: Number of stories, {\#Q}: Number of questions, {ASL(En/Zh)}: Average story length (English/Chinese). Agr.: Final agreement.}
	\label{tab-all-statistics}
	\vspace{0mm}
\end{table*}

\section{Prompts for Data Translation}
\label{appendix-translation}
We present the prompts for Chinese to English translation of \logo in Table \ref{tab-translation-prompt}.

\begin{table*}[t]
\vspace{0mm}
\small
\renewcommand\arraystretch{1.1}
\setlength{\abovecaptionskip}{1mm}
\setlength{\belowcaptionskip}{0mm}
\centering
\setlength{\tabcolsep}{1.0mm}
\begin{tabular}{|m{0.986\textwidth}|}
\hline

\hline
You are an experienced translator who only uses the *Present Tense* of English in translating all Chinese texts.\\ \\
\lbrack Task\rbrack\\ 
Translate the given Chinese text to English. You should strictly follow the below rules.\\
(1) You should use the high school-level vocabulary with frequently-used words.\\
(2) You should only present the translation results without any other explanations.\\
(3) The translation result must be in the present tense, do not use any other tenses, regardless of the text content. \\
(4) All verbs in the translation results should be in the present tense, e.g., you should use `is' instead of `was', `are' instead of `were', `give/gives\ instead of `gave', `design/designs' instead of `designed', and et al.\\
(5) The given Chinese text in json format, with the keys being story, question, option\_a, option\_b, option\_c and option\_d. Please return the English translation in the original json format. Note that the above key remains unchanged, only translate their values.\\
\\
\lbrack Input Chinese Text\rbrack:\\
\{\texttt{Chinese Sample}\}
\\ \\
\lbrack Output English Translation\rbrack:

\\ \hline

\end{tabular}
\caption{Prompt for Chinese to English Translation.}
\label{tab-translation-prompt}
\vspace{0mm}
\end{table*}

\section{Prompts for ToM Evaluation}

\label{appendix-prompt}
We present the prompts for \logo evaluation in Table \ref{tab-evaluation-prompt}.

\begin{table*}[t]
	\vspace{0mm}
	\small
	\renewcommand\arraystretch{1.0}
	\setlength{\abovecaptionskip}{1mm}
	\setlength{\belowcaptionskip}{0mm}
	\centering
	\setlength{\tabcolsep}{1.0mm}
	\begin{tabular}{|m{0.986\textwidth}|}
		\hline
		\multicolumn{1}{|c|}{\textbf{Vanilla Prompt for Chinese Evaluation }}                                                                                                                                                                                                                                                                                                                                                                                                                                                                                                                       \\ \hline
		\begin{CJK*}{UTF8}{gbsn}下面给你提供一段故事，一个问题和若干答案选项，请你根据故事内容和给定的问题，按照常理推测，选择一个最可能的答案选项，并输出答案序号。\end{CJK*}\\
		\begin{CJK*}{UTF8}{gbsn}注意：\end{CJK*}\\
		\begin{CJK*}{UTF8}{gbsn}（1）请只输出最可能的答案序号，格式为：[[答案序号]]，例如，最可能的答案选项为“A. 手提包”，则输出“[[A]]”；\end{CJK*}\\
		\begin{CJK*}{UTF8}{gbsn}（2）请必须从给定的答案选项“A、B、C、D”中选择一个做为最可能的答案作为输出，无论故事中是否提供足够的信息，如果你认为故事里没有足够的信息选出答案，请随机输出“[[A]]”，“[[B]]”，“[[C]]”，“[[D]]”其中之一；\end{CJK*}\\
		\begin{CJK*}{UTF8}{gbsn}（3）请只输出在给定的信息下最可能的答案序号，不要输出其他内容。 \end{CJK*} \\
		\begin{CJK*}{UTF8}{gbsn}\lbrack 故事\rbrack  \end{CJK*}\\
		\{\texttt{Story}\} \\
		
		\begin{CJK*}{UTF8}{gbsn}\lbrack 问题\rbrack  \end{CJK*}\\
		\{\texttt{Questions}\} \\
		
		\begin{CJK*}{UTF8}{gbsn}\lbrack 答案选项\rbrack  \end{CJK*}\\
		A. \{\texttt{Option\_a}\} ~~~~B. \{\texttt{Option\_b}\} ~~~~C. \{\texttt{Option\_d}\} ~~~~D. \{\texttt{Option\_c}\} \\

		\\ \hline
		
		\multicolumn{1}{|c|}{\textbf{CoT Prompt for Chinese Evaluation}}                                                                                                                                                                                                                                                                                                                                                                                                                                                                                                                       \\ \hline
		\begin{CJK*}{UTF8}{gbsn}下面给你提供一段故事，一个问题和若干答案选项，请你根据故事内容和给定的问题，按照常理推测，选择一个最可能的答案选项，并输出答案序号。\end{CJK*}\\
		\begin{CJK*}{UTF8}{gbsn}注意：\end{CJK*} \\
		\begin{CJK*}{UTF8}{gbsn}（1）请先一步步思考，对问题的答案进行推理分析，最后请输出最可能的答案序号，格式为：[[答案序号]]，例如，最可能的答案选项为“A. 手提包”，则输出“[[A]]”；\end{CJK*}\\
		\begin{CJK*}{UTF8}{gbsn}（2）请必须从给定的答案选项“A、B、C、D”中选择一个做为最可能的答案作为输出，无论故事中是否提供足够的信息，如果你认为故事里没有足够的信息选出答案，请随机输出“[[A]]”，“[[B]]”，“[[C]]”，“[[D]]”其中之一；\end{CJK*}\\
		\begin{CJK*}{UTF8}{gbsn}（3）再次强调，你必须先给出一步步推理的结果，最后再输出最可能的答案序号。你不应该直接输出答案。\end{CJK*} \\
		... (Same as above)
		\\ \hline
		
		\multicolumn{1}{|c|}{\textbf{Vanilla Prompt for English Evaluation}}                                                                                                                                                                                                                                                                                                                                                                                                                                                                                                                      \\ \hline
		Below is a multiple-choice question with a story and serveral answer options. Based on the content of the story and the given question, please infer the most likely answer and output the answer index. \\
		Note:\\
		(1) Please only output the most likely answer index in the format: [[Answer Index]], for example, if the most likely answer option is `A. Handbag', then output `[[A]]';\\
		(2) You must choose one of the given answer options `A, B, C, D' as the most likely answer, regardless of whether the story provides enough information. If you think there is not enough information in the story to choose an answer, please randomly output one of ``[[A]]'', ``[[B]]'', ``[[C]]'', or ``[[D]]'';\\
		(3) Please only output the most likely answer index based on the given information, and do not output any other content.\\ 
		\text{[Story]}\\
		\{\texttt{Story}\} \\
		
		[Question]\\
		\{\texttt{Questions}\} \\
		
		[Candidate Answers]\\
		A. \{\texttt{Option\_a}\} ~~~~B. \{\texttt{Option\_b}\} ~~~~C. \{\texttt{Option\_d}\} ~~~~D. \{\texttt{Option\_c}\} \\
		
		\\ \hline
		
		\multicolumn{1}{|c|}{\textbf{CoT Prompt for English Evaluation}}                                                                                                                                                                                                                                                                                                                                                                                                                                                                                                                      \\ \hline
		Below is a multiple-choice question with a story and serveral answer options. Based on the content of the story and the given question, please infer the most likely answer and output the answer index. \\
		Note:\\
		(1) Please first think step by step, conduct analysis on the answers to the questions, and finally output the most likely answer index in the format: [[Answer Index]], for example, if the most likely answer option is `A. Handbag', then output `[[A]]';\\
		(2) You must choose one of the given answer options `A, B, C, D' as the most likely answer, regardless of whether the story provides enough information. If you think there is not enough information in the story to choose an answer, please randomly output one of ``[[A]]'', ``[[B]]'', ``[[C]]'', or ``[[D]]'';\\
		(3) Again, you must first output the results of step-by-step reasoning, and finally output the most likely answer index. You should not directly output the answer index.\\ 
		
		... (Same as above)\\
		
		\hline
		
	\end{tabular}
	\caption{Prompts for evaluation.}
	\label{tab-evaluation-prompt}
	\vspace{0mm}
\end{table*}

\section{Detailed Results of 31 Abilities}
\label{appendix-all-ability-results}
\input{tables/ability-results-all.tex}

We present the detailed experimental results of 31 theory-of-mind abilities in Table \ref{tab-all-ability-results}. Overall, LLMs have not yet achieved a general and robust human-level ToM performance. The GPT-4 series shows the best performance in most specific abilities and even surpasses human performance in 9 out of 31 abilities. We attribute this to two main reasons: First, certain abilities, such as \textit{Emotion/Typical emotional reactions}, relate to samples that are quite direct and simple. LLMs have encountered many similar training datasets, thus achieving high accuracy. Second, limited by human resources, some abilities have a relatively small number of samples, insufficient to showcase the general advantage of humans over LLMs.

\section{Detailed Results of Coherent Tests}
\label{appendix-coherent-results}
\input{tables/task-results-hard.tex}

We present the detailed results of coherent tests on all LLMs with vanilla/CoT prompting in Table \ref{tab-task-coherent-results}.

\section{Case Study for CoT Prompting}
\label{appendix-cot-case}

\begin{figure}[h]
	\centering
	\vspace{3mm}
	\hspace{0mm}
	\setlength{\abovecaptionskip}{2mm}
	\setlength{\belowcaptionskip}{2mm}
	\includegraphics[width=0.49\textwidth]{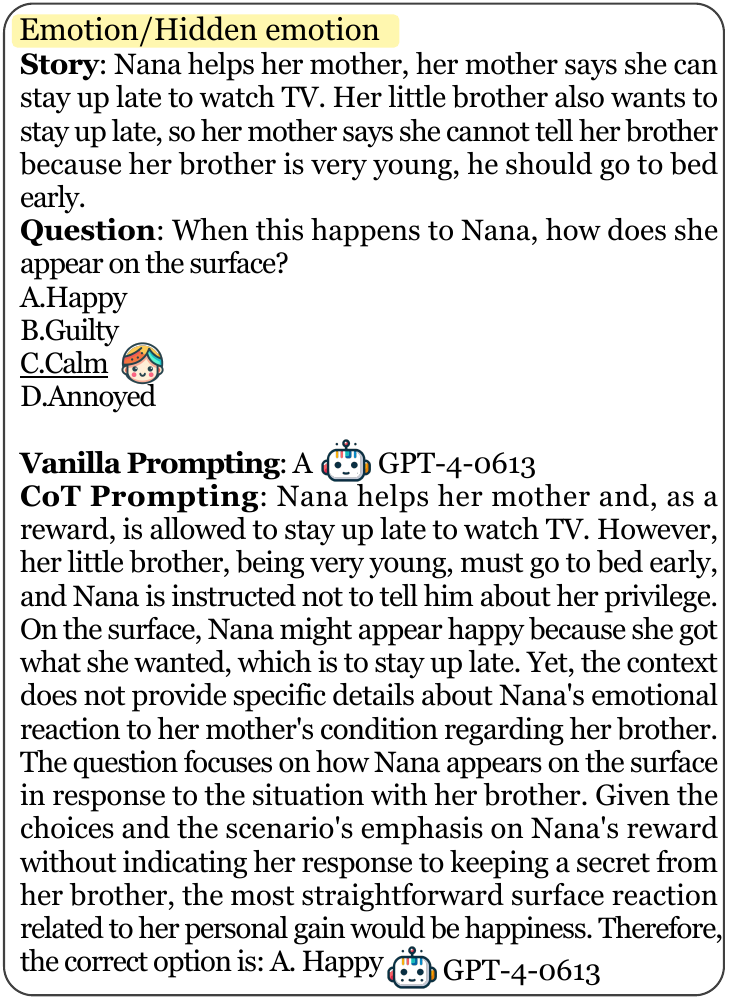}
	\centering
	\caption{
		A case of CoT prompting on GPT-4-0613.}
\label{fig-appendix-cot-case}
\vspace{-3.5mm}
\end{figure}

In Figure \ref{fig-appendix-cot-case}, we show a failed CoT reasoning process of GPT-4-0613. It's evident that while LLMs can break down questions into detailed parts, they still struggle to provide the correct answer.

\section{Use of AI Assistants}
We use ChatGPT to polish some of the content.

\end{document}

%% file: tables/task-results-soft.tex
\begin{table*}[h]
	\vspace{0mm}
	\small
	\renewcommand\arraystretch{0.9}
	\setlength{\abovecaptionskip}{0mm}
	\setlength{\belowcaptionskip}{5mm}
	\centering
	\setlength{\tabcolsep}{0.8mm}
		\resizebox{1.\textwidth}{!}{
			\begin{tabular}{l\myline cc|cc|cc|cc|cc|cc|cc|cc\myline cc}
				\toprule
				\multicolumn{19}{l}{\makecell[tl]{\textbb{UOT}: Unexpected Outcome Test ~~\textbb{SIT}: Scalar Implicature Task ~~\textbb{PST}: Persuasion Story Task ~~\textbb{FBT}: False Belief Task \\ \textbb{AST}: Ambiguous Story Task ~~~~~~~~\,\textbb{HT}: Hinting Test~~~~~~ ~~~~~~ ~~ ~~~\,~\textbb{SST}: Strange Story Task~~~~~~~ \textbb{FRT}: Faux-pas Recognition Test}}  \\ \midrule
				\multirow{2}{*}{SUBJECT}& \multicolumn{2}{c|}{\textbb{UOT}} & \multicolumn{2}{c|}{\textbb{SIT}} & \multicolumn{2}{c|}{\textbb{PST}} & \multicolumn{2}{c|}{\textbb{FBT}} & \multicolumn{2}{c|}{\textbb{AST}} & \multicolumn{2}{c|}{\textbb{HT}} & \multicolumn{2}{c|}{\textbb{SST}} & \multicolumn{2}{c\myline}{\textbb{FRT}} & \multicolumn{2}{c}{\textbf{\tbbg AVG.}}  \\
				& Zh	  & En	  & Zh	  & En	  & Zh	  & En	 & Zh	  & En	  & Zh	  & En	  & Zh	  & En	  & Zh	  & En	  & Zh	  & En	  & \tbbg Zh	  & \tbbg En	  \\ \midrule
				\textbf{Human} &  \multicolumn{2}{c|}{\textbf{89.3}}       &  \multicolumn{2}{c|}{\textbf{75.5}}&  \multicolumn{2}{c|}{\textbf{70.0}}&  \multicolumn{2}{c|}{\textbf{86.8}}&  \multicolumn{2}{c|}{\textbf{95.0}}&  \multicolumn{2}{c|}{\textbf{97.1}}&  \multicolumn{2}{c|}{\textbf{89.2}}&  \multicolumn{2}{c|}{\textbf{80.4}}&  \multicolumn{2}{c}{\tbbg \textbf{85.4}}      \\ \midrule
				ChatGLM3-6B	 					& 55.3      & 44.3      & 24.5      & 28.0      & 44.0      & 41.0      & 59.2      & 48.5      & 48.0      & 41.0      & 32.0      & 36.9      & 58.0      & 37.8      & 55.2      & 44.6      &\tbbg  47.0      & \tbbg 40.3      \\
				LLaMA2-13B-Chat    				& 43.7      & 52.7      & 28.0      & 23.5      & 38.0      & 43.0      & 42.2      & 42.8      & 38.0      & 47.5      & 32.0      & 48.5      & 58.2      & 58.0      & 47.9      & 58.4      & \tbbg 41.0      & \tbbg 46.8      \\
				Baichuan2-13B-Chat 				& 56.3      & 53.7      & 27.5      & 32.0      & 48.0      & 36.0      & 50.2      & 51.5      & 56.0      & 50.5      & 54.4      & 58.3      & 50.1      & 50.4      & 61.6      & 61.3      &\tbbg  50.5      &\tbbg  49.2      \\
				Mistral-7B	  					& 61.0      & 58.0      & 28.0      & 34.5      & 49.0      & 51.0      & 43.5      & 46.7      & 52.5      & 51.0      & 29.1      & 43.7      & 53.1      & 60.0      & 63.6      & 66.8      &\tbbg  47.5      &\tbbg  51.5      \\
				Mixtral-8x7B					& 68.0      & 58.7      & \bestd{49.5}      & 42.5      & 45.0      & 55.0      & 49.8      & 37.8      & 71.0      & 69.5      & 43.7      & 55.3      & 51.4      & 53.8      & 62.5      & 54.1      & \tbbg 55.1      & \tbbg 53.3      \\
				Qwen-14B-Chat					& 72.0      & 63.7      & 42.5      & 30.5      & 50.0      & 51.0      & 57.2      & 58.7      & 65.5      & 64.0      & 54.4      & 56.3      & 60.0      & 59.5      & 72.7      & 69.5      & \tbbg 59.3      &\tbbg  56.7      \\ 
				GPT-3.5-Turbo-0613					& 69.3      & 63.3      & 33.0      & 35.0      & 52.0      & 49.0      & 61.2      & 62.3      & 63.5      & 63.5      & 60.2      & 53.4      & 72.0      & 66.1      & 66.8      & 67.0      &\tbbg  59.8      & \tbbg 57.5      \\
				GPT-3.5-Turbo-1106					& 72.3      & 66.0      & 34.0      & 33.0      & 57.0      & 56.0      & 53.0      & 55.0      & 59.0      & 60.5      & 61.2      & 64.1      & 72.5      & 69.0      & 68.8      & 72.5      &\tbbg  59.7      & \tbbg 59.5      \\
				GPT-4-0613	  					& 71.3      & \bestd{71.3}      & 49.0      & 44.0      & 58.0      & 53.0      & 86.3      & 80.0      & \bestd{84.0}      & \bestd{78.0}      & 79.6      & 76.7      & \bestd{83.0}      & 81.1      & 76.6      & 71.8      & \tbbg 73.5      & \tbbg 69.5      \\
				GPT-4-1106	  					& \bestd{76.7}      & 71.0      & 48.0      & \bestd{49.0}      & \bestd{61.0}      & \bestd{65.0}      & \bestd{90.8}      & \bestd{88.2}      & 83.0      & 77.5      & \bestd{88.3}      & \bestd{82.5}      & 76.2      & \bestd{84.0}      & \bestd{78.6}      & \bestd{75.0}      & \tbbg \bestd{75.3}      & \tbbg \bestd{74.0}      \\\midrule
				ChatGLM3-6B + CoT				& 58.7      & 50.3      & 27.0      & 26.5      & 44.0      & 41.0      & 56.5      & 51.2      & 48.0      & 44.0      & 37.9      & 42.7      & 56.3      & 44.2      & 65.4      & 51.4      & \tbbg 49.2      & \tbbg 43.9      \\
				LLaMA2-13B-Chat + CoT			& 45.3      & 52.7      & 25.5      & 23.5      & 34.0      & 39.0      & 41.3      & 43.0      & 41.0      & 48.5      & 28.2      & 43.7      & 53.6      & 59.5      & 49.6      & 62.1      & \tbbg 39.8      & \tbbg 46.5      \\
				Baichuan2-13B-Chat + CoT		& 54.3      & 48.7      & 26.5      & 23.0      & 33.0      & 34.0      & 44.8      & 44.2      & 51.5      & 44.0      & 53.4      & 49.5      & 52.8      & 51.1      & 65.4      & 52.5      & \tbbg 47.7      &\tbbg  43.4      \\
				Mistral-7B + CoT				& 61.0      & 55.3      & 27.0      & 28.0      & 46.0      & 42.0      & 47.2      & 47.0      & 47.0      & 46.5      & 30.1      & 37.9      & 56.5      & 63.4      & 64.3      & 64.1      & \tbbg 47.4      & \tbbg 48.0      \\
				Mixtral-8x7B + CoT				& 65.3      & 52.3      & 45.0      & 29.5      & 41.0      & 39.0      & 53.7      & 43.8      & 66.0      & 59.5      & 44.7      & 54.4      & 43.7      & 39.8      & 47.5      & 54.3      & \tbbg 50.9      & \tbbg 46.6      \\ 
				Qwen-14B-Chat + CoT				& 65.3      & 58.0      & 31.5      & 31.0      & 45.0      & 44.0      & 51.3      & 54.7      & 62.5      & 63.0      & 47.6      & 48.5      & 60.2      & 53.6      & 70.7      & 67.7      & \tbbg 54.3      & \tbbg 52.6      \\
				GPT-3.5-Turbo-0613 + CoT				& 62.3      & 58.3      & 30.0      & 26.5      & 43.0      & 48.0      & 57.8      & 64.0      & 58.5      & 58.0      & 41.7      & 41.7      & 71.3      & 66.8      & 70.5      & 70.4      & \tbbg 54.4      & \tbbg 54.2      \\
				GPT-3.5-Turbo-1106 + CoT				& 68.7      & 64.7      & 27.5      & 35.0      & 45.0      & 54.0      & 57.5      & 56.3      & 61.5      & 63.0      & 46.6      & 51.5      & 71.3      & 68.6      & 72.7      & 70.9      & \tbbg 56.4      & \tbbg 58.0      \\ 
				GPT-4-0613 + CoT				& 72.3      & 64.7      & 43.5      & 54.0      & 55.0      & 52.0      & \bestc{90.3}      & 80.8      & \bestc{84.5}      & 77.5      & 78.6      & 76.7      & \bestc{83.5}      & 81.1      & 74.3      & 73.6      & \tbbg 72.8      & \tbbg 70.1      \\
				GPT-4-1106 + CoT				& \bestc{76.3}      & \bestc{72.7}      & \bestc{48.0}      & \bestc{55.0}      & \bestc{59.0}      & \bestc{55.0}      & 88.7      & \bestc{86.8}      & 84.0      & \bestc{81.0}      & \bestc{89.3}      & \bestc{82.5}      & 76.9      & \bestc{84.3}      & \bestc{79.6}      & \bestc{75.2}      & \tbbg \bestc{75.2}      & \tbbg \bestc{74.1}      \\ \midrule
				\tbbg LLM Grand Mean &\tbbg  64.6 &\tbbg 60.3 &\tbbg 36.4 &\tbbg 35.2 &\tbbg 50.2 &\tbbg 50.0 &\tbbg 59.3 &\tbbg 57.2 &\tbbg 62.1 &\tbbg 60.3 &\tbbg 53.5 &\tbbg 57.6 &\tbbg 63.5 &\tbbg 62.0 &\tbbg 65.4 &\tbbg 64.1 &\tbbgg 56.9 &\tbbgg 55.8 
				\\
				\tbbg LLM Grand Mean + CoT &\tbbg  {63.0} &\tbbg 57.8 &\tbbg 33.2 &\tbbg 33.2 &\tbbg 44.5 &\tbbg 44.8 &\tbbg 58.9 &\tbbg 57.2 &\tbbg 60.5 &\tbbg 58.5 &\tbbg 49.8 &\tbbg 52.9 &\tbbg 62.6 &\tbbg 61.2 &\tbbg 66.0 &\tbbg 64.2 &\tbbgg 54.8 &\tbbgg 53.7 
				\\

				\bottomrule
			\end{tabular}
		}
		
		\vspace{1mm}
		\caption{Task-oriented ToM performance in accuracy. The best results of vanilla prompting are in \bestd{pink}, and those of CoT prompting are in \bestc{blue}. ``LLM Grand Mean'' is the average performance of all ten LLMs (same as below). 
		}
		
		\label{tab-task-soft}
		\vspace{-6mm}
	\end{table*}

%% file: tables/ability-results-category.tex
\begin{table*}[h]
\vspace{0mm}
\small
\renewcommand\arraystretch{0.9}
\setlength{\abovecaptionskip}{0mm}
\setlength{\belowcaptionskip}{0mm}
\centering
\setlength{\tabcolsep}{1.8mm}
\resizebox{1.\textwidth}{!}{
\begin{tabular}{l\myline cc|cc|cc|cc|cc|cc\myline cc}
\toprule
\multirow{2}{*}{SUBJECT} & \multicolumn{2}{c|}{\textbf{Emotion}} & \multicolumn{2}{c|}{\textbf{Desire}} & \multicolumn{2}{c|}{\textbf{Intention}} & \multicolumn{2}{c|}{\textbf{Knowledge}} & \multicolumn{2}{c|}{\textbf{Belief}} & \multicolumn{2}{c\myline}{\textbf{NL Comm.}} & \multicolumn{2}{c}{\tbbg \textbf{AVG.}} \\ 
& Zh	  & En	  & Zh	  & En	  & Zh	  & En	 & Zh	  & En	  & Zh	  & En	  & Zh	  & En	  & \tbbg Zh	  & \tbbg En	  \\\midrule
\textbf{Human}              &  \multicolumn{2}{c|}{\textbf{86.4}}       &  \multicolumn{2}{c|}{\textbf{78.2}}&  \multicolumn{2}{c|}{\textbf{90.4}}&  \multicolumn{2}{c|}{\textbf{82.2}}&  \multicolumn{2}{c|}{\textbf{89.3}}&  \multicolumn{2}{c|}{\textbf{89.0}}&  \multicolumn{2}{c}{\tbbg \textbf{86.1}}     \\ \midrule
ChatGLM3-6B              & 54.9         & 42.2         & 52.0         & 40.7        & 52.0          & 35.9          & 16.8          & 22.0          & 55.0         & 44.5        & 49.8                  & 38.5                  & \tbbg 46.8        & \tbbg 37.3       \\
LLaMA2-13B-Chat          & 38.4         & 51.0         & 39.2         & 49.4        & 41.7          & 49.6          & 22.4          & 21.1          & 46.7         & 49.0        & 54.0                  & 54.3                  & \tbbg 40.4        & \tbbg 45.7       \\
Baichuan2-13B-Chat       & 55.9         & 53.1         & 49.6         & 46.0        & 63.5          & 52.2          & 21.3          & 20.9          & 48.5         & 49.8        & 46.2                  & 50.1                  & \tbbg 47.5        & \tbbg 45.4       \\
Mistral-7B               & 54.0         & 58.1         & 48.7         & 49.8        & 45.3          & 52.2          & 33.1          & 42.0          & 47.2         & 48.7        & 46.5                  & 57.2                  & \tbbg 45.8        & \tbbg 51.3       \\
Mixtral-8x7B             & 61.6         & 56.6         & 54.1         & 51.2        & 60.1          & 64.1          & 31.1          & 27.1          & 56.9         & 48.1        & 50.9                  & 57.9                  & \tbbg 52.5        &\tbbg  50.8       \\
Qwen-14B-Chat            & 66.8         & 65.8         & 57.0         & 52.9        & 66.4          & 58.9          & 37.9          & 33.1          & 62.2         & 60.6        & 53.2                  & 57.5                  & \tbbg 57.3        & \tbbg 54.8       \\
GPT-3.5-Turbo-0613             & 58.4         & 65.6         & 54.2         & 53.4        & 58.2          & 61.0          & 37.8          & 36.3          & 64.3         & 61.4        & 76.8                  & 66.9                  & \tbbg 58.3        & \tbbg 57.4       \\
GPT-3.5-Turbo-1106             & 61.6         & 60.6         & 57.1         & 60.7        & 56.5          & 62.6          & 30.4          & 37.4          & 60.6         & 59.4        & 76.0                  & 71.5                  & \tbbg 57.0        & \tbbg 58.7       \\
GPT-4-0613               & \bestd{79.0}         & 72.0         & \bestd{72.2}         & 60.2        & \bestd{77.8}          & 66.1          & 56.0          & 48.1          & 82.1         & 76.1        & \bestd{81.3}                  & 81.5                  & \tbbg \bestd{74.7}        & \tbbg 67.3       \\
GPT-4-1106               & 75.9         & \bestd{75.7}         & 67.5         & \bestd{69.7}        & \bestd{77.8}          &\bestd {84.7}          & \bestd{57.6}          & \bestd{52.1}          & \bestd{84.1}         & \bestd{82.8}        & 72.8                  & \bestd{84.0}                  & \tbbg 72.6        & \tbbg \bestd{74.8}       \\ \midrule

ChatGLM3-6B + CoT        & 53.0         & 46.7         & 49.1         & 43.7        & 54.8          & 49.8          & 32.0          & 28.9          & 51.7         & 48.6        & 55.8                  & 40.1                  & \tbbg 49.4        & \tbbg 43.0       \\
LLaMA2-13B-Chat + CoT    & 43.3         & 48.1         & 37.4         & 44.9        & 43.4          & 51.7          & 28.7          & 30.7          & 43.8         & 47.9        & 52.9                  & 62.7                  & \tbbg 41.6        & \tbbg 47.7       \\
Baichuan2-13B-Chat + CoT & 51.6         & 49.7         & 47.2         & 37.5        & 51.3          & 47.8          & 33.7          & 19.3          & 47.3         & 45.2        & 52.4                  & 47.5                  & \tbbg 47.3        & \tbbg 41.2       \\
Mistral-7B + CoT         & 52.0         & 57.9         & 46.9         & 45.1        & 50.5          & 51.1          & 33.4          & 44.5          & 50.9         & 50.1        & 50.7                  & 62.4                  & \tbbg 47.4        & \tbbg 51.9       \\
Mixtral-8x7B + CoT       & 56.9         & 56.0         & 47.5         & 41.5        & 57.9          & 55.3          & 30.2          & 33.2          & 54.6         & 44.3        & 44.6                  & 45.5                  & \tbbg 48.6        & \tbbg 46.0       \\
Qwen-14B-Chat + CoT      & 63.9         & 62.7         & 57.3         & 50.2        & 63.2          & 57.8          & 41.0          & 40.1          & 56.2         & 53.6        & 53.5                  & 53.2                  & \tbbg 55.9        & \tbbg 52.9       \\
GPT-3.5-Turbo-0613 + CoT       & 61.6         & 62.7         & 53.1         & 52.1        & 65.4          & 63.8          & 49.6          & 43.3          & 58.2         & 58.7        & 70.0                  & 71.6                  & \tbbg 59.7        & \tbbg 58.7       \\
GPT-3.5-Turbo-1106 + CoT       & 63.2         & 62.3         & 54.7         & 54.7        & 59.9          & 63.1          & 34.6          & 49.6          & 61.9         & 59.9        & 71.3                  & 70.8                  & \tbbg 57.6        & \tbbg 60.1       \\
GPT-4-0613 + CoT         & \bestc{76.8}         & 73.1         & 69.9         & \bestc{67.1}        & \bestc{80.1}          & 71.5          & 60.5          & 57.5          & 83.7         & 76.4        & \bestc{80.9}                  & 82.2                  & \tbbg \bestc{75.3}        & \tbbg 71.3       \\
GPT-4-1106 + CoT         & \bestc{76.8}         & \bestc{73.2}         & \bestc{71.2}         & 63.3        & 78.9          & \bestc{77.9}          & \bestc{63.1}          & \bestc{60.4}          & \bestc{84.0}        & \bestc{83.6}        & 70.9                  & \bestc{83.0}                  & \tbbg 74.2        & \tbbg \bestc{73.6}     \\ \midrule
\tbbg LLM Grand Mean &  \tbbg60.7 & \tbbg60.1 &\tbbg55.2 &\tbbg53.4 &\tbbg59.9 &\tbbg58.7 &\tbbg34.4 &\tbbg34.0 &\tbbg60.8 &\tbbg58.0 &\tbbg60.8 &\tbbg61.9 &\tbbgg55.3 &\tbbgg54.4 \\
\tbbg LLM Grand Mean + CoT&\tbbg 59.9 &\tbbg59.2 &\tbbg53.4 &\tbbg50.0 &\tbbg60.5 &\tbbg59.0 &\tbbg40.7 &\tbbg40.8 &\tbbg59.2 &\tbbg56.8 &\tbbg60.3 &\tbbg61.9 &\tbbgg55.7 &\tbbgg 54.6  \\\bottomrule 
\end{tabular}
}

\vspace{1mm}
\caption{Ability-oriented ToM performance in accuracy macro-averaged on 6 categories due to space limitation, while the complete results of all 31 abilities can be found in Appendix \ref{appendix-all-ability-results}, Table \ref{tab-all-ability-results}. 
}

\label{tab-ability-category}
\vspace{-4mm}
\end{table*}

%% file: tables/ability-results-all.tex
\begin{table*}[h]
	\vspace{0mm}
	\small
	\renewcommand\arraystretch{0.9}
	\setlength{\abovecaptionskip}{0mm}
	\setlength{\belowcaptionskip}{5mm}
	\centering
	\setlength{\tabcolsep}{2.mm}
		\resizebox{1.\textwidth}{!}{
			\begin{tabular}{ccc|cccccccccc|cccccccccc}
				\toprule
				\multicolumn{23}{l}{\makecell[tl]{\textbf{Large Language Models}: (M0) ChatGLM3-6B ~~~~ (M1) LLaMA2-13B-Chat ~~~~ (M2) Baichuan2-13B-Chat ~~~~ (M3) Mistral-7B ~~~~ (M4) Mixtral-8x7B \\~~~~ ~~~~ ~~~~ ~~~~ ~~~~ ~~~~ ~~~~ ~~~~ ~~~~(M5) Qwen-14B-Chat ~~ (M6) GPT-3.5-Turbo-0613 ~~ (M7) GPT-3.5-Turbo-1106 ~~~~ (M8) GPT-4-0613 ~~ (M9) GPT-4-1106}} \\ \midrule
				Index        & Human          & Lang.   & M0     & M1     & M2     & M3     & M4     & M5     & M6     & M7     & M8      & M9      & \makecell{M0\\+\\CoT}     & \makecell{M1\\+\\CoT}     & \makecell{M2\\+\\CoT}     & \makecell{M3\\+\\CoT}     & \makecell{M4\\+\\CoT}     & \makecell{M5\\+\\CoT}     & \makecell{M6\\+\\CoT}     & \makecell{M7\\+\\CoT}     & \makecell{M8\\+\\CoT}      & \makecell{M9\\+\\CoT}    \\ \midrule
				\multicolumn{23}{l}{\makecell[tl]{\textbf{Emotion}: (I) Typical emotional reactions ~~~ (II) Atypical emotional reactions ~~~ (III) Discrepant emotions ~~~ (IV) Mixed emotions ~~~ (V) Hidden emotions \\~~~~~~~~~~~~~~~~~(VI) Moral emotions  ~~~  ~~~  ~~~  ~~~  ~~~ (VII) Emotion regulation}}                                       \\ \midrule
				
				\multirow{2}{*}{I}  & \multirow{2}{*}{\textbf{93.0}}   & zh      & 89.0   & 63.0   & 84.0   & 84.0   & 89.0   & 96.0   & 91.0   & 91.0   & 96.0    & \bestd{97.0}    & 89.0   & 67.0   & 83.0   & 84.0   & 86.0   & 88.0   & 89.0   & 89.0   & 91.0    & \bestc{93.0}   \\
				
				&  & en    & 71.0   & 83.0   & 86.0   & 83.0   & 73.0   & 89.0   & 90.0   & 86.0   & 90.0    & \bestd{94.0}    & 79.0   & 81.0   & 81.0   & 80.0   & 67.0   & 85.0   & 73.0   & 83.0   & 89.0    & \bestc{90.0}   \\ \midrule
				
				\multirow{2}{*}{II} & \multirow{2}{*}{\textbf{94.0}}   & zh      & 33.0   & 33.0   & 42.0   & 54.0   & 52.0   & 59.0   & 54.0   & \bestd{64.0}   & 63.0    & \bestd{64.0}    & 46.0   & 37.0   & 38.0   & 50.0   & 51.0   & 58.0   & 55.0   & 61.0   & \bestc{65.0}    & 62.0   \\
				
				&    & en      & 23.0   & 30.0   & 36.0   & 49.0   & 48.0   & 50.0   & 52.0   & \bestd{56.0}   & 49.0    & 55.0    & 29.0   & 32.0   & 33.0   & 48.0   & 51.0   & 48.0   & \bestc{56.0}   & 54.0   & 53.0    & 55.0   \\ \midrule
				
				\multirow{2}{*}{III} & \multirow{2}{*}{\textbf{92.5}}  & zh      & 52.5   & 27.5   & 40.0   & 42.5   & 57.5   & 62.5   & 52.5   & 62.5   & \bestd{90.0}    & \bestd{90.0}    & 57.5   & 37.5   & 37.5   & 50.0   & 50.0   & 65.0   & 70.0   & 60.0   & 70.0    & \bestc{85.0}   \\
				
				&    & en      & 35.0   & 47.5   & 55.0   & 47.5   & 65.0   & 65.0   & 67.5   & 70.0   & 85.0    & \bestd{92.5}    & 45.0   & 50.0   & 52.5   & 57.5   & 67.5   & 70.0   & 67.5   & 67.5   & 85.0    & \bestc{90.0}   \\ \midrule
				
				\multirow{2}{*}{IV}  & \multirow{2}{*}{\textbf{70.0}}  & zh      & 60.0   & 52.5   & 55.0   & 55.0   & 47.5   & 60.0   & 30.0   & 37.5   & \bestd{70.0}    & 60.0    & 47.5   & 40.0   & 57.5   & 50.0   & 37.5   & 55.0   & 45.0   & 70.0   & \bestc{82.5}    & 72.5   \\
				
				&    & en      & 35.0   & 52.5   & 52.5   & \bestd{65.0}   & 35.0   & 62.5   & 57.5   & 45.0   & \bestd{65.0}    & \bestd{65.0}    & 55.0   & 40.0   & 45.0   & 65.0   & 32.5   & 50.0   & 55.0   & 52.5   & \bestc{72.5}    & 67.5   \\ \midrule
				
				\multirow{2}{*}{V}   & \multirow{2}{*}{\textbf{95.0}}  & zh      & 55.0   & 40.0   & 62.5   & 55.0   & 67.5   & 75.0   & 58.8   & 58.8   & \bestd{88.7}    & 82.5    & 48.7   & 46.3   & 55.0   & 47.5   & 66.2   & 73.8   & 60.0   & 62.5   & \bestc{83.8}    & 82.5   \\
				
				&    & en      & 46.3   & 46.3   & 52.5   & 60.0   & 60.0   & 68.8   & 65.0   & 62.5   & \bestd{80.0}    & 78.7    & 43.8   & 48.7   & 43.8   & 60.0   & 56.2   & 66.2   & 70.0   & 61.3   & \bestc{80.0}    & \bestc{80.0}   \\ \midrule
				
				\multirow{2}{*}{VI}  & \multirow{2}{*}{\textbf{90.0}}  & zh      & 55.0   & 27.5   & 67.5   & 57.5   & 77.5   & 70.0   & 62.5   & 72.5   & \bestd{90.0}    & 82.5    & 52.5   & 50.0   & 60.0   & 62.5   & 72.5   & 67.5   & 72.5   & 60.0   & 85.0    & \bestc{92.5}   \\
				
				&    & en      & 50.0   & 62.5   & 55.0   & 77.5   & 80.0   & 75.0   & 77.5   & 70.0   & 85.0    & \bestd{90.0}    & 55.0   & 55.0   & 62.5   & 70.0   & 82.5   & 70.0   & 67.5   & 72.5   & 87.5    & \bestc{90.0}   \\ \midrule
				
				\multirow{2}{*}{VII}  & \multirow{2}{*}{\textbf{70.0}} & zh      & 40.0   & 25.0   & 40.0   & 30.0   & 40.0   & 45.0   & \bestd{60.0}   & 45.0   & 55.0    & 55.0    & 30.0   & 25.0   & 30.0   & 20.0   & 35.0   & 40.0   & 40.0   & 40.0   & \bestc{60.0}    & 50.0   \\
				
				&    & en      & 35.0   & 35.0   & 35.0   & 25.0   & 35.0   & 50.0   & 50.0   & 35.0   & 50.0    & \bestd{55.0}    & 20.0   & 30.0   & 30.0   & 25.0   & 35.0   & \bestc{50.0}   & \bestc{50.0}   & 45.0   & 45.0    & 40.0   \\ \midrule
				
				\multicolumn{23}{l}{\makecell[tl]{\textbf{Desire}: (I) Multiple desires ~~ (II) Desires influence on actions/emotions ~~ (III) Desire-action contradiction ~~ (IV) Discrepant desires}}                                                                                   \\ \midrule
				
				\multirow{2}{*}{I} & \multirow{2}{*}{\textbf{100}}    & zh      & 55.0   & 40.0   & 65.0   & 55.0   & 80.0   & 65.0   & 60.0   & 70.0   & \bestd{100}   & 95.0    & 50.0   & 50.0   & 55.0   & 50.0   & 65.0   & 60.0   & 65.0   & 70.0   & \bestc{100}   & 95.0   \\
				
				&    & en      & 50.0   & 65.0   & 70.0   & 55.0   & 70.0   & 75.0   & 65.0   & 80.0   & 90.0    & \bestd{100}   & 55.0   & 55.0   & 50.0   & 60.0   & 60.0   & 80.0   & 50.0   & 65.0   & \bestc{95.0}    & 90.0   \\ \midrule
				
				\multirow{2}{*}{II}  & \multirow{2}{*}{\textbf{71.7}}  & zh      & 47.5   & 39.3   & 41.6   & 49.4   & 45.3   & 50.0   & 55.6   & \bestd{58.9}   & 58.1    & 58.7    & 50.4   & 30.9   & 31.7   & 47.4   & 41.3   & 49.6   & 44.0   & 46.7   & 56.1    & \bestc{63.1}   \\
				
				&    & en      & 45.5   & 41.1   & 35.1   & 52.1   & 51.9   & 43.6   & 52.2   & 55.4   & 50.6    & \bestd{64.2}    & 45.5   & 38.5   & 33.8   & 39.0   & 40.0   & 41.8   & 51.5   & \bestc{55.5}   & 52.8    & 53.3   \\\midrule
				
				\multirow{2}{*}{III} & \multirow{2}{*}{\textbf{77.5}}  & zh      & 65.0   & 42.5   & 60.0   & 55.0   & 50.0   & 65.0   & 60.0   & 62.5   & \bestd{75.0}    & \bestd{75.0}    & 70.0   & 40.0   & 62.5   & 50.0   & 40.0   & 72.5   & 72.5   & 70.0   & 72.5    & \bestc{75.0}   \\
				
				&    & en      & 32.5   & 55.0   & 55.0   & 65.0   & 52.5   & 62.5   & 62.5   & 67.5   & 70.0    & \bestd{75.0}    & 47.5   & 52.5   & 55.0   & 52.5   & 32.5   & 57.5   & 72.5   & 67.5   & 70.0    & \bestc{75.0}   \\\midrule
				
				\multirow{2}{*}{IV}  & \multirow{2}{*}{\textbf{70.0}}  & zh      & 45.0   & 35.0   & 40.0   & 35.0   & 50.0   & 55.0   & 40.0   & 35.0   & \bestd{70.0}    & 50.0    & 25.0   & 35.0   & 55.0   & 40.0   & 50.0   & 55.0   & 40.0   & 40.0   & \bestc{65.0}    & 60.0   \\
				
				&    & en      & 30.0   & \bestd{45.0}   & 35.0   & 25.0   & 30.0   & 40.0   & 35.0   & \bestd{45.0}   & 40.0    & \bestd{45.0}    & 25.0   & 40.0   & 15.0   & 35.0   & 35.0   & 30.0   & 35.0   & 30.0   & \bestc{65.0}    & 45.0   \\ \midrule
				
				\multicolumn{23}{l}{\makecell[tl]{\textbf{Intention}: (I) Discrepant intentions ~~~~(II) Prediction of actions ~~~~(III) Intentions explanations ~~~~(IV) Completion of failed actions}}                                                                                     \\ \midrule
				
				\multirow{2}{*}{I}  & \multirow{2}{*}{\textbf{95.0}}   & zh      & 47.5   & 55.0   & 67.5   & 62.5   & 72.5   & 75.0   & 57.5   & 62.5   & \bestd{92.5}    & 82.5    & 50.0   & 37.5   & 57.5   & 65.0   & 67.5   & 65.0   & 75.0   & 70.0   & \bestc{90.0}    & \bestc{90.0}   \\
				
				&    & en      & 45.0   & 65.0   & 57.5   & 60.0   & 75.0   & 72.5   & 65.0   & 72.5   & 90.0    & \bestd{95.0}    & 50.0   & 60.0   & 42.5   & 65.0   & 67.5   & 72.5   & 75.0   & 75.0   & 92.5    & \bestc{95.0}   \\ \midrule
				
				\multirow{2}{*}{II}  & \multirow{2}{*}{\textbf{95.0}}  & zh      & 60.0   & 40.0   & 75.0   & 30.0   & 70.0   & \bestd{80.0}   & 60.0   & 50.0   & \bestd{80.0}    & \bestd{80.0}    & 70.0   & 45.0   & 50.0   & 45.0   & 65.0   & 70.0   & 70.0   & 65.0   & \bestc{80.0}    & 75.0   \\
				
				&    & en      & 35.0   & 50.0   & 45.0   & 40.0   & \bestd{85.0}   & 55.0   & 65.0   & 65.0   & 50.0    & \bestd{85.0}    & 50.0   & 50.0   & 55.0   & 35.0   & 65.0   & 55.0   & 70.0   & \bestc{75.0}   & 55.0    & 70.0   \\ \midrule
				
				\multirow{2}{*}{III}  & \multirow{2}{*}{\textbf{96.5}} & zh      & 50.4   & 41.9   & 61.5   & 48.5   & 57.7   & 65.4   & 70.4   & 68.5   & 83.8    & \bestd{88.5}    & 49.2   & 41.2   & 52.7   & 46.9   & 54.2   & 62.7   & 61.5   & 59.6   & 85.4    & \bestc{90.4}   \\
				
				&    & en      & 38.5   & 53.5   & 56.2   & 53.8   & 61.2   & 63.1   & 63.8   & 67.7   & 79.2    & \bestd{83.8}    & 44.2   & 51.9   & 48.5   & 49.2   & 53.8   & 58.8   & 55.0   & 62.3   & 78.5    & \bestc{86.5}   \\ \midrule
				
				\multirow{2}{*}{IV} & \multirow{2}{*}{\textbf{75.0}}   & zh      & 50.0   & 30.0   & 50.0   & 40.0   & 40.0   & 45.0   & 45.0   & 45.0   & 55.0    & \bestd{60.0}    & 50.0   & 50.0   & 45.0   & 45.0   & 45.0   & 55.0   & 55.0   & 45.0   & \bestc{65.0}    & 60.0   \\
				
				&    & en      & 25.0   & 30.0   & 50.0   & 55.0   & 35.0   & 45.0   & 50.0   & 45.0   & 45.0    & \bestd{75.0}    & 55.0   & 45.0   & 45.0   & 55.0   & 35.0   & 45.0   & 55.0   & 40.0   & \bestc{60.0}    & \bestc{60.0}   \\ \midrule
				
				\multicolumn{23}{l}{\makecell[tl]{\textbf{Knowledge}: (I) Knowledge-pretend play links ~~~~(II) Percepts-knowledge links ~~~~(III) Information-knowledge links ~~~~(IV) Knowledge-attention links}}     
				                                                                    \\ \midrule
				\multirow{2}{*}{I}   & \multirow{2}{*}{\textbf{93.3}}  & zh      & 10.0   & 6.7    & 10.0   & 16.7   & 10.0   & 16.7   & 13.3   & 20.0   & \bestd{40.0}    & \bestd{40.0}    & 13.3   & 26.7   & 23.3   & 26.7   & 13.3   & 20.0   & 43.3   & 23.3   & 53.3    & \bestc{66.7}   \\
				
				&    & en      & 10.0   & 13.3   & 6.7    & 13.3   & 23.3   & 16.7   & 10.0   & 16.7   & 3.3     & \bestd{26.7}    & 16.7   & 16.7   & 6.7    & 20.0   & 33.3   & 16.7   & 26.7   & 33.3   & 23.3    & \bestc{46.7}   \\ \midrule
				
				\multirow{2}{*}{II} & \multirow{2}{*}{\textbf{95.0}}   & zh      & 2.5    & 5.0    & 32.5   & 62.5   & 30.0   & 62.5   & 80.0   & 47.5   & \bestd{100}   & 97.5    & 47.5   & 22.5   & 50.0   & 60.0   & 22.5   & 82.5   & 85.0   & 52.5   & 95.0    & \bestc{97.5}   \\
				
				&    & en      & 25.0   & 12.5   & 10.0   & 85.0   & 32.5   & 55.0   & 80.0   & 80.0   & \bestd{100}   & 92.5    & 42.5   & 47.5   & 17.5   & 85.0   & 50.0   & 77.5   & 90.0   & 75.0   & \bestc{97.5}    & 90.0   \\ \midrule
				
				\multirow{2}{*}{III} & \multirow{2}{*}{\textbf{75.5}}  & zh      & 24.5   & 28.0   & 27.5   & 28.0   & \bestd{49.5}   & 42.5   & 33.0   & 34.0   & 49.0    & 48.0    & 27.0   & 25.5   & 26.5   & 27.0   & 45.0   & 31.5   & 30.0   & 27.5   & 43.5    & \bestc{48.0}   \\
				
				&    & en      & 28.0   & 23.5   & 32.0   & 34.5   & 42.5   & 30.5   & 35.0   & 33.0   & 44.0    & \bestd{49.0}    & 26.5   & 23.5   & 23.0   & 28.0   & 29.5   & 31.0   & 26.5   & 35.0   & 54.0    & \bestc{55.0}   \\ \midrule
				
				\multirow{2}{*}{IV} & \multirow{2}{*}{\textbf{65.0}}   & zh      & 30.0   & \bestd{50.0}   & 15.0   & 25.0   & 35.0   & 30.0   & 25.0   & 20.0   & 35.0    & 45.0    & 40.0   & 40.0   & 35.0   & 20.0   & 40.0   & 30.0   & 40.0   & 35.0   & \bestc{50.0}    & 40.0   \\
				
				&    & en      & 25.0   & 35.0   & 35.0   & 35.0   & 10.0   & 30.0   & 20.0   & 20.0   & \bestd{45.0}    & 40.0    & 30.0   & 35.0   & 30.0   & 45.0   & 20.0   & 35.0   & 30.0   & \bestc{55.0}   & \bestc{55.0}    & 50.0   \\ \midrule
				
				\multicolumn{23}{l}{\makecell[tl]{\textbf{Belief}: (I) Content false beliefs ~~~~~~~~~~~~~~~~~~ (II) Location false beliefs ~~~~~~~~~~~~~~~~~~ (III) Identity false beliefs ~~~~~~~~~~~~~~~~~~ (IV) Second-order beliefs \\~~~~~~~~~~~~(V) Beliefs based action/emotions ~~(VI) Sequence false beliefs}}                                        \\ \midrule
				
				\multirow{2}{*}{I}  & \multirow{2}{*}{\textbf{82.5}}   & zh      & 55.0   & 50.5   & 53.0   & 38.0   & 53.5   & 51.0   & 53.0   & 54.0   & 71.0    & \bestd{75.0}    & 50.0   & 48.0   & 40.5   & 33.5   & 58.0   & 44.0   & 43.5   & 45.5   & 73.5    & \bestc{74.0}   \\
				
				&    & en      & 48.0   & 48.0   & 58.0   & 45.5   & 39.0   & 57.5   & 63.0   & 54.5   & 71.0    & \bestd{82.5}    & 53.5   & 45.5   & 32.0   & 36.5   & 41.0   & 61.0   & 58.5   & 57.5   & 71.5    & \bestc{77.5}   \\ \midrule
				
				\multirow{2}{*}{II} & \multirow{2}{*}{\textbf{97.0}}   & zh      & 67.5   & 46.0   & 76.0   & 65.5   & 74.5   & 73.0   & 75.0   & 76.0   & \bestd{100}   & \bestd{100}   & 70.5   & 51.0   & 65.5   & 67.5   & 77.0   & 73.5   & 73.0   & 75.0   & \bestc{100}   & \bestc{100}  \\
				
				&    & en      & 61.5   & 54.0   & 73.5   & 67.0   & 46.0   & 82.5   & 75.5   & 76.5   & 88.5    & \bestd{91.0}    & 59.0   & 55.5   & 59.0   & 67.0   & 54.5   & 72.0   & 81.5   & 72.0   & 89.0    & \bestc{91.0}   \\ \midrule
				
				\multirow{2}{*}{III}  & \multirow{2}{*}{\textbf{100}} & zh      & 62.5   & 75.0   & 52.5   & 55.0   & 67.5   & 80.0   & 82.5   & 90.0   & \bestd{92.5}    & 80.0    & 55.0   & 62.5   & 57.5   & 62.5   & 52.5   & 77.5   & 80.0   & 85.0   & \bestc{87.5}    & 80.0   \\
				
				&    & en      & 40.0   & 72.5   & 60.0   & 60.0   & 60.0   & 80.0   & 77.5   & 80.0   & 87.5    & \bestd{90.0}    & 50.0   & 67.5   & 57.5   & 75.0   & 45.0   & 62.5   & 60.0   & 75.0   & 85.0    & \bestc{87.5}   \\ \midrule
				
				\multirow{2}{*}{IV}  & \multirow{2}{*}{\textbf{81.0}} & zh      & 55.0   & 30.0   & 21.5   & 27.0   & 21.5   & 47.5   & 55.5   & 29.0   & 88.0    & \bestd{97.5}    & 49.0   & 25.0   & 28.5   & 40.5   & 26.0   & 36.5   & 57.0   & 52.0   & \bestc{97.5}    & 92.0   \\
				
				&    & en      & 36.0   & 26.5   & 23.0   & 27.5   & 28.5   & 36.0   & 48.5   & 34.0   & 80.5    & \bestd{91.0}    & 41.0   & 28.0   & 41.5   & 37.5   & 36.0   & 31.0   & 52.0   & 39.5   & 82.0    & \bestc{92.0}   \\ \midrule
				
				\multirow{2}{*}{V}   & \multirow{2}{*}{\textbf{94.4}}  & zh      & 45.8   & 43.7   & 45.1   & 52.8   & 61.3   & 60.6   & 57.0   & 52.8   & \bestd{85.9}    & 83.1    & 44.4   & 44.4   & 50.0   & 52.1   & 54.9   & 55.6   & 52.8   & 57.7   & 82.4    & \bestc{83.8}   \\
				
				&    & en      & 42.3   & 47.9   & 45.1   & 50.0   & 59.9   & 55.6   & 55.6   & 55.6   & \bestd{80.3}    & 78.2    & 45.1   & 45.8   & 49.3   & 46.5   & 50.0   & 54.2   & 54.2   & 58.5   & 78.9    & \bestc{80.3}   \\ \midrule
				
				\multirow{2}{*}{VI}  & \multirow{2}{*}{\textbf{81.0}}  & zh      & 44.0   & 35.0   & 43.0   & 45.0   & 63.0   & 61.0   & 63.0   & 62.0   & 55.0    & \bestd{69.0}    & 41.0   & 32.0   & 42.0   & 49.0   & 59.0   & 50.0   & 43.0   & 56.0   & 61.0    & \bestc{74.0}   \\
				
				&    & en      & 39.0   & 45.0   & 39.0   & 42.0   & 55.0   & 52.0   & 48.0   & 56.0   & 49.0    & \bestd{64.0}    & 43.0   & 45.0   & 32.0   & 38.0   & 39.0   & 41.0   & 46.0   & 57.0   & 52.0    & \bestc{73.0}   \\ \midrule
				
				\multicolumn{23}{l}{\makecell[tl]{\textbf{Non-literal Communication}: (I) Irony/Sarcasm ~~ (II) Egocentric lies ~~ (III) White lies ~~ (IV) Involuntary lies ~~ (V) Humor ~~ (VI) Faux pas}}                                                                                           \\ \midrule
				
				\multirow{2}{*}{I}   & \multirow{2}{*}{\textbf{100}}  & zh      & 30.8   & 34.6   & 42.3   & 23.1   & 34.6   & 46.2   & 73.1   & 65.4   & \bestd{92.3}    & 80.8    & 34.6   & 42.3   & 46.2   & 23.1   & 26.9   & 42.3   & 38.5   & 65.4   & \bestc{84.6}    & 80.8   \\
				
				&    & en      & 30.8   & 42.3   & 53.8   & 53.8   & 46.2   & 57.7   & 57.7   & 57.7   & \bestd{88.5}    & \bestd{88.5}    & 30.8   & 50.0   & 42.3   & 46.2   & 46.2   & 38.5   & 65.4   & 53.8   & \bestc{88.5}    & 84.6   \\ \midrule
				
				\multirow{2}{*}{II}   & \multirow{2}{*}{\textbf{95.0}} & zh      & 62.5   & 55.0   & 37.5   & 42.5   & 62.5   & 60.0   & 77.5   & \bestd{87.5}   & 82.5    & 75.0    & 55.0   & 70.0   & 47.5   & 47.5   & 57.5   & 57.5   & 77.5   & 72.5   & \bestc{80.0}    & 77.5   \\
				
				&    & en      & 40.0   & 67.5   & 40.0   & 62.5   & 65.0   & 60.0   & 72.5   & 70.0   & 80.0    & \bestd{82.5}    & 40.0   & 72.5   & 57.5   & 67.5   & 45.0   & 62.5   & 75.0   & 75.0   & 80.0    & \bestc{82.5}   \\ \midrule
				
				\multirow{2}{*}{III}  & \multirow{2}{*}{\textbf{85.0}} & zh      & 52.5   & 50.0   & 50.0   & 37.5   & 52.5   & 30.0   & \bestd{77.5}   & \bestd{77.5}   & \bestd{77.5}    & 72.5    & 62.5   & 57.5   & 60.0   & 50.0   & 47.5   & 40.0   & 70.0   & 72.5   & \bestc{90.0}    & 65.0   \\
				
				&    & en      & 35.0   & 47.5   & 40.0   & 42.5   & 55.0   & 30.0   & 65.0   & 80.0   & \bestd{82.5}    & 80.0    & 37.5   & 60.0   & 35.0   & 52.5   & 40.0   & 52.5   & \bestc{80.0}   & \bestc{80.0}   & \bestc{80.0}    & \bestc{80.0}   \\ \midrule
				
				\multirow{2}{*}{IV}  & \multirow{2}{*}{\textbf{78.6}}  & zh      & 42.9   & 66.7   & 31.0   & 57.1   & 38.1   & 45.2   & \bestd{78.6}   & 66.7   & 69.0    & 57.1    & 59.5   & 40.5   & 42.9   & 61.9   & 38.1   & 45.2   & \bestc{76.2}   & 57.1   & 71.4    & 54.8   \\
				
				&    & en      & 42.9   & 50.0   & 42.9   & 50.0   & 61.9   & 40.5   & 61.9   & 76.2   & 73.8    & \bestd{85.7}    & 38.1   & 69.0   & 42.9   & 66.7   & 45.2   & 45.2   & 69.0   & 54.8   & 73.8    & \bestc{83.3}   \\ \midrule
				
				\multirow{2}{*}{V}   & \multirow{2}{*}{\textbf{95.0}}  & zh      & 55.0   & 70.0   & 55.0   & 55.0   & 55.0   & 65.0   & 87.5   & \bestd{90.0}   & \bestd{90.0}    & 72.5    & 57.5   & 57.5   & 52.5   & 57.5   & 50.0   & 65.0   & \bestc{87.5}   & \bestc{87.5}   & 85.0    & 67.5   \\
				
				&    & en      & 37.5   & 60.0   & 62.5   & 67.5   & 65.0   & 87.5   & 77.5   & 72.5   & \bestd{92.5}    & \bestd{92.5}    & 42.5   & 62.5   & 55.0   & 77.5   & 42.5   & 52.5   & 70.0   & 90.0   & \bestc{97.5}    & 92.5   \\ \midrule
				
				\multirow{2}{*}{VI}  & \multirow{2}{*}{\textbf{80.4}}  & zh      & 55.2   & 47.9   & 61.6   & 63.6   & 62.5   & 72.7   & 66.8   & 68.8   & 76.6    & \bestd{78.6}    & 65.4   & 49.6   & 65.4   & 64.3   & 47.5   & 70.7   & 70.5   & 72.7   & 74.3    & \bestc{79.6}   \\ 
				
				&    & en      & 44.6   & 58.4   & 61.3   & 66.8   & 54.1   & 69.5   & 67.0   & 72.5   & 71.8    & \bestd{75.0}    & 51.4   & 62.1   & 52.5   & 64.1   & 54.3   & 67.7   & 70.4   & 70.9   & 73.6    & \bestc{75.2}  \\ \bottomrule
			\end{tabular}
		}
		
		\vspace{1mm}
		\caption{ToM performance for 31 specific abilities of all LLMs. The best results of vanilla prompting are in \bestd{pink}, and those of CoT prompting are in \bestc{blue}.
		}
		
		\label{tab-all-ability-results}
		\vspace{-6mm}
	\end{table*}

%% file: tables/task-results-hard.tex
\begin{table*}[h]
	\vspace{0mm}
	\small
	\renewcommand\arraystretch{1.0}
	\setlength{\abovecaptionskip}{0mm}
	\setlength{\belowcaptionskip}{5mm}
	\centering
	\setlength{\tabcolsep}{1.2mm}
		\resizebox{1.\textwidth}{!}{
			\begin{tabular}{l\myline cc|cc|cc|cc|cc|cc|cc|cc\myline cc}
				\toprule
    \multicolumn{19}{l}{\makecell[tl]{\textbb{UOT}: Unexpected Outcome Test ~~~~~~~~~~~~~~\textbb{SIT}: Scalar Implicature Task ~~~~~~~~~~~~~~\textbb{PST}: Persuasion Story Task ~~~~~~~~~~\,~~~\textbb{FBT}: False Belief Task \\ \textbb{AST}: Ambiguous Story Task ~~~~~~~~~~~~~~~~~~~~\textbb{HT}: Hinting Test~~~~~~~ ~~~~~~ ~~ ~~~~~~~~~~~~~~~~\textbb{SST}: Strange Story Task~~~~~~~ ~~~~~~~~~~~~\textbb{FRT}: Faux-pas Recognition Test}}  \\ \midrule
				& \multicolumn{2}{c|}{\textbb{UOT}} & \multicolumn{2}{c|}{\textbb{SIT}} & \multicolumn{2}{c|}{\textbb{PST$^*$}} & \multicolumn{2}{c|}{\textbb{FBT}} & \multicolumn{2}{c|}{\textbb{AST}} & \multicolumn{2}{c|}{\textbb{HT}} & \multicolumn{2}{c|}{\textbb{SST}} & \multicolumn{2}{c\myline}{\textbb{FRT}} & \multicolumn{2}{c}{\textbf{\tbbg AVG.}}  \\
				& Zh	  & En	  & Zh	  & En	  & Zh	  & En	 & Zh	  & En	  & Zh	  & En	  & Zh	  & En	  & Zh	  & En	  & Zh	  & En	  &\tbbg  Zh	  & \tbbg En	  \\ \midrule
				Human (Original)&  \multicolumn{2}{c|}{89.3}       &  \multicolumn{2}{c|}{75.5}&  \multicolumn{2}{c|}{70.0}&  \multicolumn{2}{c|}{86.8}&  \multicolumn{2}{c|}{95.0}&  \multicolumn{2}{c|}{97.1}&  \multicolumn{2}{c|}{89.2}&  \multicolumn{2}{c|}{80.4}&  \multicolumn{2}{c}{\tbbg 85.4}        \\ 
				
				Human (Coherent)&  \multicolumn{2}{c|}{74.0}       &  \multicolumn{2}{c|}{58.0}&  \multicolumn{2}{c|}{70.0}&  \multicolumn{2}{c|}{59.0}&  \multicolumn{2}{c|}{90.0}&  \multicolumn{2}{c|}{96.8}&  \multicolumn{2}{c|}{79.6}&  \multicolumn{2}{c|}{47.1}&  \multicolumn{2}{c}{\tbbg 71.8}       \\ 
				
				\textbf{Performance Drop}&  \multicolumn{2}{c|}{\textbf{15.3}}       &  \multicolumn{2}{c|}{\textbf{17.5}}&  \multicolumn{2}{c|}{\textbf{0.0}}&  \multicolumn{2}{c|}{\textbf{27.8}}&  \multicolumn{2}{c|}{\textbf{5.0}}&  \multicolumn{2}{c|}{\textbf{0.3}}&  \multicolumn{2}{c|}{\textbf{9.6}}&  \multicolumn{2}{c|}{\textbf{33.3}}&  \multicolumn{2}{c}{\textbf{\tbbg 13.6}}        \\ \midrule
				
				ChatGLM3-6B (Original)	 					& 55.3      & 44.3      & 24.5      & 28.0      & 44.0      & 41.0      & 59.2      & 48.5      & 48.0      & 41.0      & 32.0      & 36.9      & 58.0      & 37.8      & 55.2      & 44.6      & \tbbg 47.0      & \tbbg 40.3      \\
				ChatGLM3-6B (Coherent)              & 17.0 & 4.0  & 5.0  & 7.0  & 44.0 & 41.0 & 2.0  & 0.0  & 25.0 & 15.0 & 28.0 & 33.3 & 35.3 & 13.9 & 4.3  & 2.9  & \tbbg 20.1 & \tbbg 14.6 \\
				\textbf{Performance Drop}  & \textbf{38.3}   &   \textbf{40.3}   &   \textbf{19.5}   &   \textbf{21.0}   &   \textbf{0.0}   &   \textbf{0.0}   &   \textbf{57.2}   &   \textbf{48.5}   &   \textbf{23.0}   &   \textbf{26.0}   &   \textbf{4.0}   &   \textbf{3.6}   &   \textbf{22.7}   &   \textbf{23.9}   &   \textbf{50.9}   &   \textbf{41.7}   &  \tbbg  \textbf{27.0}   &   \tbbg \textbf{25.6}  \\ \midrule
				
				LLaMA2-13B-Chat (Original)    				& 43.7      & 52.7      & 28.0      & 23.5      & 38.0      & 43.0      & 42.2      & 42.8      & 38.0      & 47.5      & 32.0      & 48.5      & 58.2      & 58.0      & 47.9      & 58.4      & \tbbg 41.0      & \tbbg 46.8      \\
				LLaMA2-13B-Chat (Coherent)          & 5.0  & 8.0  & 11.0 & 5.0  & 38.0 & 43.0 & 0.0  & 1.0  & 13.0 & 23.0 & 29.0 & 44.1 & 34.8 & 32.8 & 4.3  & 7.1  & \tbbg 16.9 &\tbbg  20.5 \\
				\textbf{Performance Drop}  & \textbf{38.7}   &   \textbf{44.7}   &   \textbf{17.0}   &   \textbf{18.5}   &   \textbf{0.0}   &   \textbf{0.0}   &   \textbf{42.2}   &   \textbf{41.8}   &   \textbf{25.0}   &   \textbf{24.5}   &   \textbf{3.0}   &   \textbf{4.4}   &   \textbf{23.4}   &   \textbf{25.2}   &   \textbf{43.6}   &   \textbf{51.3}   &  \tbbg  \textbf{24.1}   &   \tbbg \textbf{26.3}  \\ \midrule
				
				Baichuan2-13B-Chat (Original) 				& 56.3      & 53.7      & 27.5      & 32.0      & 48.0      & 36.0      & 50.2      & 51.5      & 56.0      & 50.5      & 54.4      & 58.3      & 50.1      & 50.4      & 61.6      & 61.3      & \tbbg 50.5      & \tbbg 49.2      \\
				Baichuan2-13B-Chat (Coherent)       & 12.0 & 18.0 & 8.0  & 11.0 & 48.0 & 36.0 & 2.0  & 1.0  & 29.0 & 26.0 & 49.5 & 53.8 & 24.4 & 27.9 & 24.3 & 17.1 & \tbbg 24.7 & \tbbg 23.9 \\
				\textbf{Performance Drop}  & \textbf{44.3}   &   \textbf{35.7}   &   \textbf{19.5}   &   \textbf{21.0}   &   \textbf{0.0}   &   \textbf{0.0}   &   \textbf{48.2}   &   \textbf{50.5}   &   \textbf{27.0}   &   \textbf{24.5}   &   \textbf{4.9}   &   \textbf{4.5}   &   \textbf{25.7}   &   \textbf{22.5}   &   \textbf{37.3}   &   \textbf{44.2}   &  \tbbg  \textbf{25.9}   &   \tbbg \textbf{25.4}  \\ \midrule
				
				Mistral-7B (Original)	  					& 61.0      & 58.0      & 28.0      & 34.5      & 49.0      & 51.0      & 43.5      & 46.7      & 52.5      & 51.0      & 29.1      & 43.7      & 53.1      & 60.0      & 63.6      & 66.8      & \tbbg 47.5      & \tbbg 51.5      \\
				Mistral-7B (Coherent)               & 22.0 & 16.0 & 8.0  & 13.0 & 49.0 & 51.0 & 0.0  & 2.0  & 31.0 & 24.0 & 24.7 & 39.8 & 28.4 & 38.3 & 25.0 & 28.6 & \tbbg 23.5 & \tbbg 26.6 \\
				\textbf{Performance Drop}  & \textbf{39.0}   &   \textbf{42.0}   &   \textbf{20.0}   &   \textbf{21.5}   &   \textbf{0.0}   &   \textbf{0.0}   &   \textbf{43.5}   &   \textbf{44.7}   &   \textbf{21.5}   &   \textbf{27.0}   &   \textbf{4.4}   &   \textbf{3.9}   &   \textbf{24.7}   &   \textbf{21.7}   &   \textbf{38.6}   &   \textbf{38.2}   &   \tbbg \textbf{24.0}   &   \tbbg \textbf{24.9}  \\ \midrule
				
				Mixtral-8x7B (Original)					& 68.0      & 58.7      & {49.5}      & 42.5      & 45.0      & 55.0      & 49.8      & 37.8      & 71.0      & 69.5      & 43.7      & 55.3      & 51.4      & 53.8      & 62.5      & 54.1      & \tbbg 55.1      &\tbbg  53.3      \\
				Mixtral-8x7B (Coherent)             & 33.0 & 17.0 & \bestd{26.0} & \bestd{20.0} & 45.0 & 55.0 & 0.0  & 0.0  & 51.0 & 52.0 & 38.7 & 51.6 & 22.4 & 30.3 & 16.4 & 12.1 &\tbbg  29.1 & \tbbg 29.8 \\
				\textbf{Performance Drop}  & \textbf{35.0}   &   \textbf{41.7}   &   \textbf{23.5}   &   \textbf{22.5}   &   \textbf{0.0}   &   \textbf{0.0}   &   \textbf{49.8}   &   \textbf{37.8}   &   \textbf{20.0}   &   \textbf{17.5}   &   \textbf{5.0}   &   \textbf{3.7}   &   \textbf{29.0}   &   \textbf{23.5}   &   \textbf{46.1}   &   \textbf{42.0}   &   \tbbg \textbf{26.1}   &   \tbbg \textbf{23.6}  \\ \midrule
				
				Qwen-14B-Chat (Original)					& 72.0      & 63.7      & 42.5      & 30.5      & 50.0      & 51.0      & 57.2      & 58.7      & 65.5      & 64.0      & 54.4      & 56.3      & 60.0      & 59.5      & 72.7      & 69.5      & \tbbg 59.3      & \tbbg 56.7      \\ 
				Qwen-14B-Chat (Coherent)            & 33.0 & 23.0 & 21.0 & 6.0  & 50.0 & 51.0 & 0.0  & 7.0  & 44.0 & 41.0 & 49.5 & 51.6 & 35.8 & 37.3 & 32.1 & 33.6 & \tbbg 33.2 & \tbbg 31.3 \\
				\textbf{Performance Drop}  & \textbf{39.0}   &   \textbf{40.7}   &   \textbf{21.5}   &   \textbf{24.5}   &   \textbf{0.0}   &   \textbf{0.0}   &   \textbf{57.2}   &   \textbf{51.7}   &   \textbf{21.5}   &   \textbf{23.0}   &   \textbf{4.9}   &   \textbf{4.7}   &   \textbf{24.2}   &   \textbf{22.2}   &   \textbf{40.6}   &   \textbf{35.9}   &  \tbbg  \textbf{26.1}   &   \tbbg \textbf{25.3}  \\ \midrule
				
				GPT-3.5-Turbo-0613 (Original)					& 69.3      & 63.3      & 33.0      & 35.0      & 52.0      & 49.0      & 61.2      & 62.3      & 63.5      & 63.5      & 60.2      & 53.4      & 72.0      & 66.1      & 66.8      & 67.0      &\tbbg  59.8      & \tbbg 57.5      \\
				GPT-3.5-Turbo-0613 (Coherent)             & 36.0 & 24.0 & 10.0 & 14.0 & 52.0 & 49.0 & 1.0  & 1.0  & 38.0 & 42.0 & 58.1 & 49.5 & 53.2 & 44.8 & 14.3 & 16.4 & \tbbg 32.8 &\tbbg  30.1 \\
				\textbf{Performance Drop}  & \textbf{33.3}   &   \textbf{39.3}   &   \textbf{23.0}   &   \textbf{21.0}   &   \textbf{0.0}   &   \textbf{0.0}   &   \textbf{60.2}   &   \textbf{61.3}   &   \textbf{25.5}   &   \textbf{21.5}   &   \textbf{2.1}   &   \textbf{3.9}   &   \textbf{18.8}   &   \textbf{21.3}   &   \textbf{52.5}   &   \textbf{50.6}   &   \tbbg \textbf{26.9}   &   \tbbg \textbf{27.4}  \\ \midrule
				
				GPT-3.5-Turbo-1106 (Original)					& 72.3      & 66.0      & 34.0      & 33.0      & 57.0      & 56.0      & 53.0      & 55.0      & 59.0      & 60.5      & 61.2      & 64.1      & 72.5      & 69.0      & 68.8      & 72.5      & \tbbg 59.7      & \tbbg 59.5      \\
				GPT-3.5-Turbo-1106 (Coherent)             & 42.0 & 31.0 & 13.0 & 15.0 & 57.0 & 56.0 & 0.0  & 0.0  & 32.0 & 36.0 & 58.1 & 61.3 & 53.2 & 48.3 & 17.1 & 23.6 & \tbbg 34.1 & \tbbg 33.9 \\
				\textbf{Performance Drop}  & \textbf{30.3}   &   \textbf{35.0}   &   \textbf{21.0}   &   \textbf{18.0}   &   \textbf{0.0}   &   \textbf{0.0}   &   \textbf{53.0}   &   \textbf{55.0}   &   \textbf{27.0}   &   \textbf{24.5}   &   \textbf{3.1}   &   \textbf{2.8}   &   \textbf{19.3}   &   \textbf{20.7}   &   \textbf{51.7}   &   \textbf{48.9}   &   \tbbg \textbf{25.7}   &   \tbbg \textbf{25.6}  \\ \midrule
				
				GPT-4-0613 (Original)	  					& 71.3      & {71.3}      & 49.0      & 44.0      & 58.0      & 53.0      & 86.3      & 80.0      & {84.0}      & {78.0}      & 79.6      & 76.7      & {83.0}      & 81.1      & 76.6      & 71.8      & \tbbg 73.5      & \tbbg 69.5      \\
				GPT-4-0613 (Coherent)               & 37.0 & 24.0 & 21.0 & 17.0 & 58.0 & 53.0 & 27.0 & 23.0 & \bestd{72.0} & \bestd{63.0} & 77.4 & 74.2 & \bestd{68.7} & 65.7 & 41.4 & 31.4 & \tbbg 50.3 & \tbbg 43.9 \\
				\textbf{Performance Drop}  & \textbf{34.3}   &   \textbf{47.3}   &   \textbf{28.0}   &   \textbf{27.0}   &   \textbf{0.0}   &   \textbf{0.0}   &   \textbf{59.3}   &   \textbf{57.0}   &   \textbf{12.0}   &   \textbf{15.0}   &   \textbf{2.2}   &   \textbf{2.5}   &   \textbf{14.3}   &   \textbf{15.4}   &   \textbf{35.2}   &   \textbf{40.4}   &  \tbbg  \textbf{23.2}   &   \tbbg \textbf{25.6}  \\ \midrule
				
				GPT-4-1106 (Original)	  					& {76.7}      & 71.0      & 48.0      & {49.0}      & {61.0}      & {65.0}      & {90.8}      & {88.2}      & 83.0      & 77.5      & {88.3}      & {82.5}      & 76.2      & {84.0}      & {78.6}      & {75.0}      &\tbbg  {75.3}      &\tbbg  {74.0}      \\
				GPT-4-1106 (Coherent)               & \bestd{45.0} & \bestd{37.0} & 20.0 & 18.0 & \bestd{61.0} & \bestd{65.0} & \bestd{59.0} & \bestd{53.0} & 69.0 & \bestd{63.0} & \bestd{87.1} & \bestd{80.6} & 57.2 & \bestd{70.1} & \bestd{46.4} & \bestd{34.3} & \tbbg \bestd{55.6} & \tbbg \bestd{52.6} \\ 
				\textbf{Performance Drop}  & \textbf{31.7}   &   \textbf{34.0}   &   \textbf{28.0}   &   \textbf{31.0}   &   \textbf{0.0}   &   \textbf{0.0}   &   \textbf{31.8}   &   \textbf{35.2}   &   \textbf{14.0}   &   \textbf{14.5}   &   \textbf{1.2}   &   \textbf{1.9}   &   \textbf{19.0}   &   \textbf{13.9}   &   \textbf{32.2}   &   \textbf{40.7}   &  \tbbg  \textbf{19.7}   &   \tbbg \textbf{21.4}  \\ \midrule

				ChatGLM3-6B + CoT (Original)				& 58.7      & 50.3      & 27.0      & 26.5      & 44.0      & 41.0      & 56.5      & 51.2      & 48.0      & 44.0      & 37.9      & 42.7      & 56.3      & 44.2      & 65.4      & 51.4      & \tbbg 49.2      &\tbbg  43.9      \\
				ChatGLM3-6B + CoT (Coherent)        & 23.0 & 11.0 & 11.0 & 6.0  & 44.0 & 41.0 & 0.0  & 1.0  & 25.0 & 16.0 & 32.3 & 36.6 & 34.3 & 14.9 & 14.3 & 7.9  & \tbbg 23.0 & \tbbg 16.8 \\
				\textbf{Performance Drop}  & \textbf{35.7}   &   \textbf{39.3}   &   \textbf{16.0}   &   \textbf{20.5}   &   \textbf{0.0}   &   \textbf{0.0}   &   \textbf{56.5}   &   \textbf{50.2}   &   \textbf{23.0}   &   \textbf{28.0}   &   \textbf{5.6}   &   \textbf{6.1}   &   \textbf{22.0}   &   \textbf{29.3}   &   \textbf{51.1}   &   \textbf{43.5}   &  \tbbg  \textbf{26.2}   &  \tbbg  \textbf{27.1}  \\ \midrule
				
				LLaMA2-13B-Chat + CoT (Original)			& 45.3      & 52.7      & 25.5      & 23.5      & 34.0      & 39.0      & 41.3      & 43.0      & 41.0      & 48.5      & 28.2      & 43.7      & 53.6      & 59.5      & 49.6      & 62.1      &\tbbg  39.8      & \tbbg 46.5      \\
				LLaMA2-13B-Chat + CoT (Coherent)    & 8.0  & 9.0  & 6.0  & 4.0  & 34.0 & 39.0 & 0.0  & 1.0  & 16.0 & 27.0 & 24.7 & 39.8 & 31.8 & 35.8 & 3.6  & 9.3  & \tbbg 15.5 & \tbbg 20.6 \\
				\textbf{Performance Drop}  & \textbf{37.3}   &   \textbf{43.7}   &   \textbf{19.5}   &   \textbf{19.5}   &   \textbf{0.0}   &   \textbf{0.0}   &   \textbf{41.3}   &   \textbf{42.0}   &   \textbf{25.0}   &   \textbf{21.5}   &   \textbf{3.5}   &   \textbf{3.9}   &   \textbf{21.8}   &   \textbf{23.7}   &   \textbf{46.0}   &   \textbf{52.8}   &   \tbbg \textbf{24.3}   &   \tbbg \textbf{25.9}  \\ \midrule
				
				Baichuan2-13B-Chat + CoT (Original)		& 54.3      & 48.7      & 26.5      & 23.0      & 33.0      & 34.0      & 44.8      & 44.2      & 51.5      & 44.0      & 53.4      & 49.5      & 52.8      & 51.1      & 65.4      & 52.5      & \tbbg 47.7      & \tbbg 43.4      \\
				Baichuan2-13B-Chat + CoT (Coherent) & 14.0 & 7.0  & 4.0  & 6.0  & 33.0 & 34.0 & 0.0  & 0.0  & 23.0 & 21.0 & 48.4 & 46.2 & 25.9 & 25.9 & 21.4 & 8.6  & \tbbg 21.2 & \tbbg 18.6 \\
				\textbf{Performance Drop}  & \textbf{40.3}   &   \textbf{41.7}   &   \textbf{22.5}   &   \textbf{17.0}   &   \textbf{0.0}   &   \textbf{0.0}   &   \textbf{44.8}   &   \textbf{44.2}   &   \textbf{28.5}   &   \textbf{23.0}   &   \textbf{5.0}   &   \textbf{3.3}   &   \textbf{26.9}   &   \textbf{25.2}   &   \textbf{44.0}   &   \textbf{43.9}   &   \tbbg \textbf{26.5}   &   \tbbg \textbf{24.8}  \\ \midrule
				
				Mistral-7B + CoT (Original)				& 61.0      & 55.3      & 27.0      & 28.0      & 46.0      & 42.0      & 47.2      & 47.0      & 47.0      & 46.5      & 30.1      & 37.9      & 56.5      & 63.4      & 64.3      & 64.1      & \tbbg 47.4      & \tbbg 48.0      \\
				Mistral-7B + CoT (Coherent)         & 19.0 & 13.0 & 8.0  & 7.0  & 46.0 & 42.0 & 4.0  & 1.0  & 23.0 & 24.0 & 25.8 & 32.3 & 33.3 & 40.8 & 27.9 & 25.0 & \tbbg 23.4 & \tbbg 23.1 \\
				\textbf{Performance Drop}  & \textbf{42.0}   &   \textbf{42.3}   &   \textbf{19.0}   &   \textbf{21.0}   &   \textbf{0.0}   &   \textbf{0.0}   &   \textbf{43.2}   &   \textbf{46.0}   &   \textbf{24.0}   &   \textbf{22.5}   &   \textbf{4.3}   &   \textbf{5.6}   &   \textbf{23.2}   &   \textbf{22.6}   &   \textbf{36.4}   &   \textbf{39.1}   &   \tbbg \textbf{24.0}   &   \tbbg \textbf{24.9}  \\ \midrule
				
				Mixtral-8x7B + CoT (Original)				& 65.3      & 52.3      & 45.0      & 29.5      & 41.0      & 39.0      & 53.7      & 43.8      & 66.0      & 59.5      & 44.7      & 54.4      & 43.7      & 39.8      & 47.5      & 54.3      & \tbbg 50.9      & \tbbg 46.6      \\ 
				Mixtral-8x7B + CoT (Coherent)       & 25.0 & 15.0 & \bestc{22.0} & 9.0  & 41.0 & 39.0 & 1.0  & 0.0  & 45.0 & 37.0 & 39.8 & 51.6 & 13.9 & 12.9 & 6.4  & 10.0 & \tbbg 24.3 & \tbbg 21.8 \\
				\textbf{Performance Drop}  & \textbf{40.3}   &   \textbf{37.3}   &   \textbf{23.0}   &   \textbf{20.5}   &   \textbf{0.0}   &   \textbf{0.0}   &   \textbf{52.7}   &   \textbf{43.8}   &   \textbf{21.0}   &   \textbf{22.5}   &   \textbf{4.9}   &   \textbf{2.8}   &   \textbf{29.8}   &   \textbf{26.9}   &   \textbf{41.1}   &   \textbf{44.3}   &  \tbbg  \textbf{26.6}   &\tbbg    \textbf{24.8}  \\ \midrule
				
				Qwen-14B-Chat + CoT (Original)				& 65.3      & 58.0      & 31.5      & 31.0      & 45.0      & 44.0      & 51.3      & 54.7      & 62.5      & 63.0      & 47.6      & 48.5      & 60.2      & 53.6      & 70.7      & 67.7      & \tbbg 54.3      &\tbbg  52.6      \\
				Qwen-14B-Chat + CoT (Coherent)      & 22.0 & 16.0 & 9.0  & 8.0  & 45.0 & 44.0 & 1.0  & 1.0  & 43.0 & 43.0 & 45.2 & 43.0 & 37.3 & 27.4 & 27.1 & 25.7 & \tbbg 28.7 & \tbbg 26.0 \\
				\textbf{Performance Drop}  & \textbf{43.3}   &   \textbf{42.0}   &   \textbf{22.5}   &   \textbf{23.0}   &   \textbf{0.0}   &   \textbf{0.0}   &   \textbf{50.3}   &   \textbf{53.7}   &   \textbf{19.5}   &   \textbf{20.0}   &   \textbf{2.4}   &   \textbf{5.5}   &   \textbf{22.9}   &   \textbf{26.2}   &   \textbf{43.6}   &   \textbf{42.0}   &   \tbbg \textbf{25.6}   &   \tbbg \textbf{26.6}  \\ \midrule
				
				GPT-3.5-Turbo-0613 + CoT (Original)				& 62.3      & 58.3      & 30.0      & 26.5      & 43.0      & 48.0      & 57.8      & 64.0      & 58.5      & 58.0      & 41.7      & 41.7      & 71.3      & 66.8      & 70.5      & 70.4      & \tbbg 54.4      & \tbbg 54.2      \\
				GPT-3.5-Turbo-0613 + CoT (Coherent)       & 20.0 & 17.0 & 9.0  & 9.0  & 43.0 & 48.0 & 3.0  & 8.0  & 33.0 & 33.0 & 37.6 & 36.6 & 53.2 & 45.3 & 35.7 & 33.6 & \tbbg 29.3 &\tbbg  28.8 \\
				\textbf{Performance Drop}  & \textbf{42.3}   &   \textbf{41.3}   &   \textbf{21.0}   &   \textbf{17.5}   &   \textbf{0.0}   &   \textbf{0.0}   &   \textbf{54.8}   &   \textbf{56.0}   &   \textbf{25.5}   &   \textbf{25.0}   &   \textbf{4.1}   &   \textbf{5.1}   &   \textbf{18.1}   &   \textbf{21.5}   &   \textbf{34.8}   &   \textbf{36.8}   &   \tbbg \textbf{25.1}   &   \tbbg \textbf{25.4}  \\ \midrule
				
				GPT-3.5-Turbo-1106 + CoT (Original)				& 68.7      & 64.7      & 27.5      & 35.0      & 45.0      & 54.0      & 57.5      & 56.3      & 61.5      & 63.0      & 46.6      & 51.5      & 71.3      & 68.6      & 72.7      & 70.9      &\tbbg  56.4      & \tbbg 58.0      \\ 
				GPT-3.5-Turbo-1106 + CoT (Coherent)       & 32.0 & 25.0 & 8.0  & 12.0 & 45.0 & 54.0 & 0.0  & 3.0  & 40.0 & 41.0 & 43.0 & 47.3 & 49.8 & 47.8 & 28.6 & 27.1 & \tbbg 30.8 & \tbbg 32.2 \\
				\textbf{Performance Drop}  & \textbf{36.7}   &   \textbf{39.7}   &   \textbf{19.5}   &   \textbf{23.0}   &   \textbf{0.0}   &   \textbf{0.0}   &   \textbf{57.5}   &   \textbf{53.3}   &   \textbf{21.5}   &   \textbf{22.0}   &   \textbf{3.6}   &   \textbf{4.2}   &   \textbf{21.5}   &   \textbf{20.8}   &   \textbf{44.1}   &   \textbf{43.8}   &  \tbbg  \textbf{25.6}   &   \tbbg \textbf{25.9}  \\ \midrule
				
				GPT-4-0613 + CoT (Original)				& 72.3      & 64.7      & 43.5      & 54.0      & 55.0      & 52.0      & {90.3}      & 80.8      & {84.5}      & 77.5      & 78.6      & 76.7      & {83.5}      & 81.1      & 74.3      & 73.6      & \tbbg 72.8      & \tbbg 70.1      \\
				GPT-4-0613 + CoT (Coherent)         & \bestc{43.0} & 24.0 & 14.0 & 26.0 & 55.0 & 52.0 & 46.0 & 20.0 & 72.0 & 63.0 & 77.4 & 74.2 & \bestc{69.2} & 68.2 & 37.1 & \bestc{34.3} & \tbbg 51.7 & \tbbg 45.2 \\
				\textbf{Performance Drop}  & \textbf{29.3}   &   \textbf{40.7}   &   \textbf{29.5}   &   \textbf{28.0}   &   \textbf{0.0}   &   \textbf{0.0}   &   \textbf{44.3}   &   \textbf{60.8}   &   \textbf{12.5}   &   \textbf{14.5}   &   \textbf{1.2}   &   \textbf{2.5}   &   \textbf{14.3}   &   \textbf{12.9}   &   \textbf{37.2}   &   \textbf{39.3}   &   \tbbg \textbf{21.0}   &   \tbbg \textbf{24.8}  \\ \midrule
				
				GPT-4-1106 + CoT (Original)				& {76.3}      & {72.7}      & {48.0}      & {55.0}      & {59.0}      & {55.0}      & 88.7      & {86.8}      & 84.0      & {81.0}      & {89.3}      & {82.5}      & 76.9      & {84.3}      & {79.6}      & {75.2}      & \tbbg {75.2}      & \tbbg {74.1}      \\
				GPT-4-1106 + CoT (Coherent)         & 42.0 & \bestc{38.0} & 20.0 & \bestc{28.0} & \bestc{59.0} & \bestc{55.0} & \bestc{47.0} & \bestc{39.0} & \bestc{74.0} & \bestc{69.0} & \bestc{88.2} & \bestc{80.6} & 59.2 & \bestc{71.1} & \bestc{46.4} & 32.9 & \tbbg \bestc{54.5} & \tbbg \bestc{51.7} \\
				\textbf{Performance Drop}  & \textbf{34.3}   &   \textbf{34.7}   &   \textbf{28.0}   &   \textbf{27.0}   &   \textbf{0.0}   &   \textbf{0.0}   &   \textbf{41.7}   &   \textbf{47.8}   &   \textbf{10.0}   &   \textbf{12.0}   &   \textbf{1.1}   &   \textbf{1.9}   &   \textbf{17.7}   &   \textbf{13.2}   &   \textbf{33.2}   &   \textbf{42.3}   &\tbbg   \textbf{20.8}   &  \tbbg \textbf{22.4}  \\ \bottomrule
			\end{tabular}
		}
		
		\vspace{1mm}
		\caption{The ToM performance drop of all LLMs in the coherent test. 
  The best results of vanilla prompting in coherent tests are in \bestd{pink}, and those of CoT prompting are in \bestc{blue}. No performance drop occurs in the persuasion story task since each story in \texttt{PST} corresponds to only one question.
		}
		
		\label{tab-task-coherent-results}
		\vspace{-6mm}
	\end{table*}

%% file: acl_latex.bbl
\begin{thebibliography}{74}
\expandafter\ifx\csname natexlab\endcsname\relax\def\natexlab#1{#1}\fi

\bibitem[{Achiam et~al.(2023)Achiam, Adler, Agarwal, Ahmad, Akkaya, Aleman,
  Almeida, Altenschmidt, Altman, Anadkat et~al.}]{achiam2023gpt}
Josh Achiam, Steven Adler, Sandhini Agarwal, Lama Ahmad, Ilge Akkaya,
  Florencia~Leoni Aleman, Diogo Almeida, Janko Altenschmidt, Sam Altman,
  Shyamal Anadkat, et~al. 2023.
\newblock Gpt-4 technical report.
\newblock \emph{arXiv preprint arXiv:2303.08774}.

\bibitem[{AN(1933)}]{an1933sulla}
KOLMOGOROV AN. 1933.
\newblock Sulla determinazione empirica di una legge didistribuzione.
\newblock \emph{Giorn Dell'inst Ital Degli Att}, 4:89--91.

\bibitem[{Aronson and Golomb(1999)}]{aronson1999preschoolers}
James~N Aronson and Claire Golomb. 1999.
\newblock Preschoolers' understanding of pretense and presumption of congruity
  between action and representation.
\newblock \emph{Developmental Psychology}, 35(6):1414.

\bibitem[{Bai et~al.(2023)Bai, Bai, Chu, Cui, Dang, Deng, Fan, Ge, Han, Huang
  et~al.}]{bai2023qwen}
Jinze Bai, Shuai Bai, Yunfei Chu, Zeyu Cui, Kai Dang, Xiaodong Deng, Yang Fan,
  Wenbin Ge, Yu~Han, Fei Huang, et~al. 2023.
\newblock Qwen technical report.
\newblock \emph{arXiv preprint arXiv:2309.16609}.

\bibitem[{{Baichuan-Inc}(2023)}]{Baichuan2}
{Baichuan-Inc}. 2023.
\newblock \href {https://github.com/baichuan-inc/Baichuan2} {Baichuan 2}.
\newblock Online.

\bibitem[{Baron-Cohen et~al.(1985)Baron-Cohen, Leslie, and
  Frith}]{baron1985does}
Simon Baron-Cohen, Alan~M Leslie, and Uta Frith. 1985.
\newblock Does the autistic child have a “theory of mind”?
\newblock \emph{Cognition}, 21(1):37--46.

\bibitem[{Baron-Cohen et~al.(1999)Baron-Cohen, O'riordan, Stone, Jones, and
  Plaisted}]{baron1999recognition}
Simon Baron-Cohen, Michelle O'riordan, Valerie Stone, Rosie Jones, and Kate
  Plaisted. 1999.
\newblock Recognition of faux pas by normally developing children and children
  with asperger syndrome or high-functioning autism.
\newblock \emph{Journal of autism and developmental disorders}, 29:407--418.

\bibitem[{Beaudoin et~al.(2020)Beaudoin, Leblanc, Gagner, and
  Beauchamp}]{beaudoin2020systematic}
Cindy Beaudoin, {\'E}lizabel Leblanc, Charlotte Gagner, and Miriam~H Beauchamp.
  2020.
\newblock Systematic review and inventory of theory of mind measures for young
  children.
\newblock \emph{Frontiers in psychology}, 10:2905.

\bibitem[{Bennett and Galpert(1993)}]{bennett1993children}
Mark Bennett and Linda Galpert. 1993.
\newblock Children's understanding of multiple desires.
\newblock \emph{International Journal of Behavioral Development}, 16(1):15--33.

\bibitem[{Borke(1971)}]{borke1971interpersonal}
Helene Borke. 1971.
\newblock Interpersonal perception of young children: Egocentrism or empathy?
\newblock \emph{Developmental psychology}, 5(2):263.

\bibitem[{Bosacki and Wilde~Astington(1999)}]{bosacki1999theory}
Sandra Bosacki and Janet Wilde~Astington. 1999.
\newblock Theory of mind in preadolescence: Relations between social
  understanding and social competence.
\newblock \emph{Social development}, 8(2):237--255.

\bibitem[{Bradford et~al.(2018)Bradford, Jentzsch, Gomez, Chen, Zhang, and
  Su}]{bradford2018cross}
Elisabeth~EF Bradford, Ines Jentzsch, Juan-Carlos Gomez, Yulu Chen, Da~Zhang,
  and Yanjie Su. 2018.
\newblock Cross-cultural differences in adult theory of mind abilities: a
  comparison of native-english speakers and native-chinese speakers on the
  self/other differentiation task.
\newblock \emph{Quarterly Journal of Experimental Psychology},
  71(12):2665--2676.

\bibitem[{Brambring and Asbrock(2010)}]{brambring2010validity}
Michael Brambring and Doreen Asbrock. 2010.
\newblock Validity of false belief tasks in blind children.
\newblock \emph{Journal of Autism and Developmental Disorders}, 40:1471--1484.

\bibitem[{Bubeck et~al.(2023)Bubeck, Chandrasekaran, Eldan, Gehrke, Horvitz,
  Kamar, Lee, Lee, Li, Lundberg et~al.}]{bubeck2023sparks}
S{\'e}bastien Bubeck, Varun Chandrasekaran, Ronen Eldan, Johannes Gehrke, Eric
  Horvitz, Ece Kamar, Peter Lee, Yin~Tat Lee, Yuanzhi Li, Scott Lundberg,
  et~al. 2023.
\newblock Sparks of artificial general intelligence: Early experiments with
  gpt-4.
\newblock \emph{arXiv preprint arXiv:2303.12712}.

\bibitem[{Caputi et~al.(2012)Caputi, Lecce, Pagnin, and
  Banerjee}]{caputi2012longitudinal}
Marcella Caputi, Serena Lecce, Adriano Pagnin, and Robin Banerjee. 2012.
\newblock Longitudinal effects of theory of mind on later peer relations: the
  role of prosocial behavior.
\newblock \emph{Developmental psychology}, 48(1):257.

\bibitem[{Carlson and Moses(2001)}]{carlson2001individual}
Stephanie~M Carlson and Louis~J Moses. 2001.
\newblock Individual differences in inhibitory control and children's theory of
  mind.
\newblock \emph{Child development}, 72(4):1032--1053.

\bibitem[{Colonnesi et~al.(2008)Colonnesi, Rieffe, Koops, and
  Perucchini}]{colonnesi2008precursors}
Cristina Colonnesi, Carolien Rieffe, Willem Koops, and Paola Perucchini. 2008.
\newblock Precursors of a theory of mind: A longitudinal study.
\newblock \emph{British Journal of Developmental Psychology}, 26(4):561--577.

\bibitem[{Corcoran et~al.(1995)Corcoran, Mercer, and
  Frith}]{corcoran1995schizophrenia}
Rhiannon Corcoran, Gavin Mercer, and Christopher~D Frith. 1995.
\newblock Schizophrenia, symptomatology and social inference: investigating
  “theory of mind” in people with schizophrenia.
\newblock \emph{Schizophrenia research}, 17(1):5--13.

\bibitem[{Decety and Jackson(2004)}]{decety2004functional}
Jean Decety and Philip~L Jackson. 2004.
\newblock The functional architecture of human empathy.
\newblock \emph{Behavioral and cognitive neuroscience reviews}, 3(2):71--100.

\bibitem[{Denham(1986)}]{denham1986social}
Susanne~A Denham. 1986.
\newblock Social cognition, prosocial behavior, and emotion in preschoolers:
  Contextual validation.
\newblock \emph{Child development}, pages 194--201.

\bibitem[{Dyck et~al.(2001)Dyck, Ferguson, and Shochet}]{dyck2001autism}
Murray~J Dyck, Kara Ferguson, and Ian~M Shochet. 2001.
\newblock Do autism spectrum disorders differ from each other and from
  non-spectrum disorders on emotion recognition tests?
\newblock \emph{European child \& adolescent psychiatry}, 10:105--116.

\bibitem[{Flavell et~al.(1986)Flavell, Green, Flavell, Watson, and
  Campione}]{flavell1986development}
John~H Flavell, Frances~L Green, Eleanor~R Flavell, Malcolm~W Watson, and
  Joseph~C Campione. 1986.
\newblock Development of knowledge about the appearance-reality distinction.
\newblock \emph{Monographs of the society for research in child development},
  pages i--87.

\bibitem[{Golchin and Surdeanu(2023)}]{golchin2023time}
Shahriar Golchin and Mihai Surdeanu. 2023.
\newblock Time travel in llms: Tracing data contamination in large language
  models.
\newblock \emph{arXiv preprint arXiv:2308.08493}.

\bibitem[{Goodman and Stuhlm{\"u}ller(2013)}]{goodman2013knowledge}
Noah~D Goodman and Andreas Stuhlm{\"u}ller. 2013.
\newblock Knowledge and implicature: Modeling language understanding as social
  cognition.
\newblock \emph{Topics in cognitive science}, 5(1):173--184.

\bibitem[{Gordis et~al.(1989)}]{gordis1989young}
Felice~W Gordis et~al. 1989.
\newblock Young children's understanding of simultaneous conflicting emotions.

\bibitem[{Happ{\'e}(1994)}]{happe1994advanced}
Francesca~GE Happ{\'e}. 1994.
\newblock An advanced test of theory of mind: Understanding of story
  characters' thoughts and feelings by able autistic, mentally handicapped, and
  normal children and adults.
\newblock \emph{Journal of autism and Developmental disorders}, 24(2):129--154.

\bibitem[{Harris et~al.(1986)Harris, Donnelly, Guz, and
  Pitt-Watson}]{harris1986children}
Paul~L Harris, Kara Donnelly, Gabrielle~R Guz, and Rosemary Pitt-Watson. 1986.
\newblock Children's understanding of the distinction between real and apparent
  emotion.
\newblock \emph{Child development}, pages 895--909.

\bibitem[{Hogrefe et~al.(1986)Hogrefe, Wimmer, and
  Perner}]{hogrefe1986ignorance}
G-Juergen Hogrefe, Heinz Wimmer, and Josef Perner. 1986.
\newblock Ignorance versus false belief: A developmental lag in attribution of
  epistemic states.
\newblock \emph{Child development}, pages 567--582.

\bibitem[{Jiang et~al.(2023)Jiang, Sablayrolles, Mensch, Bamford, Chaplot,
  Casas, Bressand, Lengyel, Lample, Saulnier et~al.}]{jiang2023mistral}
Albert~Q Jiang, Alexandre Sablayrolles, Arthur Mensch, Chris Bamford,
  Devendra~Singh Chaplot, Diego de~las Casas, Florian Bressand, Gianna Lengyel,
  Guillaume Lample, Lucile Saulnier, et~al. 2023.
\newblock Mistral 7b.
\newblock \emph{arXiv preprint arXiv:2310.06825}.

\bibitem[{Jones et~al.(2023)Jones, Trott, and Bergen}]{jones2023epitome}
Cameron~Robert Jones, Sean Trott, and Ben Bergen. 2023.
\newblock Epitome: Experimental protocol inventory for theory of mind
  evaluation.
\newblock In \emph{First Workshop on Theory of Mind in Communicating Agents}.

\bibitem[{Killen et~al.(2011)Killen, Mulvey, Richardson, Jampol, and
  Woodward}]{killen2011accidental}
Melanie Killen, Kelly~Lynn Mulvey, Cameron Richardson, Noah Jampol, and Amanda
  Woodward. 2011.
\newblock The accidental transgressor: Morally-relevant theory of mind.
\newblock \emph{Cognition}, 119(2):197--215.

\bibitem[{Kim et~al.(2023)Kim, Sclar, Zhou, Bras, Kim, Choi, and
  Sap}]{kim2023fantom}
Hyunwoo Kim, Melanie Sclar, Xuhui Zhou, Ronan Bras, Gunhee Kim, Yejin Choi, and
  Maarten Sap. 2023.
\newblock Fantom: A benchmark for stress-testing machine theory of mind in
  interactions.
\newblock In \emph{EMNLP}, pages 14397--14413.

\bibitem[{Kinderman et~al.(1998)Kinderman, Dunbar, and
  Bentall}]{kinderman1998theory}
Peter Kinderman, Robin Dunbar, and Richard~P Bentall. 1998.
\newblock Theory-of-mind deficits and causal attributions.
\newblock \emph{British journal of Psychology}, 89(2):191--204.

\bibitem[{Klie et~al.(2023)Klie, Webber, and Gurevych}]{klie2023annotation}
Jan-Christoph Klie, Bonnie Webber, and Iryna Gurevych. 2023.
\newblock Annotation error detection: Analyzing the past and present for a more
  coherent future.
\newblock \emph{Computational Linguistics}, 49(1):157--198.

\bibitem[{Knafo et~al.(2009)Knafo, Zahn-Waxler, Davidov, Van~Hulle, Robinson,
  and Rhee}]{knafo2009empathy}
Ariel Knafo, Carolyn Zahn-Waxler, Maayan Davidov, Carol Van~Hulle, JoAnn~L
  Robinson, and Soo~Hyun Rhee. 2009.
\newblock Empathy in early childhood: Genetic, environmental, and affective
  contributions.
\newblock \emph{Annals of the New York Academy of Sciences}, 1167(1):103--114.

\bibitem[{Ko{\l}odziejczyk and Bosacki(2016)}]{kolodziejczyk2016young}
Anna~M Ko{\l}odziejczyk and Sandra~L Bosacki. 2016.
\newblock Young-school-aged children’s use of direct and indirect persuasion:
  role of intentionality understanding.
\newblock \emph{Psychology of Language and Communication}, 20(3):292--315.

\bibitem[{Kosinski(2023)}]{kosinski2023theory}
Michal Kosinski. 2023.
\newblock Theory of mind may have spontaneously emerged in large language
  models.
\newblock \emph{arXiv preprint arXiv:2302.02083}.

\bibitem[{Lapuschkin et~al.(2019)Lapuschkin, W{\"a}ldchen, Binder, Montavon,
  Samek, and M{\"u}ller}]{lapuschkin2019unmasking}
Sebastian Lapuschkin, Stephan W{\"a}ldchen, Alexander Binder, Gr{\'e}goire
  Montavon, Wojciech Samek, and Klaus-Robert M{\"u}ller. 2019.
\newblock Unmasking clever hans predictors and assessing what machines really
  learn.
\newblock \emph{Nature communications}, 10(1):1096.

\bibitem[{Le et~al.(2019)Le, Boureau, and Nickel}]{Le2019RevisitingTE}
Matt Le, Y-Lan Boureau, and Maximilian Nickel. 2019.
\newblock Revisiting the evaluation of theory of mind through question
  answering.
\newblock In \emph{EMNLP}.

\bibitem[{Li and Flanigan(2023)}]{li2023task}
Changmao Li and Jeffrey Flanigan. 2023.
\newblock Task contamination: Language models may not be few-shot anymore.
\newblock \emph{arXiv preprint arXiv:2312.16337}.

\bibitem[{Ma et~al.(2023{\natexlab{a}})Ma, Gao, and Xu}]{ma2023tomchallenges}
Xiaomeng Ma, Lingyu Gao, and Qihui Xu. 2023{\natexlab{a}}.
\newblock Tomchallenges: A principle-guided dataset and diverse evaluation
  tasks for exploring theory of mind.
\newblock \emph{arXiv preprint arXiv:2305.15068}.

\bibitem[{Ma et~al.(2023{\natexlab{b}})Ma, Sansom, Peng, and
  Chai}]{ma2023towards}
Ziqiao Ma, Jacob Sansom, Run Peng, and Joyce Chai. 2023{\natexlab{b}}.
\newblock Towards a holistic landscape of situated theory of mind in large
  language models.
\newblock In \emph{Findings of the Association for Computational Linguistics:
  EMNLP 2023}, pages 1011--1031.

\bibitem[{Meltzoff(1995)}]{meltzoff1995understanding}
Andrew~N Meltzoff. 1995.
\newblock Understanding the intentions of others: Re-enactment of intended acts
  by 18-month-old children.
\newblock \emph{Developmental psychology}, 31(5):838.

\bibitem[{{Mistral AI}(2023)}]{Mixtral}
{Mistral AI}. 2023.
\newblock \href {https://mistral.ai/news/mixtral-of-experts/} {Mixtral of
  experts: A high quality sparse mixture-of-experts}.
\newblock Online.

\bibitem[{Moll et~al.(2006)Moll, Koring, Carpenter, and
  Tomasello}]{moll2006infants}
Henrike Moll, Cornelia Koring, Malinda Carpenter, and Michael Tomasello. 2006.
\newblock Infants determine others' focus of attention by pragmatics and
  exclusion.
\newblock \emph{Journal of Cognition and Development}, 7(3):411--430.

\bibitem[{{OpenAI}(2023{\natexlab{a}})}]{GPT35-0613}
{OpenAI}. 2023{\natexlab{a}}.
\newblock \href
  {https://community.openai.com/t/gpt-3-5-turbo-0613-function-calling-16k-context-window-and-lower-prices/263263}
  {Gpt-3.5-turbo-0613: Function calling, 16k context window, and lower prices}.
\newblock Online.

\bibitem[{{OpenAI}(2023{\natexlab{b}})}]{GPT4-1106}
{OpenAI}. 2023{\natexlab{b}}.
\newblock \href
  {https://openai.com/blog/new-models-and-developer-products-announced-at-devday}
  {New models and developer products announced at devday}.
\newblock Online.

\bibitem[{Perner and Wimmer(1985)}]{perner1985john}
Josef Perner and Heinz Wimmer. 1985.
\newblock “john thinks that mary thinks that…” attribution of
  second-order beliefs by 5-to 10-year-old children.
\newblock \emph{Journal of experimental child psychology}, 39(3):437--471.

\bibitem[{Peskin et~al.(2014)Peskin, Prusky, and Comay}]{peskin2014keeping}
Joan Peskin, Carly Prusky, and Julie Comay. 2014.
\newblock Keeping the reader's mind in mind: development of perspective-taking
  in children's dictations.
\newblock \emph{Journal of applied developmental psychology}, 35(1):35--43.

\bibitem[{Phillips et~al.(2002)Phillips, Wellman, and
  Spelke}]{phillips2002infants}
Ann~T Phillips, Henry~M Wellman, and Elizabeth~S Spelke. 2002.
\newblock Infants' ability to connect gaze and emotional expression to
  intentional action.
\newblock \emph{Cognition}, 85(1):53--78.

\bibitem[{Pillow(1989)}]{pillow1989early}
Bradford~H Pillow. 1989.
\newblock Early understanding of perception as a source of knowledge.
\newblock \emph{Journal of experimental child psychology}, 47(1):116--129.

\bibitem[{Pons and Harris(2000)}]{pons2000test}
Francisco Pons and Paul Harris. 2000.
\newblock \emph{Test of emotion comprehension: TEC}.
\newblock University of Oxford.

\bibitem[{Premack and Woodruff(1978)}]{premack1978tom}
David Premack and Guy Woodruff. 1978.
\newblock Does the chimpanzee have a theory of mind?
\newblock \emph{Behavioral and brain sciences}, pages 515--526.

\bibitem[{Quesque and Rossetti(2020)}]{quesque2020theory}
Fran{\c{c}}ois Quesque and Yves Rossetti. 2020.
\newblock What do theory-of-mind tasks actually measure? theory and practice.
\newblock \emph{Perspectives on Psychological Science}, 15(2):384--396.

\bibitem[{Repacholi and Gopnik(1997)}]{repacholi1997early}
Betty~M Repacholi and Alison Gopnik. 1997.
\newblock Early reasoning about desires: evidence from 14-and 18-month-olds.
\newblock \emph{Developmental psychology}, 33(1):12.

\bibitem[{Sap et~al.(2022)Sap, Bras, Fried, and Choi}]{Sap2022NeuralTO}
Maarten Sap, Ronan~Le Bras, Daniel Fried, and Yejin Choi. 2022.
\newblock Neural theory-of-mind? on the limits of social intelligence in large
  lms.
\newblock \emph{ArXiv}.

\bibitem[{Sap et~al.(2019)Sap, Rashkin, Chen, LeBras, and
  Choi}]{sap2019socialiqa}
Maarten Sap, Hannah Rashkin, Derek Chen, Ronan LeBras, and Yejin Choi. 2019.
\newblock Socialiqa: Commonsense reasoning about social interactions.
\newblock \emph{arXiv preprint arXiv:1904.09728}.

\bibitem[{Shapira et~al.(2023)Shapira, Levy, Alavi, Zhou, Choi, Goldberg, Sap,
  and Shwartz}]{shapira2023clever}
Natalie Shapira, Mosh Levy, Seyed~Hossein Alavi, Xuhui Zhou, Yejin Choi, Yoav
  Goldberg, Maarten Sap, and Vered Shwartz. 2023.
\newblock Clever hans or neural theory of mind? stress testing social reasoning
  in large language models.
\newblock \emph{arXiv preprint arXiv:2305.14763}.

\bibitem[{Simon and Simon(1977)}]{simon1977spurious}
Herbert~A Simon and Herbert~A Simon. 1977.
\newblock \emph{Spurious correlation: A causal interpretation}.
\newblock Springer.

\bibitem[{Slaughter et~al.(2002)Slaughter, Dennis, and
  Pritchard}]{slaughter2002theory}
Virginia Slaughter, Michelle~J Dennis, and Michelle Pritchard. 2002.
\newblock Theory of mind and peer acceptance in preschool children.
\newblock \emph{British journal of developmental psychology}, 20(4):545--564.

\bibitem[{Smiley(2001)}]{smiley2001intention}
Patricia~A Smiley. 2001.
\newblock Intention understanding and partner-sensitive behaviors in young
  children's peer interactions.
\newblock \emph{Social Development}, 10(3):330--354.

\bibitem[{Sullivan et~al.(1995)Sullivan, Winner, and
  Hopfield}]{sullivan1995children}
Kate Sullivan, Ellen Winner, and Natalie Hopfield. 1995.
\newblock How children tell a lie from a joke: The role of second-order mental
  state attributions.
\newblock \emph{British journal of developmental psychology}, 13(2):191--204.

\bibitem[{Swettenham(1996)}]{swettenham1996can}
J~Swettenham. 1996.
\newblock Can children be taught to understand false belief using computers?
  child psychology \& psychiatry \& allied disciplines, 37 (2), 157--165.

\bibitem[{{THUDM}(2023)}]{ChatGLM3}
{THUDM}. 2023.
\newblock \href {https://github.com/THUDM/ChatGLM3} {Chatglm3}.
\newblock Online.

\bibitem[{Touvron et~al.(2023)Touvron, Martin, Stone, Albert, Almahairi,
  Babaei, Bashlykov, Batra, Bhargava, Bhosale et~al.}]{touvron2023llama}
Hugo Touvron, Louis Martin, Kevin Stone, Peter Albert, Amjad Almahairi, Yasmine
  Babaei, Nikolay Bashlykov, Soumya Batra, Prajjwal Bhargava, Shruti Bhosale,
  et~al. 2023.
\newblock Llama 2: Open foundation and fine-tuned chat models.
\newblock \emph{arXiv preprint arXiv:2307.09288}.

\bibitem[{Ullman(2023)}]{ullman2023large}
Tomer Ullman. 2023.
\newblock Large language models fail on trivial alterations to theory-of-mind
  tasks.
\newblock \emph{arXiv preprint arXiv:2302.08399}.

\bibitem[{van Duijn et~al.(2023)van Duijn, van Dijk, Kouwenhoven, de~Valk,
  Spruit, and van~der Putten}]{van2023theory}
Max~J van Duijn, Bram van Dijk, Tom Kouwenhoven, Werner de~Valk, Marco~R
  Spruit, and Peter van~der Putten. 2023.
\newblock Theory of mind in large language models: Examining performance of 11
  state-of-the-art models vs. children aged 7-10 on advanced tests.
\newblock \emph{arXiv preprint arXiv:2310.20320}.

\bibitem[{Wellman and Bartsch(1988)}]{wellman1988young}
Henry~M Wellman and Karen Bartsch. 1988.
\newblock Young children's reasoning about beliefs.
\newblock \emph{Cognition}, 30(3):239--277.

\bibitem[{Wilcoxon(1947)}]{wilcoxon1947individual}
Frank Wilcoxon. 1947.
\newblock Individual comparisons of grouped data by ranking methods.

\bibitem[{Wilf et~al.(2023)Wilf, Lee, Liang, and Morency}]{wilf2023think}
Alex Wilf, Sihyun~Shawn Lee, Paul~Pu Liang, and Louis-Philippe Morency. 2023.
\newblock Think twice: Perspective-taking improves large language models'
  theory-of-mind capabilities.
\newblock \emph{arXiv preprint arXiv:2311.10227}.

\bibitem[{Wimmer and Perner(1983)}]{wimmer1983beliefs}
Heinz Wimmer and Josef Perner. 1983.
\newblock Beliefs about beliefs: Representation and constraining function of
  wrong beliefs in young children's understanding of deception.
\newblock \emph{Cognition}, 13(1):103--128.

\bibitem[{Wu et~al.(2023)Wu, He, Jia, Mihalcea, Chen, and Deng}]{wu2023hi}
Yufan Wu, Yinghui He, Yilin Jia, Rada Mihalcea, Yulong Chen, and Naihao Deng.
  2023.
\newblock Hi-tom: A benchmark for evaluating higher-order theory of mind
  reasoning in large language models.
\newblock In \emph{Findings of the Association for Computational Linguistics:
  EMNLP 2023}, pages 10691--10706.

\bibitem[{Zheng et~al.(2023)Zheng, Zhou, Meng, Zhou, and
  Huang}]{zheng2023large}
Chujie Zheng, Hao Zhou, Fandong Meng, Jie Zhou, and Minlie Huang. 2023.
\newblock On large language models' selection bias in multi-choice questions.
\newblock \emph{arXiv preprint arXiv:2309.03882}.

\bibitem[{Zhou et~al.(2023)Zhou, Madaan, Potharaju, Gupta, McKee, Holtzman,
  Pujara, Ren, Mishra, Nematzadeh et~al.}]{zhou2023far}
Pei Zhou, Aman Madaan, Srividya~Pranavi Potharaju, Aditya Gupta, Kevin~R McKee,
  Ari Holtzman, Jay Pujara, Xiang Ren, Swaroop Mishra, Aida Nematzadeh, et~al.
  2023.
\newblock How far are large language models from agents with theory-of-mind?
\newblock \emph{arXiv preprint arXiv:2310.03051}.

\end{thebibliography}
